%% file: 0-main.tex
\documentclass{article}

\usepackage[final]{corl_2026}

\input{0-preamble}

\input{0-defs}

\hypersetup{
  pdftitle={MESA: An Evaluation Framework for Compositional, Semantic, and Spatial Generalization in Robotics},
  pdfauthor={Albert Wilcox, Frank Chang, Nhi Nguyen, Aishani Chakraborty, Jeremy A. Collins, Vaibhav Saxena, Benjamin Joffe, Siddharth Karamcheti, Animesh Garg},
  pdfsubject={Robotics, imitation learning, and generalization benchmarks}
}

\title{%
\texorpdfstring{%
\includegraphics[height=1.1em]{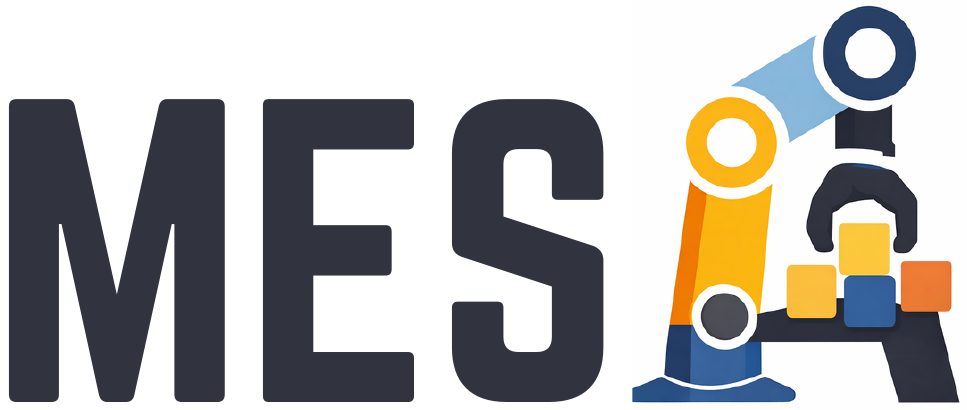}%
}{MESA}%
: An Evaluation Framework for Compositional, Semantic, and Spatial Generalization in Robotics}

\author{
  \normalfont Albert Wilcox$^{1,2,*}$ \quad Frank Chang$^{1,*}$ \quad Nhi Nguyen$^1$\\
  \normalfont Aishani Chakraborty$^1$ \quad Jeremy A.~Collins$^1$ \quad Vaibhav Saxena$^1$\\
  \normalfont Benjamin Joffe$^{1,2}$ \quad Siddharth Karamcheti$^1$ \quad Animesh Garg$^1$\\
  \normalfont $^1$Georgia Institute of Technology\\
  \normalfont $^2$Georgia Tech Research Institute\\
  \normalfont $^*$Equal contribution.
}

\begin{document}
\maketitle
\begin{figure}[h]
    \centering
    \includegraphics[width=0.81\textwidth]{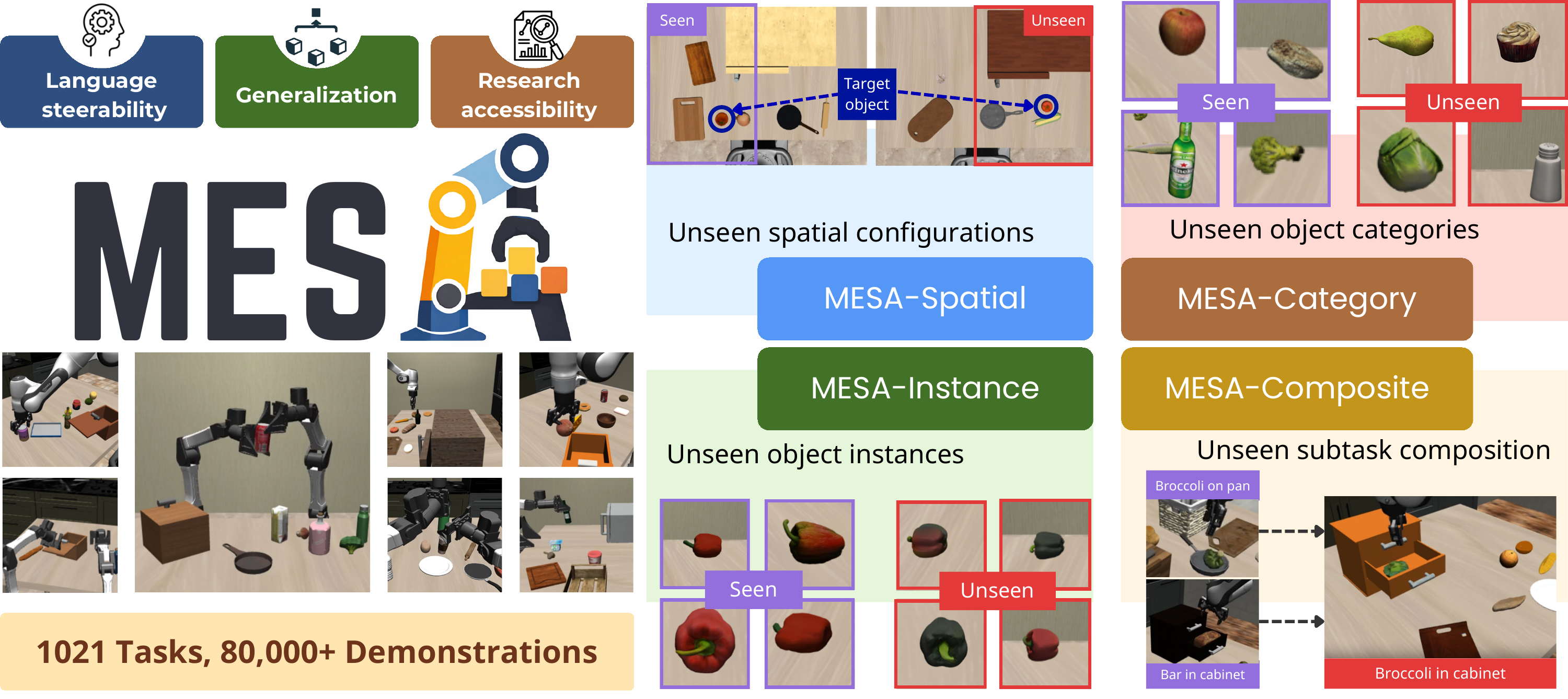}
    \caption{\abbr{} (\textbf{Left}) is an evaluation framework designed to enable researchers to probe language steering and generalization capabilities in settings accessible to researchers with academic budgets. We introduce \abbr{}-Bench (\textbf{Right}), which includes single-arm and bimanual datasets and test sets designed to probe generalization to unseen spatial configurations, object instances, object categories and subtask compositions.}
    \label{fig:teaser}
\end{figure}


\begin{abstract}
    Despite growing interest in foundation models for robotic manipulation, systematic evaluation of their capabilities remains elusive. Existing benchmarks are either narrow and saturated, offering little signal on SOTA methods, broadly scoped, providing diversity at the cost of controlled experimentation, or require training on large, messy open-source datasets. We introduce MESA, an evaluation framework designed for precise, reproducible measurement of language-conditioned policy generalization. MESA provides five evaluation suites that isolate in-distribution performance and four distinct axes of generalization: spatial configuration, object instance, object category, and subtask composition, enabling researchers to diagnose where and why policies fail, not just whether they fail. 
    To populate these suites at scale, we develop MESA-Gen, a pipeline for task and demonstration generation built by improving MimicGen. 
    We instantiate the framework on two platforms and release both: the single-arm MESA suite, and BiMESA, the first large-scale benchmark for bimanual generalization.
    We evaluate seven baseline methods spanning diffusion policies, VLM-based approaches, and pretrained VLAs, finding that the strongest model achieves only 52\% on MESA and 49\% on BiMESA, providing meaningful signal while leaving ample room for progress.
    MESA rankings are broadly consistent with RoboArena scores, suggesting that MESA captures method-level differences relevant beyond simulation.
    Project website: \url{https://pairlab.github.io/MESA}.
\end{abstract}



\input{1-intro}

\input{2-related-work}
\input{3-mesa-gen}
\input{4-mesa-70}

\input{5-experiments}
\input{6-conclusion}


\clearpage
\acknowledgments{The authors would like to acknowledge the State of Georgia, the Agricultural Technology Research Program at Georgia Tech, and microagi for supporting the work described in this paper. We are thankful to all the reviewers whose comments helped to strengthen the paper.}


\bibliography{references}  

\input{7-appendix}

\end{document}

%% file: 0-preamble.tex
\usepackage{etoolbox} 
\usepackage[utf8]{inputenc} 
\usepackage[T1]{fontenc}    
\usepackage{fonttable}
\usepackage[english]{babel} 
\usepackage[protrusion=true,expansion=true]{microtype}
\usepackage{comment}
\usepackage{cancel}
\usepackage{fvextra}
\usepackage{csquotes}
\usepackage{listings}
\usepackage{nameref}
\usepackage{paralist}
\usepackage{xspace}
\usepackage{setspace}
\usepackage{makeidx}
\usepackage{pdfpages}
\usepackage{etoc} 
\usepackage[dvipsnames,table]{xcolor}

\usepackage{times}
\usepackage{microtype}

\usepackage{amsmath,amssymb} 
\usepackage{amsfonts}       
\usepackage{mathrsfs} 
\usepackage{mathtools}
\usepackage{array} 
\usepackage{nicefrac}       
\usepackage{breqn}          
\usepackage{textcomp}
\usepackage{bm} 
\usepackage{pifont}
\usepackage[
  separate-uncertainty = true,
  multi-part-units = repeat
]{siunitx}
\usepackage{afterpage}
\usepackage{wasysym}

\usepackage{graphicx}
\usepackage{wrapfig}
\usepackage{adjustbox} 
\usepackage{epsfig}

\usepackage{float}
\usepackage{stfloats}

\usepackage{placeins}
\usepackage{subcaption}
\usepackage[font=small,labelfont=bf]{caption}
\usepackage{lscape}      

\usepackage{tikz}  
\usepackage{makecell}
\usepackage{overpic}
\usepackage{algorithm}
\usepackage[noend]{algpseudocode}

\usepackage{array} 
\usepackage{tabularx}
\usepackage{multirow}
\usepackage{multicol}
\usepackage{booktabs}   
\usepackage{tablefootnote}
\usepackage{longtable}

\usepackage{paralist}
\usepackage{enumitem}
\setlist[itemize]{noitemsep,leftmargin=*,topsep=0in}
\setlist[enumerate]{noitemsep,leftmargin=*,topsep=0in}

\usepackage{url}            
\PassOptionsToPackage{table}{xcolor}
\usepackage{color,colortbl} 
\usepackage{soul}

\PassOptionsToPackage{capitalize, noabbrev}{cleveref}

\hypersetup{pagebackref=false,breaklinks=true,colorlinks,urlcolor=blue,citecolor=blue,linkcolor=blue,bookmarks=false}

\usepackage{blindtext}
\usepackage{bbm}

\usepackage{lipsum}

\usepackage{titlesec}
\titlespacing{\section}{0pt}{0.25\baselineskip}{0.2\baselineskip}
\titlespacing{\subsection}{0pt}{0.15\baselineskip}{0.1\baselineskip}
\titlespacing{\subsubsection}{0pt}{0.05\baselineskip}{0.03\baselineskip}

\renewcommand{\paragraph}[1]{\vspace{0.2em}\noindent\textit{#1} --}

\usepackage[rightcaption]{sidecap} 
\sidecaptionvpos{table}{c}         

\input{figures/code-highlighting}
\VerbatimPygments{\PYG}{\PYG}
\def\PYGZhy{\mbox{-}}

%% file: 0-defs.tex
\newcommand*\abbr{MESA}

\makeatletter

\renewcommand*{\sectionautorefname}{\Snospace}
\renewcommand*{\subsectionautorefname}{\Snospace}
\renewcommand*{\subsubsectionautorefname}{\Snospace}

\makeatother

\def\subsubsectionautorefname{\sectionautorefname}

\newcommand{\appref}[1]{App.~\ref{#1}}

\definecolor[named]{xDarkGray}{HTML}{404040}

\definecolor[named]{TODORed}{HTML}{FF0000}
\definecolor[named]{TLDRViolet}{HTML}{800080}
\definecolor[named]{MissingCyan}{HTML}{47D4FF}
\definecolor[named]{PendingSaffron}{HTML}{F9C22E}
\definecolor[named]{SkeletonGray}{HTML}{808080}

\definecolor[named]{cBlue}{HTML}{18647E}
\definecolor[named]{cOrange}{HTML}{FF9B00}  
\definecolor[named]{cPink}{HTML}{F26DF9}
\definecolor[named]{cYellow}{HTML}{F0C808}

\def \papermode{final}  

\def \draftmode{draft}

\ifx \papermode \draftmode
    \newcommand{\todo}[1]{{\color{TODORed}#1}}
    \newcommand{\TODO}[1]{\textbf{\color{TODORed}[TODO]: #1}}
    
    \newcommand{\tldr}[1]{\medskip \noindent {\color{TLDRViolet}{[TL;DR] :: #1}}}

    \newcommand{\placeholder}[1]{{\color{PendingSaffron}\textbf{[PLACEHOLDER]:} #1}}
    \newcommand{\skeleton}[1]{{\color{SkeletonGray} #1}}

    \newcommand{\makecomment}[3]{\noindent {\color{#2}[\textbf{#1}]: #3}}

\else
    \newcommand{\todo}[1]{}
    \newcommand{\TODO}[1]{}
    \newcommand{\tldr}[1]{}

    \newcommand{\placeholder}[1]{}
    \newcommand{\skeleton}[1]{}
    \newcommand{\warning}[1]{}
    \newcommand{\maybe}[1]{}
    \newcommand{\makecomment}[3]{}
\fi


%% file: 1-intro.tex
\section{Introduction}

Benchmarks shape the trajectory of research.
They define what the community measures, and hence what it optimizes for.
In robotic manipulation, recent progress in robotic foundation models (RFMs) has produced open-ended language-conditioned policies, long-horizon policies, and impressively dexterous manipulation policies \cite{rt-2, kim2024openvla, black2024pi0, pertsch2025fastefficientactiontokenization,  geminiroboticsteam2025geminiroboticsbringingai, nvidia2025gr00tn1openfoundation, molmoact2025, intelligence2025pi05, geminiroboticsteam2025geminirobotics15pushing, eo1, cheang2025gr3technicalreport, shukor2025smolvlavisionlanguageactionmodelaffordable, intelligence2026pi07steerablegeneralistrobotic, ye2026worldactionmodelszeroshot, pai2025mimicvideo}.
Despite rapid progress, the capabilities, limitations, and failure modes of these models remain poorly understood.
Current evaluation practices make systematic understanding of RFMs difficult, with prominent VLA efforts like $\pi$ \cite{black2024pi0,intelligence2025pi05,intelligence2026pi07steerablegeneralistrobotic}, GR00T \cite{nvidia2025gr00tn1openfoundation} and MolmoAct \cite{molmoact2025, fang2026molmoact2actionreasoningmodels} using completely different backbones, training data, architectural choices and evaluation protocols. This makes it difficult to isolate which design choices actually matter for generalization, sample efficiency, or language grounding.

Simulation offers a path toward reproducible evaluation, but existing benchmarks suffer from several complementary limitations.
Some are narrow and saturated: LIBERO~\cite{liu2024libero} is still used frequently despite state-of-the-art RFMs exceeding 98\% success rate~\cite{kim2026cosmospolicyfinetuningvideo}.
Other simulation benchmarks are diverse, but imprecise: RoboCasa~\cite{nasiriany2024robocasa} and Behavior-1K~\cite{li2024behavior1khumancenteredembodiedai} provide myriad assets, but offer no clear methodology for experimentation with controlled distribution shifts. 
Recent work proposes evaluating real data-trained policies with a digital twin \cite{yang2026robolab, jain2025polarisscalablerealtosimevaluations, jangir2025robotarenainftyscalablerobot, molmospaces2026, sedlacek2025realmrealtosimvalidatedbenchmark}, but this requires training on the large, messy DROID \cite{khazatsky2024droid} dataset. 
Finally, despite growing interest in bimanual manipulation, there are no multitask bimanual manipulation benchmarks commonly used by RFM researchers.

We propose that evaluation sets should isolate \textbf{specific axes of generalization}, enabling diagnosis of \textit{where} and \textit{why} policies fail. Evaluation should require policies to \textbf{follow language instructions}, rather than memorize task-specific behaviors or deduce tasks from visual observation alone. Frameworks should be \textbf{extensible}, enabling researchers to generate new tasks and evaluation suites as model capabilities evolve.  Finally, they should support \textbf{fast iteration}, with training and evaluation feasible under limited compute.

To meet these objectives, we introduce \abbr{}, an extensible evaluation framework for language-conditioned robotic manipulation.
Our contributions are as follows:

\noindent \textbf{1. \abbr{}-Bench}, a benchmark containing single-arm Franka and bimanual YAM embodiments, each instantiated with a canonical training set and five evaluation suites targeting in-distribution performance, spatial generalization, instance generalization, category generalization, and compositional generalization.
We introduce a language-following metric based on distractor-task completion to disentangle instruction-grounding failures from control-execution failures. 

\smallskip \noindent \textbf{2. \abbr{}-Gen}, a framework for scalable task and dataset generation.
Researchers can curate targeted evaluations for emerging questions and synthesize the corresponding data to power their experiments.
We introduce improvements to MimicGen \cite{mandlekar2023mimicgen} that enable higher-quality data generation under greater spatial variation and support generation for novel tasks.

\smallskip \noindent \textbf{3. Rigorous experiments} evaluating seven methods spanning diffusion policies, VLM-based approaches, and pretrained VLAs.
In the bimanual setting, our strongest baseline achieves only 48.7\% success rate, dropping to 18.6\% on novel object categories, suggesting substantial headroom for future improvement.
Results correlate with real-world rankings, providing preliminary evidence that MESA captures method-level differences relevant beyond simulation.

%% file: 2-related-work.tex
\section{Related Work}
\label{sec:related-work}

\input{tables/rw-comparisons}

Simulation offers reproducible, scalable evaluation without the cost and variability of real-world rollouts.
However, existing simulated benchmarks for RFMs are insufficient, and we share a detailed taxonomy in \autoref{tab:taxonomy}.

\medskip \noindent \textbf{Simulated Benchmarks.}
A useful benchmark for modern RFMs should provide axis-aligned splits that isolate specific generalization axes, remain tractable to train and evaluate on academic compute, and target a setting compatible with RFM pretraining distributions.
No existing simulated benchmark satisfies all three.
LIBERO~\cite{liu2024libero} and its predecessors~\cite{yu2020meta,james2019rlbench,mees2022calvin} are tractable and embodiment-compatible but saturated, with recent RFMs exceeding 98\% success~\cite{kim2026cosmospolicyfinetuningvideo}. 
OOD follow-ups attempt to address this~\cite{fei2025liberoplusindepthrobustnessanalysis,zhou2025liberopro} but inherit the original overfit-prone training data and yield near-zero success rate for all current RFMs, offering little signal for ranking relative progress.
On the other hand, broad suites such as RoboCasa~\cite{nasiriany2024robocasa,robocasa365} and Behavior-1K~\cite{li2024behavior1khumancenteredembodiedai} target large mobile-manipulation or long-horizon household activities that fall outside typical RFM pretraining regimes and lack targeted OOD evaluation sets. 
Orthogonally, several benchmarks provide bimanual settings without sufficient diversity to benchmark language following and generalization \cite{grotz2024peract2,robomimic2021,chen2025robotwin20scalabledata,wang2025roboevalroboticmanipulationmeets,gu2023maniskill2unifiedbenchmarkgeneralizable}.
\abbr{} satisfies all these requirements by providing axis-aligned evaluation suites in a tabletop manipulation setting, with a training set that is tractable to train on academic compute while still providing meaningful signal for modern RFMs. 
Additionally, \abbr{} is, to our knowledge, the first benchmark to test language steerability and axis-aligned generalization splits for both single-arm and bimanual embodiments.

\medskip \noindent \textbf{Real-to-Sim Benchmarks.}
Many recent works evaluate policies trained from large-scale real datasets in simulation \cite{li24simpler, jangir2025robotarenainftyscalablerobot, jain2025polarisscalablerealtosimevaluations, molmospaces2026, sedlacek2025realmrealtosimvalidatedbenchmark,yang2026robolab}. In particular, several involve training on the expansive DROID dataset \cite{khazatsky2024droid} and evaluating on a simulated digital twin \cite{jain2025polarisscalablerealtosimevaluations, molmospaces2026, sedlacek2025realmrealtosimvalidatedbenchmark,yang2026robolab}. 
This paradigm has several drawbacks. 
First, DROID's scale and opaque coverage make it expensive to train on and difficult to construct OOD settings that are guaranteed to be absent from training.
Second, the dataset is messy, with many inaccurate language instructions or long meaningless actions, such that simply filtering it can yield outsized performance gains~\cite{fang2026molmoact2actionreasoningmodels}, confounding the true source of any reported improvement. 
On the other hand, \abbr{} provides small, canonical datasets with sufficient diversity to benchmark language grounding and generalization without excessive scale slowing iteration.

%% file: tables/rw-comparisons.tex
\begin{table*}[t]
\centering
\scriptsize
\begin{tabular}{lccccccc}
\toprule
\rowcolor[HTML]{FFDEB4}
Benchmark & Num. Tasks & Obj. / Insts & MT & Action Space & Train Data & OOD & Bim. \\
\midrule
RLBench \cite{james2019rlbench}
& 100 & 28 / 28 & {\color{ForestGreen}\checkmark} & Keyframe & Scripted
& {\color{red}\ding{55}} & {\color{red}\ding{55}} \\

\rowcolor[HTML]{F8F8F8}
PerAct2 \cite{grotz2024peract2}
& 13 & 15 / 15 & {\color{ForestGreen}\checkmark} & Keyframe & Scripted
& {\color{red}\ding{55}} & {\color{ForestGreen}\checkmark} \\

COLOSSEUM \cite{pumacay2024colosseum}
& 20 & 28 / 28 & {\color{red}\ding{55}} & Keyframe & Scripted
& {\color{ForestGreen}\checkmark} & {\color{red}\ding{55}} \\

\rowcolor[HTML]{F8F8F8}
CALVIN \cite{mees2022calvin}
& 34 & 5 / 30 & {\color{ForestGreen}\checkmark} & Low-level & Human
& {\color{red}\ding{55}} & {\color{red}\ding{55}} \\

MetaWorld \cite{yu2020meta}
& 50 & 38 / 38 & {\color{red}\ding{55}} & Low-level & None
& {\color{red}\ding{55}} & {\color{red}\ding{55}} \\

\rowcolor[HTML]{F8F8F8}
LIBERO \cite{liu2024libero}
& 130 & 51 / 75 & {\color{ForestGreen}\checkmark} & Low-level & Human
& {\color{red}\ding{55}} & {\color{red}\ding{55}} \\

SIMPLER \cite{li24simpler}
& 8 & 17 / 17 & {\color{ForestGreen}\checkmark} & Low-level & External
& {\color{red}\ding{55}} & {\color{red}\ding{55}} \\

\rowcolor[HTML]{F8F8F8}
Behavior-1K \cite{li2024behavior1khumancenteredembodiedai}
& 1000 & 2211 / 9331 & {\color{ForestGreen}\checkmark} & Both & None
& {\color{red}\ding{55}} & {\color{ForestGreen}\checkmark} \\

VLABench \cite{zhang2024vlabench}
& 100 & 163 / 2164 & {\color{ForestGreen}\checkmark} & Keyframe & Scripted
& {\color{ForestGreen}\checkmark} & {\color{red}\ding{55}} \\

\rowcolor[HTML]{F8F8F8}
Robomimic \cite{robomimic2021}
& 8 & 12 / 12 & {\color{red}\ding{55}} & Low-level & Human
& {\color{red}\ding{55}} & {\color{ForestGreen}\checkmark} \\

RoboCasa \cite{nasiriany2024robocasa}
& 100 & 153 / 2509 & {\color{ForestGreen}\checkmark} & Low-level & MG 1$\rightarrow$1
& {\color{red}\ding{55}} & {\color{red}\ding{55}} \\

\rowcolor[HTML]{F8F8F8}
RoboCasa365 \cite{robocasa365}
& 365 & 153 / 2509 & {\color{ForestGreen}\checkmark} & Low-level & MG 1$\rightarrow$1
& {\color{ForestGreen}\checkmark} & {\color{red}\ding{55}} \\

RoboTwin 2.0 \cite{chen2025robotwin20scalabledata}
& 50 & 147 / 731 & {\color{red}\ding{55}} & Low-level & Scripted
& {\color{red}\ding{55}} & {\color{ForestGreen}\checkmark} \\

\rowcolor[HTML]{F8F8F8}
VLA-Arena \cite{zhang2025vlaarenaopensourceframeworkbenchmarking}
& 170 & 153 / 2509 & {\color{ForestGreen}\checkmark} & Low-level & Human
& {\color{ForestGreen}\checkmark} & {\color{red}\ding{55}} \\

MolmoSpaces \cite{molmospaces2026}
& 8 & 2.8k / 130k & {\color{ForestGreen}\checkmark} & Both & Scripted
& {\color{red}\ding{55}} & {\color{red}\ding{55}} \\

\rowcolor[HTML]{F8F8F8}
REALM \cite{sedlacek2025realmrealtosimvalidatedbenchmark}
& 7 & 3500+ & {\color{ForestGreen}\checkmark} & Low-level & External
& {\color{red}\ding{55}} & {\color{red}\ding{55}} \\

RoboLab \cite{yang2026robolab}
& 120 & $\sim$96 / 298 & {\color{ForestGreen}\checkmark} & Low-level & External
& {\color{red}\ding{55}} & {\color{red}\ding{55}} \\

\rowcolor[HTML]{F8F8F8}
\midrule
MESA
& 1021 & 153 / 2509 & {\color{ForestGreen}\checkmark} & Low-level & MG few$\rightarrow$many
& {\color{ForestGreen}\checkmark} & {\color{ForestGreen}\checkmark} \\
\bottomrule
\end{tabular}
\caption{\textbf{Comparison of robot manipulation evaluation suites.}
MESA is, to our knowledge, the first benchmark to combine large-scale multitask language-conditioned evaluation with low-level control, controlled OOD splits, and bimanual tasks.
It uses MESA-Gen to generate demonstrations for many tasks from a small set of source demonstrations and reusable subtasks, enabling benchmark-specific training data at scale.
For MESA, Num.\ Tasks counts all single-arm and bimanual training and evaluation tasks. Obj.\ / Insts: object categories / unique object instances; MT: multitask language-conditioned; Train Data: source of provided training demonstrations, where MG 1$\rightarrow$1 uses MimicGen to expand human demonstrations of each task and MG few$\rightarrow$many generates data for many tasks from source demonstrations of a few; OOD: controlled out-of-distribution evaluation splits; Bim.: bimanual tasks.
}
\label{tab:taxonomy}
\end{table*}

%% file: 3-mesa-gen.tex
\section{Generating Benchmarks with \abbr{}-Gen}
\label{sec:mesa}

\begin{figure*}[t!]
  \centering
  \includegraphics[width=\textwidth]{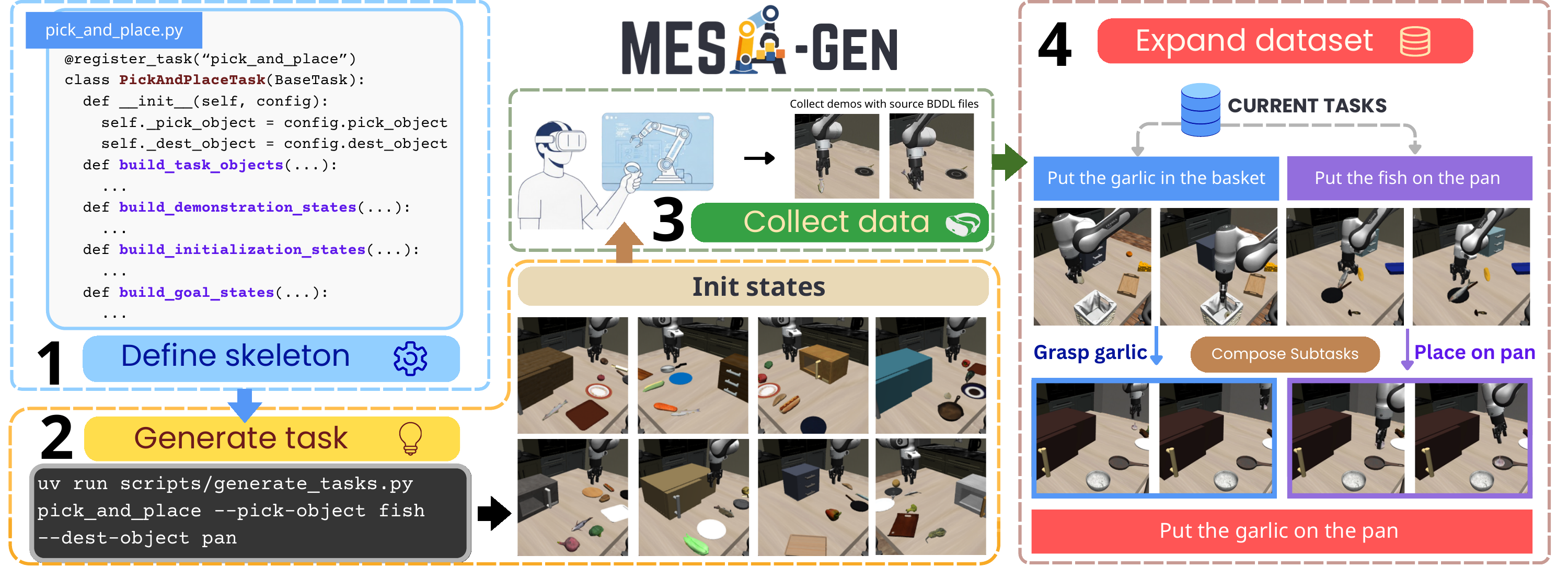}
  \caption{\textbf{\abbr{}-Gen experiment building framework}. (1) If a suitable task skeleton does not already exist, define it in Python. (2) Instantiate a task from the skeleton and generate BDDL files for the training and evaluation sets. (3) Optionally, collect a small number of source demonstrations. (4) Generate a large dataset for imitation learning. Using subtask stitching enables data generation for tasks without human data.  }
  \label{fig:generation}
\end{figure*}

\abbr{}-Gen is a framework for procedurally generating manipulation tasks and corresponding demonstration data (\autoref{fig:generation}), which we use to instantiate \abbr{}-Bench (\autoref{sec:benchmark}).
A core principle is that every demonstration and evaluation trajectory uses a unique scene with a unique set of distractor objects, making language following critical even in-distribution: the policy must disambiguate between multiple plausible goals.
Concretely, we specify each task as in- and out-of-distribution BDDL files \cite{McDermott1998PDDLthePD,srivastava2021behaviorbenchmarkeverydayhousehold} and use MimicGen \cite{mandlekar2023mimicgen} to generate in-distribution demonstrations.

\subsection{Generating Diverse Manipulation Tasks}
\label{sec:task-gen}

The first step in \abbr{}-Gen is to procedurally generate task variants for training and evaluation from a \emph{task skeleton}: a Python class specifying initialization, demonstration, and goal predicates (\appref{sec:supp_pick_and_place_skeleton}).
Given a skeleton and a set of object categories, \abbr{} emits BDDL files for the training and evaluation sets, each with a unique scene layout, distractor set, and, optionally, varied textures, camera pose, and robot embodiment (\autoref{fig:generation}).

\noindent \textbf{Ensuring Visual Confusion:} To benchmark language steerability, we must prevent the policy from inferring the task solely from visual observations.
If a scene contains only one interactable object, the policy may ignore the language instruction and simply act on that object, a form of \textit{causal confusion} \cite{dehaan2019causalconfusionimitationlearning}.
To ensure high observation-conditioned task distribution entropy, we sample distractors such that every scene contains multiple seen object-destination pairs, as elaborated in \appref{sec:supp_distractors}.

\subsection{Scalable Data Generation with MimicGen}
\label{sec:data_gen}

To generate demonstrations at scale, \abbr{}-Gen builds on MimicGen~\cite{mandlekar2023mimicgen}, an object-centric data generation pipeline. We extend it along two axes.
First, we introduce \emph{subtask stitching}: rather than requiring a human demonstration for every new task, \abbr{}-Gen composes object-centric subtasks (e.g., \texttt{Grasp(apple)}, \texttt{On($\cdot$, plate)}) drawn from a shared library indexed by predicate type and object semantics, enabling data generation for tasks with no human source data (\appref{sec:supp_subtask_stitching}).
Second, we introduce \emph{trajectory optimizations} that improve data generation quality under large source-to-target object pose differences, avoiding the unnatural sweeping motions and self-collisions that vanilla MimicGen produces under large spatial variation (\appref{sec:supp_traj_opt}).
Together, these enable \abbr{}-Gen to generate high-quality data for novel tasks under substantially greater spatial variation than prior MimicGen-based pipelines.

\subsection{Evaluating Language Following}
\label{sec:metrics}

Recent VLA benchmarks have emphasized \textit{language following} as a key capability \cite{pertsch2025fastefficientactiontokenization, intelligence2025pi05, driess2025knowledgeinsulatingvisionlanguageactionmodels, nvidia2025gr00tn1openfoundation}, but the term is often defined imprecisely. 
To address this, \abbr{}-Gen uses the BDDL predicate system to explicitly measure whether a policy completed the intended task or an alternative distractor task. 
For each scene, we enumerate feasible distractor tasks from pairs of distractor objects and detect when a failed rollout completes one. We define the \textbf{language following rate} as $L = \frac{S}{S + D}$ where $S$ is the success rate and $D$ is the distractor task completion rate. Intuitively, this measures the probability that, conditioned on completing some feasible task, the policy completed the correct one.

%% file: 4-mesa-70.tex
\section{MESA-Bench}
\label{sec:benchmark}

We use \abbr{}-Gen to instantiate \abbr{}-Bench, a fixed benchmark to support accessible and reproducible policy evaluation.
\abbr{}-Bench includes two embodiments: a DROID-like single-arm Franka Panda \cite{khazatsky2024droid} and a bimanual YAM setup. For each embodiment \textbf{we provide a training set (\autoref{sec:training-set})} and five evaluation suites (\autoref{sec:suites}) probing in-distribution performance and OOD performance along four axes of generalization.
Benchmark construction and per-suite details are in \appref{sec:supp_bench_lists}.

\subsection{Training Set}
\label{sec:training-set}

We provide one shared training set per embodiment, used for all evaluation sets. The single-arm set, \abbr{}-70, contains 7,000 demonstrations across 70 tasks, while the bimanual set, BiMESA-57, contains 6,833 demonstrations across 57 tasks. We additionally provide 736 auxiliary single-arm tasks, \abbr{}-Aux, for studying diversity scaling (\autoref{sec:scaling-diversity}). Further dataset details are provided in \appref{sec:app-dataset-detail}.

\subsection{Evaluation Suites}
\label{sec:suites}

For all trials, we sample an unseen set of distractor objects and poses and generate a fixed set of initial states for unbiased comparisons across policies.
The full task lists and partition definitions are in \appref{sec:supp_bench_lists}, and per-task success rates are in \appref{sec:per-task-sr}.

\textbf{\abbr{}-ID} tests policy performance on in-distribution tasks: 70 single-arm tasks (\abbr{}-70) and 57 bimanual tasks (BiMESA-57).
Each scene has an unseen set of distractor objects, meaning that policies must precisely follow language instructions to succeed in this setting.

\textbf{\abbr{}-Spatial} evaluates policies under changes in object location.
The single-arm setting contains 10 tasks: we partition the table into two halves $\mathcal{R}_{\mathrm{ID}}$ and $\mathcal{R}_{\mathrm{OOD}}$ (one left, one right) and only initialize the target objects in $\mathcal{R}_{\mathrm{ID}}$ during training and $\mathcal{R}_{\mathrm{OOD}}$ during evaluation.
The bimanual setting contains 38 tasks and tests a form of spatial generalization with no single-arm analogue: relocating the scene changes which arm must act and how the arms coordinate, requiring policies to solve held-out arm assignments, reversed handoff directions, and cross-body interactions (\appref{sec:supp_bimanual_suites}).

\textbf{\abbr{}-Instance} tests generalization to unseen object instances with 20 single-arm and 18 bimanual tasks.
For each task, we train on one subset of object assets and evaluate on a disjoint held-out subset.

\textbf{\abbr{}-Composite} consists of unseen compositions of seen subtasks with 20 single-arm and 15 bimanual tasks.
These include tasks with familiar structure but novel object combinations and longer-horizon tasks with novel structure (e.g., two-step pick-and-place).

\textbf{\abbr{}-Category} consists of tasks involving object categories never seen during training with 20 single-arm and 17 bimanual tasks.
Policies must rely on broad vision-language understanding beyond the finetuning dataset.

%% file: 5-experiments.tex
\section{Experiments}
\label{sec:experiments}

We organize our experiments around the following questions:
(1) How do state-of-the-art RFMs and baseline policies perform under controlled distribution shifts?
(2) To what extent is poor language following responsible for policy failures?
(3) How does naïvely scaling data quantity and diversity influence policy performance?
(4) Are the results from \abbr{} well-correlated with results from real-world policy evaluations?

We evaluate seven baselines on \abbr{}-Bench spanning three families: (i) a tabula rasa \textbf{Multitask Diffusion Policy (DP)}~\cite{diffusionpolicy,dasari2024ditpi}; (ii) VLM finetuning baselines that adapt a PaliGemma-3B~\cite{beyer2024paligemmaversatile3bvlm} backbone with either action binning (\textbf{PG-Bin}) \cite{rt-1,rt-2,kim2024openvla} or a flow-matching action head (\textbf{PG-FM}, equivalent to $\pi_0$~\cite{black2024pi0} without robot pretraining); and (iii) pretrained VLAs, namely \textbf{GR00T-N1.6}~\cite{nvidia2025gr00tn1openfoundation}, $\boldsymbol{\pi}\mathbf{_0}$~\cite{black2024pi0}, $\boldsymbol{\pi}\mathbf{_0}$-FAST~\cite{pertsch2025fastefficientactiontokenization}, and $\boldsymbol{\pi}\mathbf{_{0.5}}$~\cite{intelligence2025pi05}. Full architectural and training details are in \appref{sec:supp_training_details}.
We finetune each baseline on the \abbr{}-70 and BiMESA-57 datasets.

\subsection{Baseline Results}

\input{tables/main_results}

We present success rates for MESA and BiMESA in \autoref{tab:main-results}, with confidence intervals and paired significance tests in \appref{sec:uncertainty}. In the single-arm setting, a clear hierarchy emerges: $\pi_{0.5}$ achieves the strongest average performance, followed by $\pi_0$-FAST, $\pi_0$, PG-FM and GR00T-N1.6. The BiMESA results are more nuanced, with GR00T-N1.6 achieving more competitive performance and $\pi_0$ underperforming PG-FM despite its robot pretraining. 
DP's poor performance, especially in the single-arm setting, reflects poor multi-task scaling properties (\appref{sec:dp-study}).

\textbf{Robot pretraining does not uniformly transfer.}
The comparison between $\pi_0$ and PG-FM is particularly informative: they share a backbone and architecture, differing primarily in $\pi_0$'s pretraining on a large robot dataset that includes both single-arm Franka and bimanual YAM-like embodiments \cite{pertsch2025fastefficientactiontokenization}.
In the single-arm setting, robot pretraining leads to a consistent improvement, but in the bimanual setting PG-FM slightly outperforms $\pi_0$ on average.
This surprising reversal suggests that embodiment similarity alone does not guarantee transfer from pretraining.

\textbf{Action representation matters.}
PG-Bin's performance is worse than PG-FM on BiMESA despite sharing its VLM backbone.
This indicates that directly discretizing continuous action dimensions into language tokens is a poor fit for the fine-grained control required by BiMESA. 

\textbf{\abbr{}-Category represents an open frontier.}
Performance drops sharply for all finetuned models on novel object categories.
GR00T-N1.6 outperforms $\pi_0$ and $\pi_0$-FAST here despite weaker single-arm performance, possibly because its frozen Cosmos backbone provides good priors for novel objects.
In contrast, the zero-shot GPT-6.1-Sol agent (\appref{sec:supp_llm_agent}) is reasonable at pick-and-place (64.1\% on BiMESA-ID) but weak on handoffs (26.5\%) and articulated manipulation (13.1\%), yet does not degrade on unseen instances or categories (43.9\% and 52.4\% vs.\ 41.2\%).
This suggests that open-vocabulary grounding in current open VLAs has substantial room to improve.

\textbf{\abbr{}-Bench provides strong signal.}
The strongest model achieves 52.0\% on MESA and 48.7\% on BiMESA, indicating substantial headroom.
At the same time, modern VLA baselines make meaningful progress across all settings, making \abbr{}-Bench a high-signal benchmark for future work.

\subsection{Controlled Distribution Shifts}
Since \abbr{}-Spatial and \abbr{}-Instance each contain tasks with in-distribution counterparts in \abbr{}-ID, we can directly compare success rates with and without distribution shift (\autoref{fig:comparisons}).

Across both spatial and instance-level shifts, performance generally drops relative to matched in-distribution tasks, indicating that \abbr{} evaluates more than task memorization.
For spatial shifts, GR00T-N1.6 and $\pi_0$ show large degradations, while $\pi_{0.5}$, FAST and PG-FM are comparatively robust.
This pattern is consistent with the hypothesis that preserving the base VLM's spatial priors, either through knowledge insulation or direct finetuning from the VLM without broad robot pretraining, can improve transfer to unseen spatial configurations.
With unseen object instances, degradation is more uniform, indicating that instance generalization remains a broadly challenging axis.
These comparisons show how \abbr{} can isolate the effect of specific distribution shifts: rather than only reporting aggregate success, it reveals \textit{which} models generalize and \textit{when} they do so.

\input{figures/comparisons/comparisons}

\subsection{Probing Language Following}

\input{tables/language_following}

Reliably following language instructions is a core objective for RFMs, but existing benchmarks offer no way to study language following beyond using success rate as a proxy.
We present language following rate (LFR) (\autoref{sec:metrics}) in \autoref{tab:language-following}.

\textbf{Strong VLM backbones are important for language following.}
On in-distribution tasks (\abbr{}-ID and BiMESA-ID), models with VLM backbones exhibit strong LFR: PG-FM, $\pi_0$, and $\pi_0$-FAST all exceed 70\%, and $\pi_{0.5}$ reaches 92.9\% for BiMESA-ID.
When these models fail, they typically fail to complete any task rather than completing the wrong one.
DP, which has weaker language conditioning, achieves a very low language following rate.

Furthermore, we find that \textbf{pretraining recipe has an outsize effect on LFR,} and this is especially obvious with unseen object categories, which present the greatest language following challenge. GR00T-N1.6, which uses a frozen (and thus better-preserved) VLM backbone, consistently achieves strong LFR here, and is strongest by far among finetuned policies in BiMESA-Category. This suggests that preserving open-vocabulary visual-language knowledge can be important when policies must ground instructions involving object categories not seen during robot training.

Our language following metric provides a diagnostic that success rate alone cannot: it distinguishes successful execution of a distractor task from completion of the intended task, revealing grounding failures despite competent execution. Episodes in which neither task is completed remain ambiguous, since they may involve grounding errors, execution failures, or both.

\subsection{Scaling Data Quantity and Diversity}
\label{sec:scaling-diversity}

\input{figures/data-quantity/quantity}
\input{tables/diversity}

We study the effect of increasing demonstration quantity for $\pi_0$ and $\pi_{0.5}$ in the single-arm setting (\autoref{fig:quantity}) with a fixed compute budget.
A striking result is that \emph{$\pi_{0.5}$ trained with 10 demonstrations per task consistently outperforms $\pi_0$ with 100}.
In this regime, pretraining quality dominates data quantity, suggesting that investing in better pretraining or representation preservation yields greater returns than collecting larger fine-tuning datasets.

We investigate whether co-training on diverse auxiliary tasks improves single-arm generalization under a fixed compute budget by adding 10 demonstrations from each of 736 auxiliary tasks. These tasks reuse the \abbr{}-70 training verbs with different objects and do not overlap the held-out evaluation settings. We observe no consistent gains across models and suites (\autoref{tab:diversity}). One exception is $\pi_0$ on the 10-task Spatial suite, where success increases from 42.2\% to 54.4\%; however, this isolated result does not establish a general benefit for spatial generalization. On the remaining axes, $\pi_0$ shows no statistically significant improvements, and Category performance decreases from 11.6\% to 8.7\%. For $\pi_{0.5}$, no statistically significant changes are observed across the five suites. Overall, increasing auxiliary task diversity alone does not yield consistent generalization gains in this evaluation setting.

\subsection{Studying Cross-Arm Transfer}

\input{figures/spatial-comparison/spatial-comparison}

Because BiMESA-Spatial holds out the arm assignment rather than relocating objects corresponding to a single end-effector, it probes whether a policy has acquired \textit{arm-symmetric skills} or has instead memorized per-arm trajectories, a failure mode with no analogue in the single-arm setting. Each task has an in-distribution counterpart in BiMESA-57, enabling a direct trained-vs.-arm-switched comparison. We find that for tasks involving one arm (all task-relevant objects are on one side) or handoff tasks, policies are relatively capable of switching arms or handoff directions. However, they struggle to transfer articulated manipulation skills across arms.
This suggests that current policies can often reuse reaching and object-transfer behaviors across arms, but struggle when cross-arm transfer requires precise contacts with articulated objects.

\subsection{Comparison with Real-World Performance}

\input{tables/roboarena_comp}

While simulated benchmarks primarily provide controlled and reproducible diagnostics, their results are more useful when they align with trends observed in real-world evaluations.
Inspired by \citet{jain2025polarisscalablerealtosimevaluations}, in \autoref{tab:roboarena-mesa-results} we compare \abbr{} success rates to Elo scores on RoboArena \cite{atreya2025roboarena}, a crowdsourced evaluation platform where higher Elo scores reflect aggregate pairwise blind human preferences.

The rankings are broadly consistent: $\pi_{0.5}$ and $\pi_0$-FAST are the strongest methods across evaluations, while PG-FM and $\pi_0$ occupy the middle of the ranking.
The main disagreement is the ordering of PG-FM and $\pi_0$, which differs between RoboArena and the single-arm \abbr{} suite.
Overall, the agreement between simulated benchmark rankings and RoboArena scores provides preliminary evidence that \abbr{} captures method-level differences relevant beyond simulation. Since RoboArena evaluates policies trained on real data while \abbr{} evaluates the same architectures finetuned on \abbr{} data, we also evaluate the real-trained DROID checkpoints zero-shot in \abbr{}; these runs had success rates too low for reliable ranking (\appref{sec:supp_zero_shot}).

%% file: tables/main_results.tex
\begin{table}[t]
\centering
\captionsetup{position=below,skip=7pt}
\small
\setlength{\tabcolsep}{5pt}
\begin{tabular}{lcccccccc}
\toprule
\rowcolor[HTML]{FFDEB4}
Setting & DP & PG-Bin & PG-FM & GR00T-N1.6 & $\pi_0$ & FAST & $\pi_{0.5}$ & GPT-6.1-Sol$^\dagger$ \\
\midrule
MESA-ID & 1.6 & -- & 30.7 & 25.0 & 43.1 & 46.5 & \textbf{62.8} & -- \\
MESA-Spatial & 1.0 & -- & 35.0 & 23.2 & 42.2 & 53.6 & \textbf{68.6} & -- \\
MESA-Instance & 1.9 & -- & 22.7 & 22.7 & 34.0 & 39.1 & \textbf{55.4} & -- \\
MESA-Composite & 0.3 & -- & 17.7 & 15.0 & 27.5 & 28.9 & \textbf{46.7} & -- \\
MESA-Category & 0.7 & -- & 7.7 & 14.1 & 11.6 & 8.4 & \textbf{26.4} & -- \\
\midrule
MESA Avg. & 1.1 & -- & 22.8 & 20.0 & 31.7 & 35.3 & \textbf{52.0} & -- \\
\specialrule{\heavyrulewidth}{0.3ex}{0.8ex}
BiMESA-ID & 13.0 & 26.7 & 43.8 & 47.0 & 45.9 & 53.7 & \textbf{68.7} & 41.2 \\
BiMESA-Spatial & 18.8 & 30.3 & 50.8 & 46.2 & 49.2 & 58.1 & \textbf{70.5} & 41.8 \\
BiMESA-Instance & 9.4 & 13.9 & 28.6 & 36.7 & 23.9 & 36.7 & \textbf{48.1} & 43.9 \\
BiMESA-Composite & 5.7 & 14.4 & 17.9 & 22.1 & 18.4 & 28.8 & \textbf{37.5} & 34.7 \\
BiMESA-Category & 5.8 & 6.0 & 10.9 & 26.1 & 10.6 & 19.8 & 18.6 & \textbf{52.4} \\
\midrule
BiMESA Avg. & 10.6 & 18.3 & 30.4 & 35.6 & 29.6 & 39.4 & \textbf{48.7} & 42.8 \\
\bottomrule
\end{tabular}

\caption{\textbf{Success rates (\%)} on the single-arm and bimanual suites.
$\pi_{0.5}$ achieves the strongest average performance on both embodiments.
FAST also performs well, while GR00T-N1.6 is particularly competitive on BiMESA despite weaker single-arm performance.
DP performs poorly, and PG-Bin underperforms PG-FM.
$^\dagger$Zero-shot LLM agent with an end-effector tool-calling harness (\appref{sec:supp_llm_agent}); 10 rollouts per task instead of 50.}
\label{tab:main-results}
\end{table}

%% file: figures/comparisons/comparisons.tex
\begin{figure*}[t]
    \centering
    \includegraphics[width=0.47\linewidth]{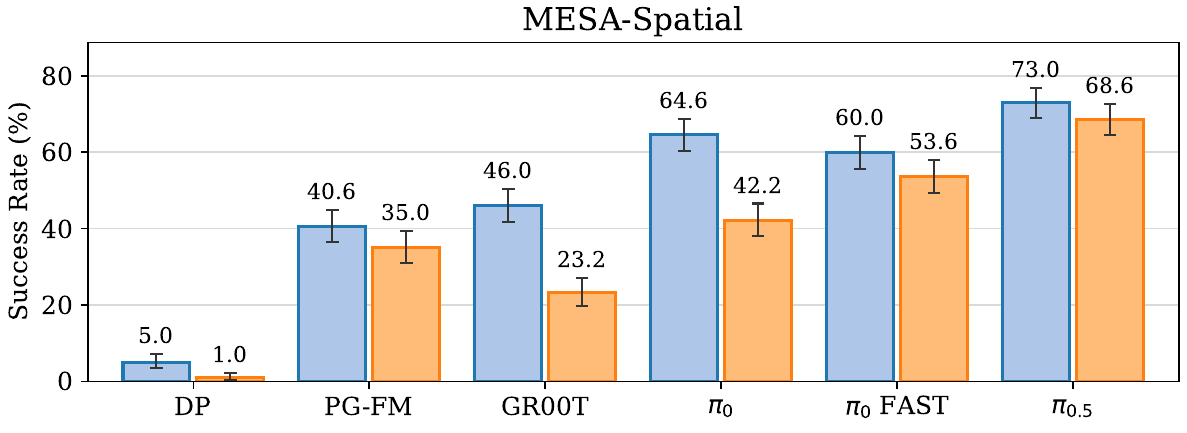}
    \includegraphics[width=0.47\linewidth]{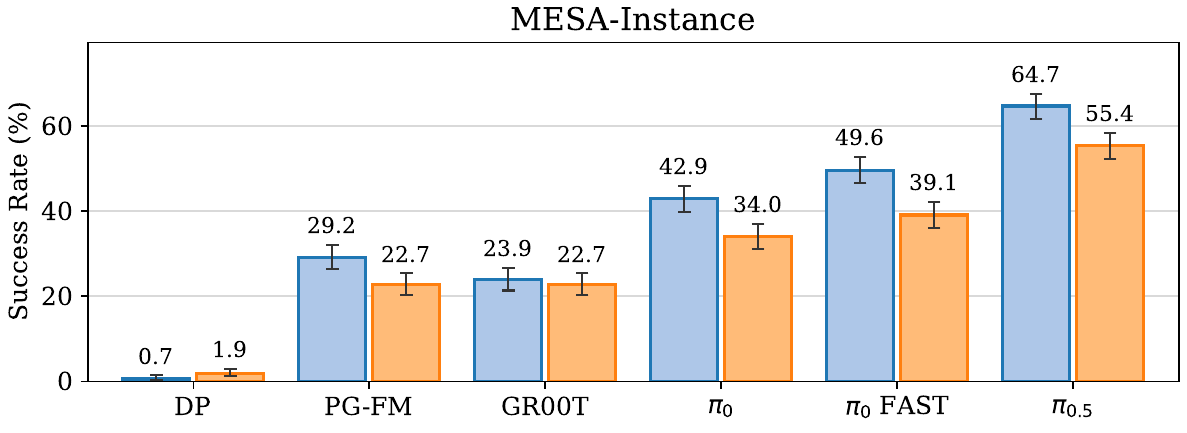}
    \includegraphics[width=0.4\linewidth]{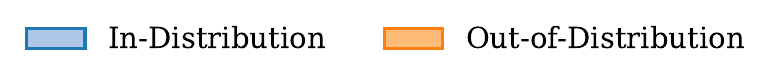}
    \caption{\textbf{Controlled spatial and instance shifts reduce performance.}
For tasks with matched in-distribution counterparts, policies generally perform worse under held-out spatial configurations and object instances, showing that MESA isolates failures under these distribution shifts. Error bars are 95\% Wilson intervals.}
    \label{fig:comparisons}
\end{figure*}

%% file: tables/language_following.tex
\begin{table}[t]
\centering
\captionsetup{position=below,skip=7pt}
\small
\setlength{\tabcolsep}{5pt}
\begin{tabular}{lcccccccc}
\toprule
\rowcolor[HTML]{FFDEB4}
Setting & DP & PG-Bin & PG-FM & GR00T-N1.6 & $\pi_0$ & FAST & $\pi_{0.5}$ & GPT-6.1-Sol \\
\midrule
MESA-ID & 18.9 & -- & 80.0 & 71.4 & 79.7 & 83.7 & \textbf{87.6} & -- \\
MESA-Spatial & 11.4 & -- & 77.4 & 53.7 & 73.0 & 78.6 & \textbf{84.5} & -- \\
MESA-Instance & 23.5 & -- & 66.6 & 60.5 & 66.5 & 69.0 & \textbf{76.9} & -- \\
MESA-Composite & 6.4 & -- & 65.1 & 56.2 & 65.5 & 66.3 & \textbf{81.5} & -- \\
MESA-Category & 12.1 & -- & 32.6 & 42.5 & 33.1 & 28.0 & \textbf{44.1} & -- \\
\midrule
MESA Avg. & 14.4 & -- & 64.3 & 56.9 & 63.6 & 65.1 & \textbf{74.9} & -- \\

\specialrule{\heavyrulewidth}{0.3ex}{0.8ex}
BiMESA-ID & 34.9 & 78.7 & 84.6 & 87.7 & 73.4 & 86.9 & \textbf{92.9} & 73.2 \\
BiMESA-Spatial & 45.3 & 75.7 & 86.2 & 77.5 & 72.6 & 83.0 & \textbf{89.1} & 69.7 \\
BiMESA-Instance & 25.5 & 53.0 & 64.4 & 73.3 & 48.0 & 64.8 & \textbf{74.0} & 65.8 \\
BiMESA-Composite & 14.5 & 65.1 & 66.7 & 71.2 & 42.3 & 66.7 & \textbf{80.1} & 74.3 \\
BiMESA-Category & 15.7 & 27.4 & 30.1 & 55.6 & 21.2 & 35.1 & 28.7 & \textbf{79.5} \\
\midrule
BiMESA Avg. & 27.2 & 60.0 & 66.4 & \textbf{73.1} & 51.5 & 67.3 & 73.0 & 72.5 \\

\bottomrule
\end{tabular}
\caption{\textbf{Language following rate (\%, \autoref{sec:metrics})}. 
Models with VLM backbones exhibit strong language following, while models trained from scratch do worse. Lower performance in \abbr{}-Category reflects that with unfamiliar objects, our baselines are bottlenecked by their ability to ground language to novel visual referents rather than low-level execution. \\
$^\dagger$ See (\appref{sec:supp_llm_agent})}
\label{tab:language-following}
\end{table}

%% file: figures/data-quantity/quantity.tex
\begin{figure*}[t]
    \centering
    \includegraphics[width=0.18\linewidth]{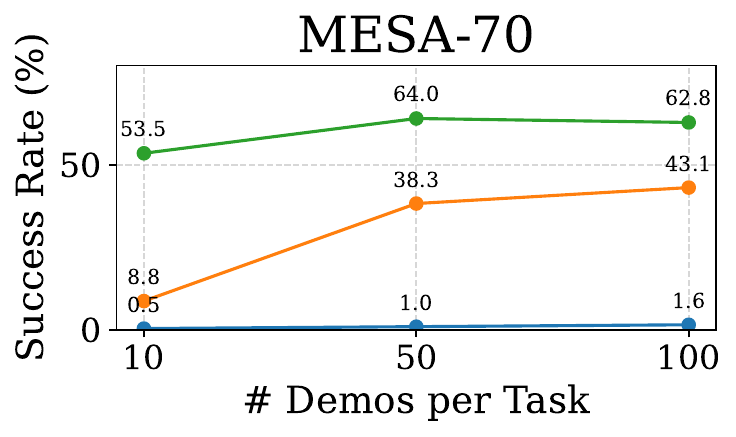}
    \includegraphics[width=0.18\linewidth]{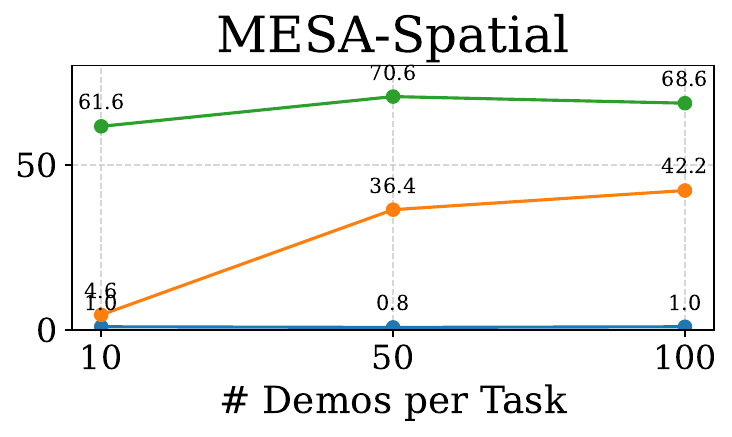}
    \includegraphics[width=0.18\linewidth]{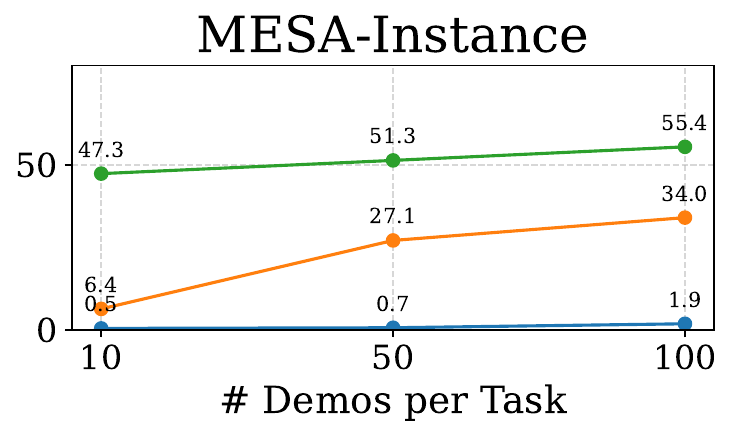}
    \includegraphics[width=0.18\linewidth]{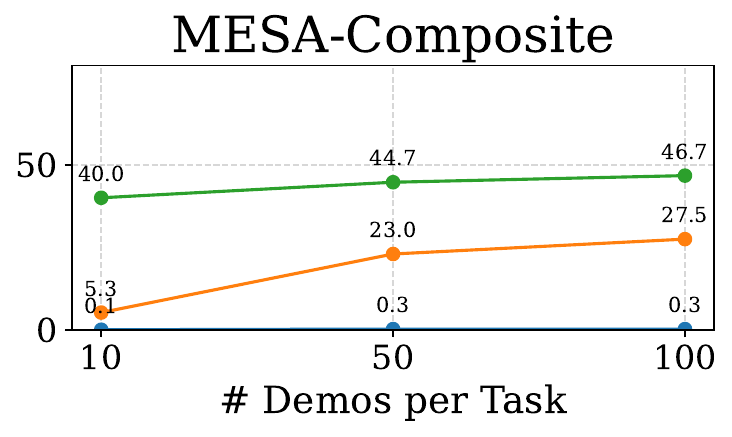}
    \includegraphics[width=0.18\linewidth]{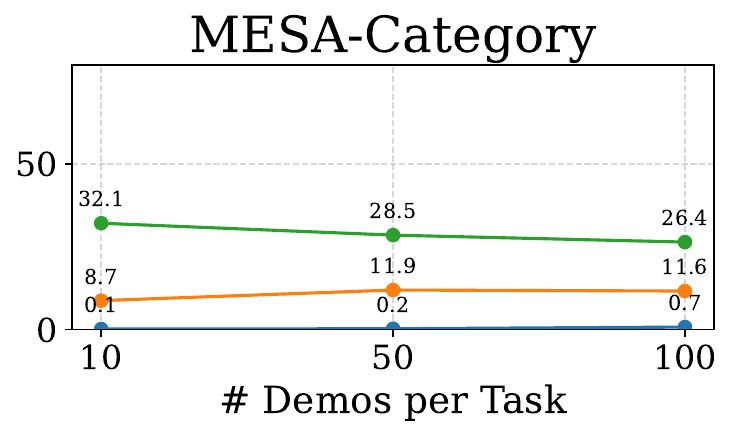}
    \includegraphics[width=0.3\linewidth]{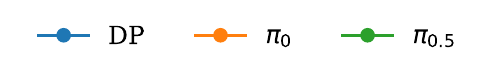}
    \caption{\textbf{Scaling data quantity.} Scaling data quantity almost monotonically improves performance. However, pretraining dominates as $\pi_{0.5}$ with 10 demonstrations consistently outperforms $\pi_0$ with 100.}
    \label{fig:quantity}
\end{figure*}

%% file: tables/diversity.tex
\begin{SCtable}[1][t]
\centering
\scriptsize
\begin{tabular}{lccccc}
\toprule
\rowcolor[HTML]{FFDEB4}
 & ID & Spatial & Instance & Composite & Category \\
\midrule
DP & 1.6 & 1.0 & 1.9 & 0.3 & 0.7 \\
DP+Aux & 2.3 & 2.4 & 2.2 & 0.6 & 0.3 \\
\midrule
$\pi_0$ & 43.1 & 42.2 & 34.0 & 27.5 & 11.6 \\
$\pi_0$+Aux & 43.4 & 54.4 & 33.2 & 28.8 & 8.7 \\
\midrule
$\pi_{0.5}$ & \textbf{62.8} & 68.6 & \textbf{55.4} & \textbf{46.7} & \textbf{26.4} \\
$\pi_{0.5}$+Aux & 59.0 & \textbf{69.2} & 53.6 & 45.6 & 23.0 \\
\bottomrule
\end{tabular}
\caption{\textbf{Scaling task diversity via auxiliary co-training yields negligible gains.} We double the training set size by adding auxiliary demonstrations from \abbr{}-Aux and train for the same number of steps. We find that the co-trained policies show little if any improvement, and hypothesize that naively scaling co-training task diversity does not automatically improve generalization.}
\label{tab:diversity}
\end{SCtable}

%% file: figures/spatial-comparison/spatial-comparison.tex
\begin{figure*}[t]
    \centering
    \includegraphics[width=0.32\linewidth]{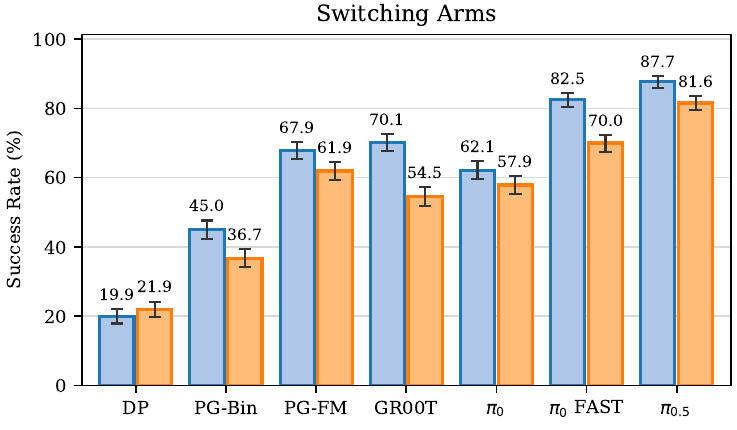}
    \includegraphics[width=0.32\linewidth]{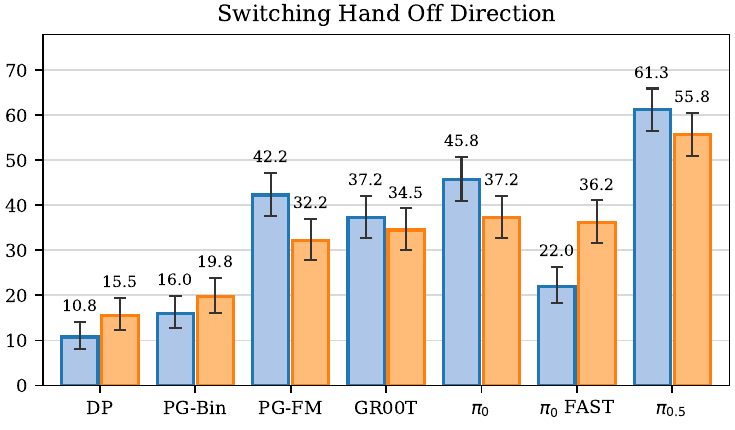}
    \includegraphics[width=0.32\linewidth]{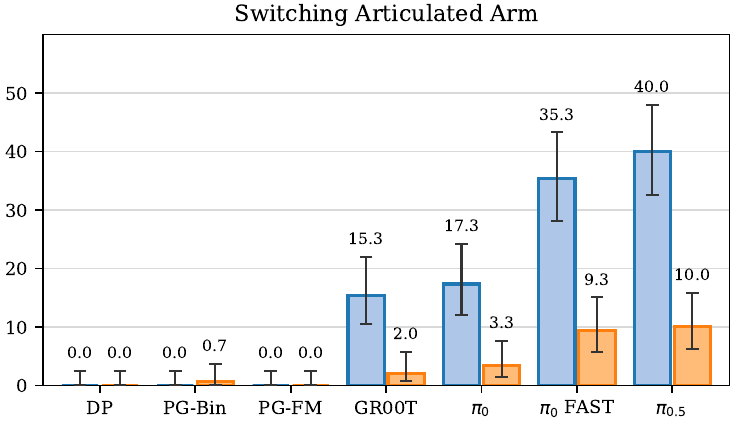}
    \includegraphics[width=0.4\linewidth]{figures/comparisons/spatial_legend.pdf}

    \caption{\textbf{BiMESA reveals that cross-arm transfer is hardest for articulated manipulation.}
    BiMESA-Spatial holds out arm assignments, allowing us to compare in-distribution tasks with matched arm-switched variants.
    Policies show modest degradation when switching single-arm tasks (left) or reversing handoff direction (middle), but much larger drops when transferring articulated manipulation skills across arms (right).
    DP, PG-Bin and $\pi_0$-FAST are exceptions and do not degrade on reversed handoffs. Error bars are 95\% Wilson intervals.}
    \label{fig:spatial-comparisons}
\end{figure*}

%% file: tables/roboarena_comp.tex
\begin{SCtable}[1][t]
\centering
\scriptsize
\begin{tabular}{lccccc}
\toprule
\rowcolor[HTML]{FFDEB4}
 & PG-Bin & PG-FM & $\pi_0$ & $\pi_0$-FAST & $\pi_{0.5}$ \\
\midrule
RoboArena 
& 771 
& 1527 
& 1449 
& 1564 
& \textbf{1594} \\
MESA      
& --
& 22.8 
& 31.7 
& 35.3 
& \textbf{52.0} \\
BiMESA
& 18.3
& 30.4
& 29.6
& 39.4
& \textbf{48.7} \\
\bottomrule
\end{tabular}
\caption{\textbf{Comparing RoboArena \cite{atreya2025roboarena} Elo with MESA success rates.}
MESA and BiMESA exhibit broadly similar stack rankings to RoboArena, suggesting that \abbr{} captures method-level differences relevant beyond simulation.
}
\label{tab:roboarena-mesa-results}
\end{SCtable}

%% file: 6-conclusion.tex
\section{Conclusion and Limitations}

We introduced \abbr{}, a language-conditioned robotic manipulation benchmarking framework consisting of \abbr{}-Gen, a scalable scene and data generation pipeline, and \abbr{}-Bench, a fixed, reproducible benchmark. 
Our experiments show that current RFMs leave substantial room for improvement in language following and out-of-distribution generalization, and that
MESA rankings are broadly consistent with real-world evaluation trends.
These results position \abbr{} as a scalable testbed for measuring progress in general-purpose robotic manipulation.

A key limitation of \abbr{} is its task distribution: most tasks involve pick-and-place or simple articulated manipulation, leaving more difficult settings such as dexterous, long-horizon, and mobile manipulation to future extensions. 
Additionally, our language-following metric's simplicity makes it robust at the cost of granularity, since it does not capture partial task completion or cases where a policy attempts the correct instruction but fails during execution.

\abbr{} prioritizes controlled distribution shifts and accessible experimentation. Establishing how its method rankings and axis-specific findings carry over to real robots remains an important direction for future validation. 
Rankings are conditional on the supplied training data and evaluation protocol; we have not tested whether alternative data sources, such as a matched human-teleoperation dataset, would preserve the ordering.

%% file: 7-appendix.tex
\clearpage
\newpage

\begin{appendix}

\renewcommand*{\sectionautorefname}{App.}
\renewcommand*{\subsectionautorefname}{App.}
\renewcommand*{\subsubsectionautorefname}{App.}

\input{7.1-supp-overview}

\clearpage
\input{7.2-additional-results}

\clearpage
\input{7.3-mesa-details}

\clearpage
\input{7.4-data-gen-details}

\clearpage
\input{7.5-bench-details}

\clearpage
\input{7.6-training-details}

\end{appendix}

%% file: 7.1-supp-overview.tex
\section{Supplement Overview and FAQs}

\begin{itemize}
    \item In \autoref{sec:supp_faq}, we answer some commonly asked questions.
    \item In \autoref{sec:supp_additional_analyses}, we provide additional experimental results.
    \item In \autoref{sec:supp_task_skeletons}, we provide detailed information about the \abbr{}-Gen framework.
    \item In \autoref{sec:supp_mesa_gen}, we provide additional information about our data generation pipeline.
    \item In \autoref{sec:supp_bench_lists}, we provide additional details about \abbr{}-Bench.
    \item In \autoref{sec:supp_training_details}, we provide detailed policy training information.
\end{itemize}

\subsection{Frequently Asked Questions (FAQs)}
\label{sec:supp_faq}

\medskip \noindent \textbf{Why should I use MESA?}
MESA enables quick iteration while giving strong signal about language steering and performance along our axes of generalization.
Furthermore, it is an easily extensible platform for designing new experimental settings.

\medskip \noindent \textbf{What explains diffusion policy's unexpectedly poor performance?}
We study this question in depth in \autoref{sec:dp-study}.
To summarize, we find that our implementation works well in LIBERO and that DP simply struggles to fit the diverse setting presented by the benchmark.

\medskip \noindent \textbf{What is this benchmark testing?}
We designed this benchmark to test language steering and policy performance under changes in spatial arrangement, object instance, object category and with unseen subtask compositions.
BiMESA additionally tests bimanual generalization through held-out arm assignments, reversed handoff directions, and cross-arm articulated manipulation.

\medskip \noindent \textbf{What does this benchmark not test for?}
This benchmark is not geared towards dexterous or precise tasks, open-ended language steerability, long-horizon tasks or mobile manipulation.

\medskip \noindent \textbf{What motivates the lack of scene and camera pose diversity?}
We purposely limit diversity along axes we don't test in order to minimize the amount of training required.
That being said, this is implemented and easy to add back if desired.

\medskip \noindent \textbf{What are the limits of MimicGen data?}
MimicGen data has several limitations, especially for our domain with such scene variety.
It is not well suited to deal with dense clutter, and even with our optimizations it sometimes outputs suboptimal trajectories.
We do not claim in this paper to present a general solution for robot data generation, or that our data is optimal.
Rather, we claim that MimicGen data is useful for inexpensively generating data for experiments, and that the stack ranking of policies trained on this data is broadly consistent with real-world rankings.

\medskip \noindent \textbf{What types of tasks are included in the benchmark?}
The benchmark mainly focuses on pick and place and articulated manipulation tasks.
This limited scope allows us to quickly generate data with MimicGen and train policies that achieve meaningful generalization under reasonable compute budgets.

%% file: 7.2-additional-results.tex
\section{Additional Experimental Results}
\label{sec:supp_additional_analyses}

\subsection{Why Diffusion Policy Scales Poorly in Multi-Task \abbr{}}
\label{sec:dp-study}
One surprising finding is that, although we expected the diffusion policy to underperform relative to its VLA counterparts, it performs substantially worse than anticipated on the single-arm \abbr{} benchmark, achieving an average success rate of 1.1\%.
We investigate this in \autoref{fig:lime-overfit}, where we train a DiT policy \cite{dasari2024ditpi} for a single task with increasing numbers of demonstrations and cotraining tasks.
We see that it achieves a reasonable success rate with 1000 demonstrations, but performance drops quickly as the number of demonstrations decreases.
Furthermore, we see that it especially struggles when switching from a single-task to multi-task setting, consistently achieving a 0\% success rate when cotraining with other tasks.
A likely explanation is that, compared to the baselines using a VLM backbone, the DiT policy has a much weaker vision-language conditioning pathway and cannot adequately fit the diverse data included in the \abbr{} dataset.

\input{figures/lime/overfit}

\medskip \noindent \textbf{Verifying with LIBERO}

To verify our DiT Policy implementation we train and evaluate on LIBERO-90 \cite{liu2024libero}.
We achieve a success rate of 90.7\%, which is comparable to results reported elsewhere \cite{mete2024quest, wilcox2025adapt3r} and provides evidence against a general implementation failure.
We hypothesize that \abbr{}'s substantial variation in object sets and placements across episodes increases the demands on vision-language grounding and contributes to this implementation's poor multi-task performance.

\subsection{Ablating Language Instructions}
\input{figures/rand-language/rand-lang}

A core claim from our work is that the diverse initial configurations create a consistent requirement for language steerability.
To verify this claim, we rerun single-arm evaluations for $\pi_{0.5}$, the strongest baseline, while scrambling the natural language instructions (precisely, for each evaluation we sample a random instruction from the benchmark), and present results in \autoref{fig:rand-lang}.
We see that this change leads to a substantial drop in success rate, with successes only occurring when the policy successfully guesses the correct instruction.
This verifies our claim that accurate language steering is essential to perform well in the \abbr{} benchmark.

\subsection{Ablating MimicGen Changes}

\input{figures/mg-ablation/mg-ablation}

In \appref{sec:supp_mesa_gen} we present several modifications to MimicGen to improve data generation in our setting with many tasks and objects.
These changes include subtask stitching and the trajectory starting point optimization procedure.
The optimization procedure has two main components: (i) we remove roll, pitch and optionally yaw from the source and target object poses and (ii) we optimize the starting point of the trajectory as described in \autoref{sec:supp_traj_opt}.
Subtask stitching has a clear benefit -- enabling data generation across over 800 tasks after only collecting data from 80 -- which is difficult to ablate.
We instead ablate the changes to the optimization procedure in \autoref{fig:mg-abl}, running the experiment in the single-arm setting.
As shown in \autoref{fig:abl-data-gen}, these changes both lead to an improvement in data generation rate, reducing compute requirements when generating large datasets.
In \autoref{fig:abl-succ}, we additionally observe that removing the optimization procedure leads to a substantial drop in performance.
This is likely because without optimizing the start of the trajectory, the generated data can include large, unnecessary interpolations to distant regions before moving towards the target object.
In effect, the data includes many suboptimal motions that can confuse the policy.
This interpretation is further supported by trajectory-length statistics: removing the optimization procedure increases the average trajectory length from 141.7 to 177.9.

\subsection{Bias in Generated Data Toward High Success-Rate Initial Conditions}
\label{sec:supp_bias_study}
MimicGen works by initializing the environment, attempting to solve the task using its trajectory transformations, and then reinitializing it once it exceeds some retry budget.
This process can bias the dataset towards initial configurations where MimicGen is likely to succeed, creating issues for the downstream policy.
We study this issue by computing normalized initial position frequency for the single-arm setting in \autoref{fig:init-dist}.
We observe minor bias in the dataset with more data in the lower-right quadrant and fewer in the upper-left quadrant.
However, within each tile, the deviation from an unbiased distribution is at most about 15\%, reflecting a relatively evenly distributed dataset.

\input{figures/init_dist/init-dist}

\subsection{Visualizing Confusion Matrices}

\input{figures/confusion_matrix/confusion}

In \autoref{fig:confusion} we visualize confusion matrices for DP and $\pi_{0.5}$ from the single-arm setting, which we can compute directly using the logged distractor task completion data.
One notable observation is that, despite being more performant overall, $\pi_{0.5}$ is confused more often.
This is likely because it is simply more capable of successfully grasping and placing objects; as a result, even if it occasionally takes an incorrect action, it is still likely to complete the distractor task and register as a confusion.
A second observation is that the confusion matrix for $\pi_{0.5}$ is much more ``spiky'', with a handful of frequently repeated mistakes, whereas diffusion policy's confusions are scattered more uniformly across the confusion matrix.
Importantly, the high-frequency confusions made by $\pi_{0.5}$ are intuitive.
For example, peaches and tomatoes look visually similar to apples and are therefore often confused with them.
This suggests that $\pi_{0.5}$ is generally making reasonable attempts to select the intended object, with errors driven by visual similarity.
In contrast, diffusion policy seems to select incorrect objects more indiscriminately, consistent with behavior that resembles grabbing random objects.

\subsection{Zero-Shot Evaluation of Real-Trained Policies}
\label{sec:supp_zero_shot}

We evaluate the DROID checkpoints of $\pi_0$, $\pi_0$-FAST, and $\pi_{0.5}$ in the single-arm \abbr{} environment without finetuning on \abbr{} data.
We match action and observation spaces, normalization, action horizon, and joint dynamics, and use system identification to tune the controller gain from $K_p=500$ to $234$; this adjustment yields no observed performance benefit.
The intermediate subtask completion rates are 0.60\% for $\pi_0$, 0.23\% for $\pi_0$-FAST, and 1.77\% for $\pi_{0.5}$.
These measurements concern intermediate subtask completion rather than the full-task success metric used in our main results.
The low absolute rates provide insufficient signal for reliable ranking conclusions, so this experiment does not establish alignment with real-world performance without \abbr{} finetuning.

\subsection{Zero-Shot Evaluation of an LLM Agent}
\label{sec:supp_llm_agent}

We evaluate GPT-6.1-Sol on BiMESA without BiMESA training data, using the RoboDojo L3 Inspect EEF harness~\cite{zhang2026astra}.
Each call, the model receives the instruction, the egocentric and both wrist images at $224\times224$, each arm's end-effector pose, joint angles, and measured jaw opening, and the remaining step budget; it receives no object poses, depth, or detections.
It responds with one tool call that sets end-effector and gripper targets for either or both arms, or gives up.
We replace the harness's CuRobo planner with damped least-squares IK, which does not check collisions; the resulting joint-position waypoints drive the same controller as the baselines.
We adapt the harness's embodiment notes to the YAM arms and drop RoboDojo's return-to-home requirement.
Otherwise, we use the published configuration (170 calls, medium reasoning effort) without per-task recipes, and the baselines' 500-step limit.
We evaluate rollouts 0--9 of each task (1,450 in total) to reduce API call costs.
Model latency is a median 13$\times$ the simulated robot time, so the agent does not run in real time.

\subsection{Per-Task Success Rates for MESA}
\label{sec:per-task-sr}
Here we list success rates across all tasks present in the benchmark.
We present success rates for \abbr{}-ID in \autoref{tab:mesa-70-per-task-sr}, \abbr{}-Spatial in \autoref{tab:mesa-spatial-per-task-sr}, \abbr{}-Instance in \autoref{tab:mesa-instance-per-task-sr}, \abbr{}-Composite in \autoref{tab:mesa-composite-per-task-sr}, and \abbr{}-Category in \autoref{tab:mesa-category-per-task-sr}.

\input{tables/mesa-70-sr-table}

\input{tables/mesa-spatial-sr-table}
\input{tables/mesa-instance-sr-table}
\input{tables/mesa-composite-sr-table}
\input{tables/mesa-object-sr-table}

\subsection{Per-Task Success Rates for BiMESA}

Several BiMESA tasks share a language instruction and differ only in arm assignment, so each row is annotated with its configuration: which arm performs a single-arm task, the pick and place sides for cross-arm pick-and-place, the handoff direction (when held out), or the acting arm and fixture side for articulated tasks.

\input{tables/bimesa_id_per_task_success}
\input{tables/bimesa_spatial_per_task_success}
\input{tables/bimesa_instance_per_task_success}
\input{tables/bimesa_composite_per_task_success}
\input{tables/bimesa_category_per_task_success}

%% file: figures/lime/overfit.tex
\begin{figure}[htb!]
    \centering
    \begin{subfigure}[t]{0.35\linewidth}
        \centering
        \includegraphics[width=\linewidth]{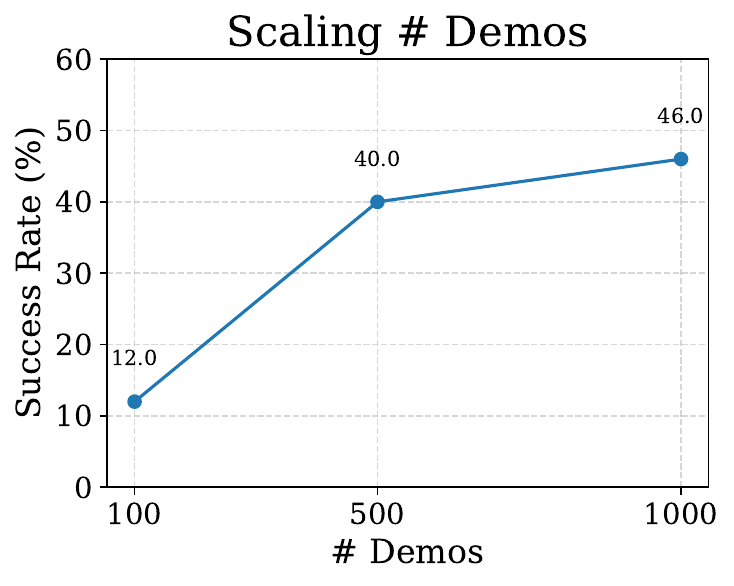}
        \caption{Single-task training}
        \label{fig:lime-overfit-a}
    \end{subfigure}
    \begin{subfigure}[t]{0.35\linewidth}
        \centering
        \includegraphics[width=\linewidth]{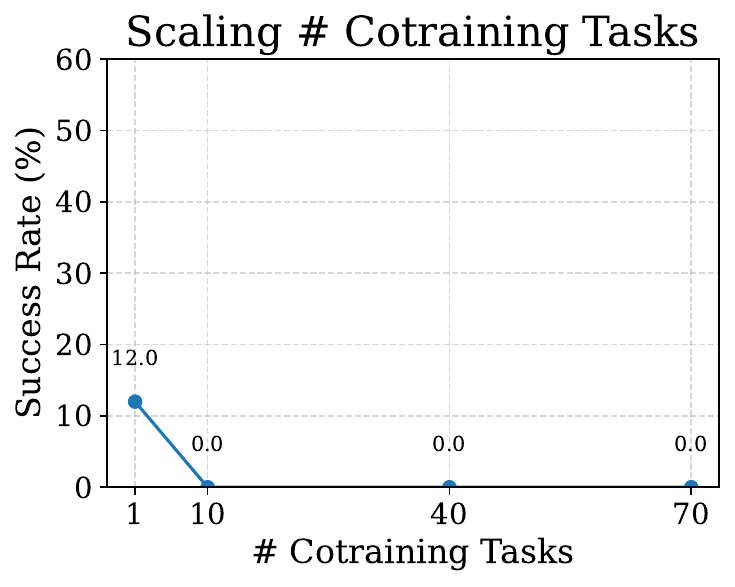}
        \caption{Co-training with 100 demonstrations}
        \label{fig:lime-overfit-b}
    \end{subfigure}
    \caption{We study the performance of diffusion policy on a single task (\texttt{``put the lime in the bowl''}) as we scale the number of demonstrations and cotraining tasks. (a) We see that with 1000 demonstrations it achieves reasonable performance, but with 100 demonstrations, the number present for each task in \abbr{}-70, its performance drops sharply. (b) With 100 demonstrations we see that as we increase the number of cotraining tasks, performance quickly drops to zero.}
    \label{fig:lime-overfit}
\end{figure}

%% file: figures/rand-language/rand-lang.tex
\begin{figure}[htb!]
    \centering
    \includegraphics[width=0.5\linewidth]{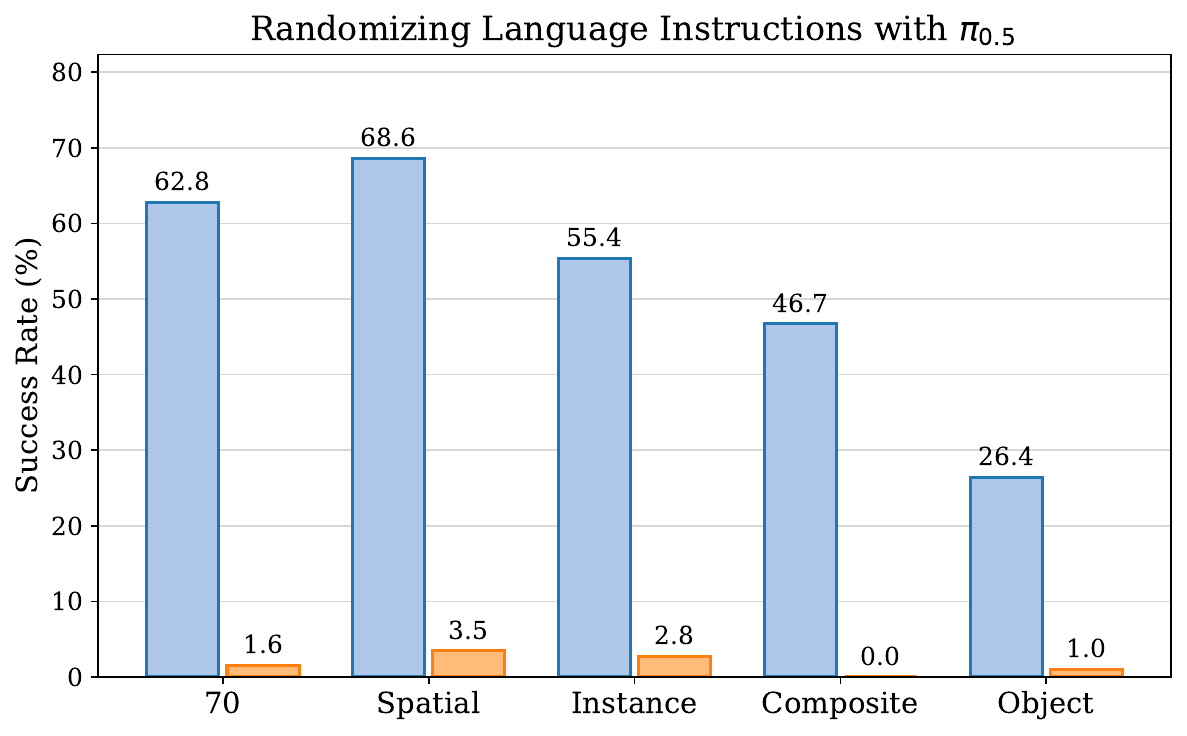}
    
    \includegraphics[width=0.3\linewidth]{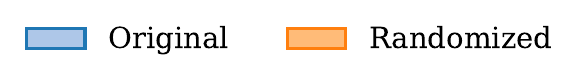}
    \caption{We evaluate $\pi_{0.5}$ with the original and randomized language instructions in the single-arm MESA setting. This leads to a substantial performance drop, indicating the necessity of language following for \abbr{}.}
    \label{fig:rand-lang}
\end{figure}

%% file: figures/mg-ablation/mg-ablation.tex
\begin{figure}[htb!]
    \centering
    \begin{subfigure}[t]{0.45\linewidth}
        \centering
        \includegraphics[width=\linewidth]{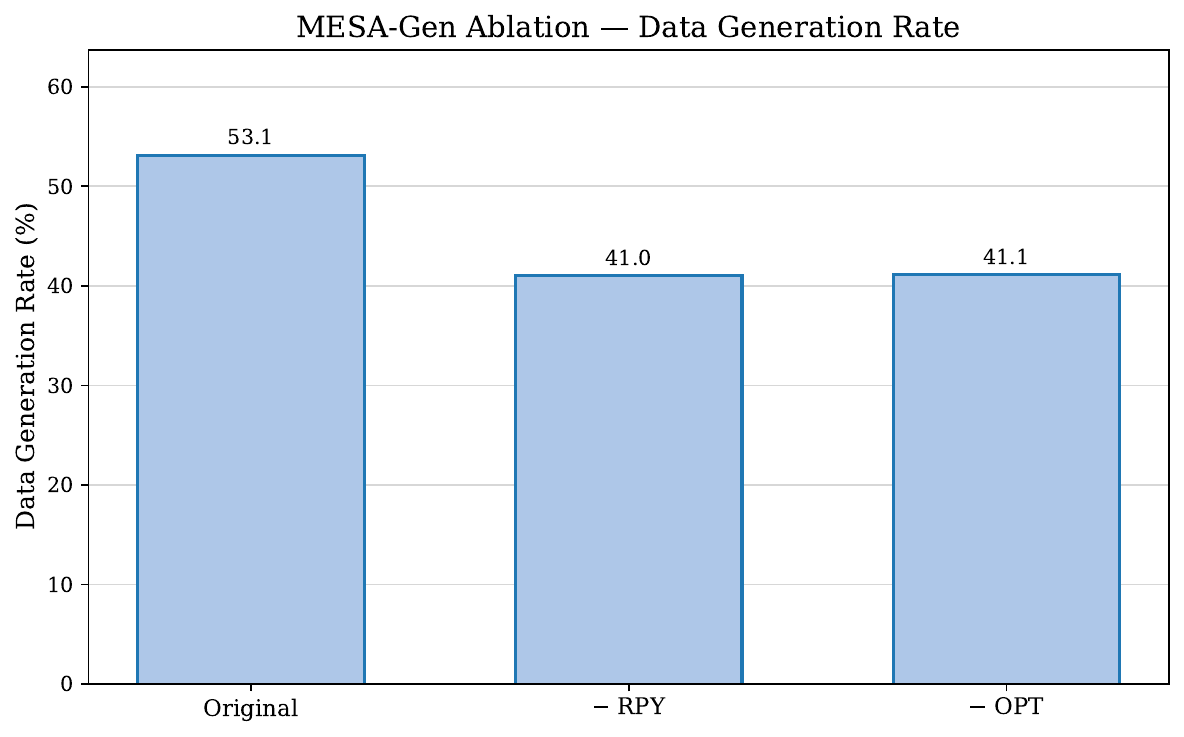}
        \caption{Data Generation Rate}
        \label{fig:abl-data-gen}
    \end{subfigure}
    \begin{subfigure}[t]{0.45\linewidth}
        \centering
        \includegraphics[width=\linewidth]{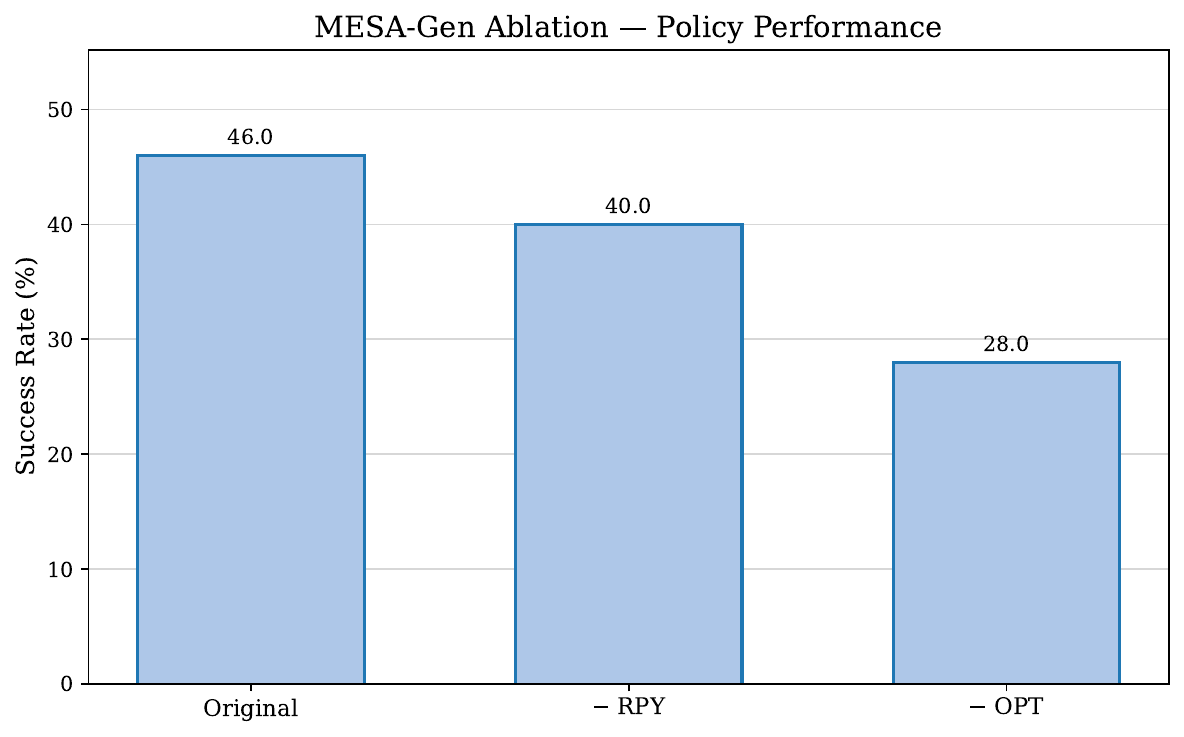}
        \caption{Trained Policy Performance}
        \label{fig:abl-succ}
    \end{subfigure}
    \caption{We study the effects of our MimicGen changes when generating and training on 1000 demos for the \texttt{``put the lime in the bowl''} task. We ablate the omission of roll, pitch and yaw ($-$ RPY) and the starting point optimization procedure ($-$ Opt). In (a) we study data generation rate, or the success rate of MimicGen at data generation time, and see that our changes lead to a consistent improvement. In (b) we study downstream policy performance and see that especially the starting point optimization leads to a substantial policy improvement.}
    \label{fig:mg-abl}
\end{figure}

%% file: figures/init_dist/init-dist.tex
\begin{figure}[htb!]
    \centering
    \begin{subfigure}[t]{0.45\linewidth}
        \centering
        \includegraphics[width=\linewidth]{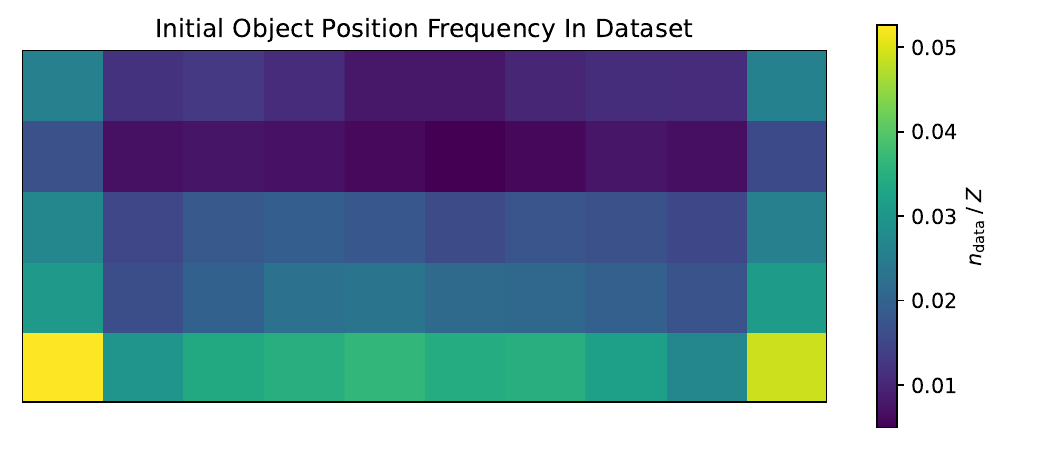}
        \caption{Frequency in the Dataset}
        \label{fig:init-dist-data}
    \end{subfigure}
    \begin{subfigure}[t]{0.45\linewidth}
        \centering
        \includegraphics[width=\linewidth]{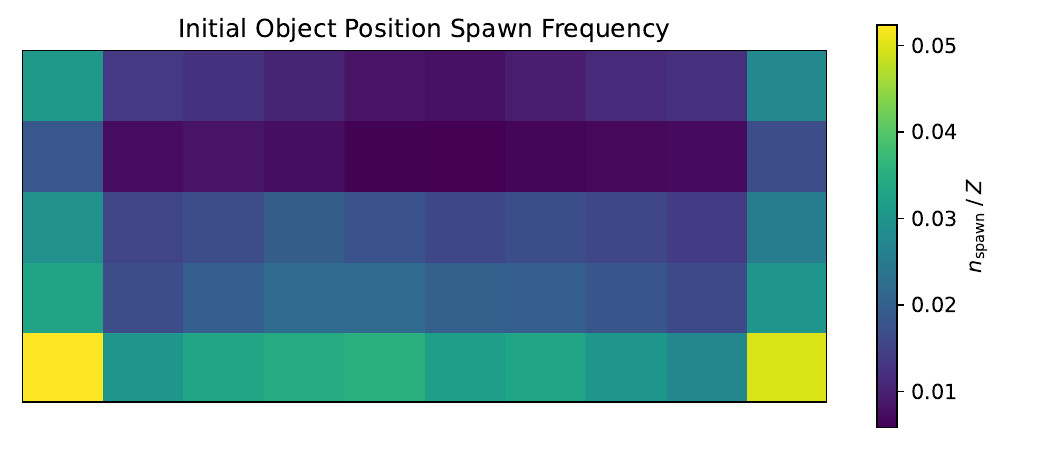}
        \caption{Spawn Frequency}
        \label{fig:init-dist-spawn}
    \end{subfigure}
    \begin{subfigure}[t]{0.45\linewidth}
        \centering
        \includegraphics[width=\linewidth]{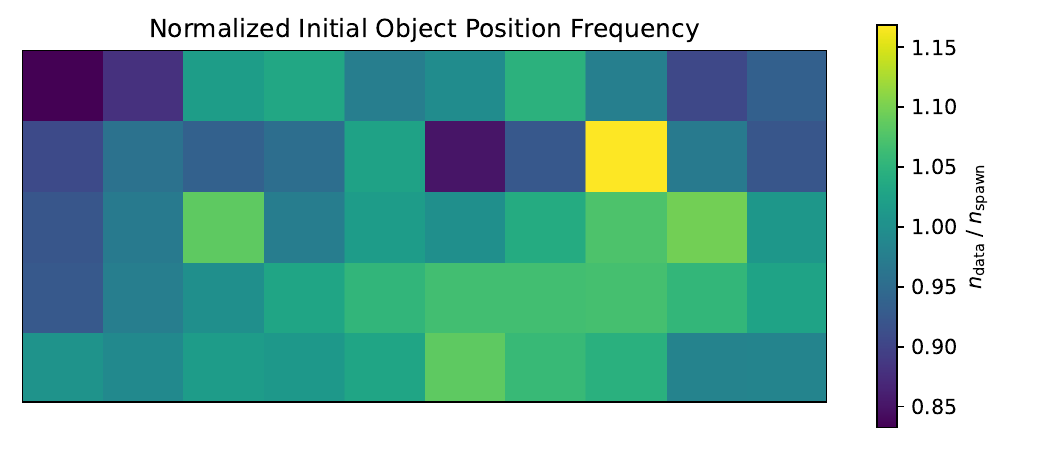}
        \caption{Normalized Frequency}
        \label{fig:init-dist-norm}
    \end{subfigure}
    \caption{We study whether MimicGen led to a biased dataset by visualizing the frequency with which different target object initial positions appear in our dataset. In (a), we show this value, and in (b) we show the frequency with which initial states spawn before MimicGen. Since these distributions are highly correlated, we show the normalized distribution in (c). We see that there is some bias towards the lower right corner of the region and away from the upper left corner. In (c), values greater than one reflect higher frequency in the dataset than an unbiased dataset would have and values less than one are the opposite.}
    \label{fig:init-dist}
\end{figure}

%% file: figures/confusion_matrix/confusion.tex
\begin{figure}[htb!]
    \centering
    \includegraphics[width=0.45\linewidth]{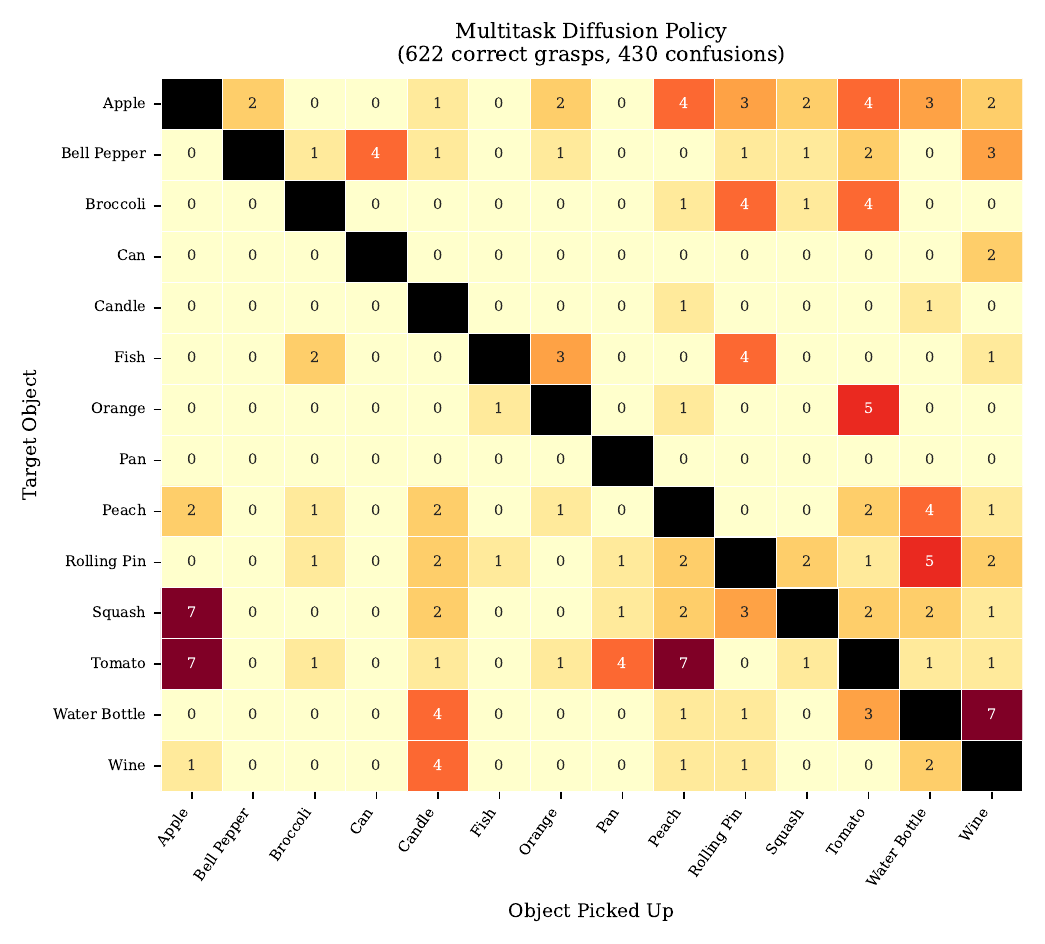}
    \includegraphics[width=0.45\linewidth]{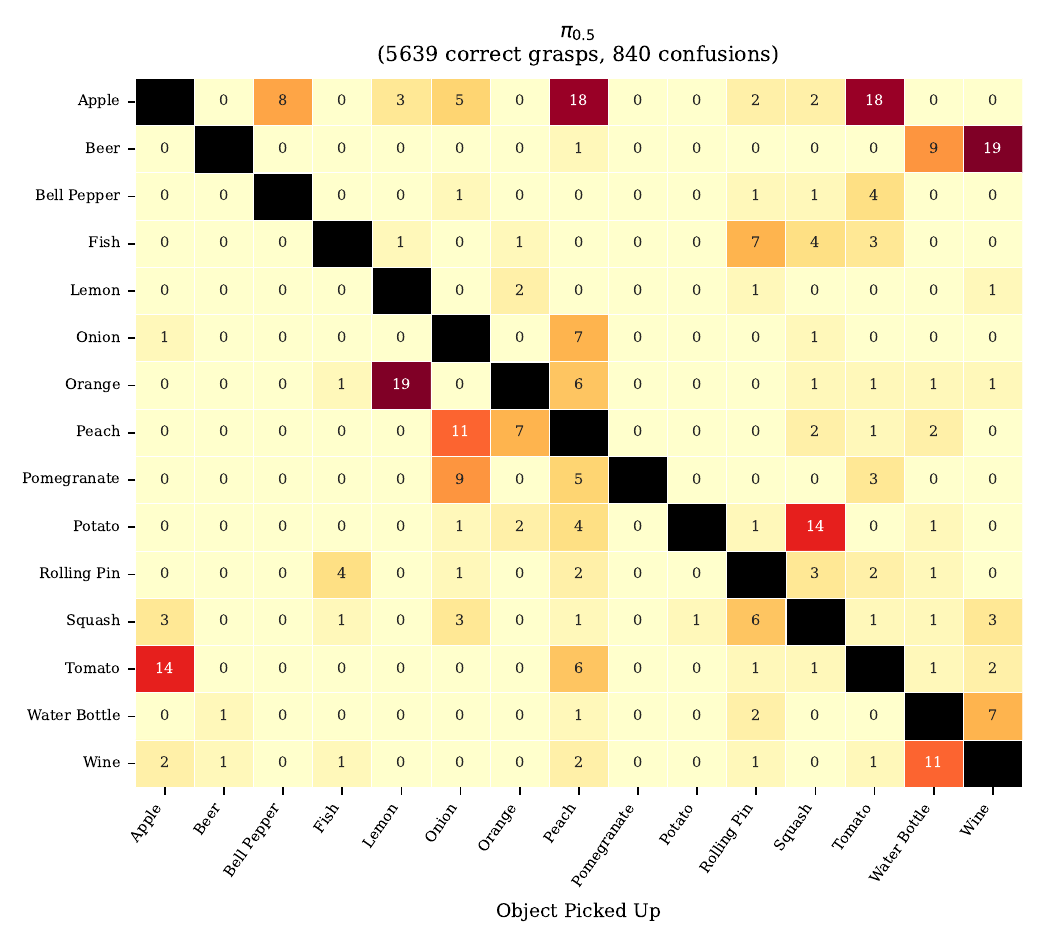}
    \caption{We compare confusion matrices for diffusion policy and $\pi_{0.5}$. Entry $i,j$ in these plots shows the frequency with which a given policy completed a distractor task by grasping $j$ when it was supposed to grasp $i$ (rows are target objects; columns are the objects picked up). For example, both confusion matrices show a lot of confusion for $\mathrm{tomato},\mathrm{apple}$ because the policy frequently grasped an apple when it was supposed to grasp a tomato. We condensed them to a reasonable size by selecting the top 15 most confusing entries in each matrix and removing rows and columns corresponding to objects not involved in those entries. For example, no object was commonly confused for mushrooms, so their rows and columns were removed.}
    \label{fig:confusion}
\end{figure}

%% file: tables/mesa-70-sr-table.tex
\begin{longtable}{lcccccc}
\caption{Success rates for all tasks in the MESA-ID task set.}\label{tab:mesa-70-per-task-sr} \\
\toprule
\rowcolor[HTML]{FFDEB4} \textbf{MESA-ID} & \textbf{DP} & \textbf{PG-FM} & \textbf{GR00T-N1.6} & $\boldsymbol{\pi}\mathbf{_0}$ & \textbf{FAST} & $\boldsymbol{\pi}\mathbf{_{0.5}}$ \\
\midrule
\endfirsthead
\toprule
\rowcolor[HTML]{FFDEB4} \textbf{MESA-ID} & \textbf{DP} & \textbf{PG-FM} & \textbf{GR00T-N1.6} & $\boldsymbol{\pi}\mathbf{_0}$ & \textbf{FAST} & $\boldsymbol{\pi}\mathbf{_{0.5}}$ \\
\midrule
\endhead
\rowcolor[HTML]{FFFFFF}
\parbox[t]{0.35\linewidth}{put the apple on the tray\smallskip} & 2.0 & 54.0 & 62.0 & \textbf{66.0} & 64.0 & \textbf{66.0} \\
\rowcolor[HTML]{F2F2F2}
\parbox[t]{0.35\linewidth}{put the avocado in the basket\smallskip} & 0.0 & 58.0 & 18.0 & 50.0 & 70.0 & \textbf{88.0} \\
\rowcolor[HTML]{FFFFFF}
\parbox[t]{0.35\linewidth}{put the bagel on the cutting board\smallskip} & 8.0 & 26.0 & 36.0 & 48.0 & 36.0 & \textbf{86.0} \\
\rowcolor[HTML]{F2F2F2}
\parbox[t]{0.35\linewidth}{put the bar in the bottom drawer of the cabinet and close it\smallskip} & 0.0 & 20.0 & 12.0 & 22.0 & 30.0 & \textbf{46.0} \\
\rowcolor[HTML]{FFFFFF}
\parbox[t]{0.35\linewidth}{put the bar soap in the top drawer of the cabinet and close it\smallskip} & 0.0 & 24.0 & 20.0 & 24.0 & 58.0 & \textbf{60.0} \\
\rowcolor[HTML]{F2F2F2}
\parbox[t]{0.35\linewidth}{put the beer in the slide cabinet and close it\smallskip} & 0.0 & 6.0 & 6.0 & 10.0 & 10.0 & \textbf{34.0} \\
\rowcolor[HTML]{FFFFFF}
\parbox[t]{0.35\linewidth}{put the bell pepper in the box and close it\smallskip} & 0.0 & 10.0 & 10.0 & 14.0 & 26.0 & \textbf{52.0} \\
\rowcolor[HTML]{F2F2F2}
\parbox[t]{0.35\linewidth}{put the bottled water on the tray\smallskip} & 4.0 & 36.0 & 30.0 & 42.0 & 66.0 & \textbf{84.0} \\
\rowcolor[HTML]{FFFFFF}
\parbox[t]{0.35\linewidth}{put the bowl in the microwave and close it\smallskip} & 0.0 & 22.0 & 10.0 & 34.0 & 12.0 & \textbf{40.0} \\
\rowcolor[HTML]{F2F2F2}
\parbox[t]{0.35\linewidth}{put the broccoli in the pan\smallskip} & 4.0 & 56.0 & 52.0 & 64.0 & 62.0 & \textbf{72.0} \\
\rowcolor[HTML]{FFFFFF}
\parbox[t]{0.35\linewidth}{put the candle on the tray\smallskip} & 6.0 & 48.0 & 30.0 & 52.0 & 48.0 & \textbf{62.0} \\
\rowcolor[HTML]{F2F2F2}
\parbox[t]{0.35\linewidth}{put the carrot in the bowl\smallskip} & 0.0 & 30.0 & 16.0 & 44.0 & 50.0 & \textbf{68.0} \\
\rowcolor[HTML]{FFFFFF}
\parbox[t]{0.35\linewidth}{put the cheese on the plate\smallskip} & 0.0 & 32.0 & 24.0 & 46.0 & 70.0 & \textbf{84.0} \\
\rowcolor[HTML]{F2F2F2}
\parbox[t]{0.35\linewidth}{put the corn on the cutting board\smallskip} & 6.0 & 56.0 & 30.0 & 72.0 & 76.0 & \textbf{88.0} \\
\rowcolor[HTML]{FFFFFF}
\parbox[t]{0.35\linewidth}{put the croissant in the slide cabinet and close it\smallskip} & 0.0 & 16.0 & 10.0 & 24.0 & 28.0 & \textbf{36.0} \\
\rowcolor[HTML]{F2F2F2}
\parbox[t]{0.35\linewidth}{put the cucumber in the bottom drawer of the cabinet and close it\smallskip} & 0.0 & 8.0 & 12.0 & 16.0 & 24.0 & \textbf{48.0} \\
\rowcolor[HTML]{FFFFFF}
\parbox[t]{0.35\linewidth}{put the cup in the right side of the dish drainer\smallskip} & 2.0 & 32.0 & 20.0 & 42.0 & 48.0 & \textbf{62.0} \\
\rowcolor[HTML]{F2F2F2}
\parbox[t]{0.35\linewidth}{put the egg in the bowl\smallskip} & 0.0 & 24.0 & 24.0 & 40.0 & 42.0 & \textbf{74.0} \\
\rowcolor[HTML]{FFFFFF}
\parbox[t]{0.35\linewidth}{put the eggplant on the cutting board\smallskip} & 0.0 & 38.0 & 28.0 & 44.0 & 68.0 & \textbf{78.0} \\
\rowcolor[HTML]{F2F2F2}
\parbox[t]{0.35\linewidth}{put the fish in the pan\smallskip} & 2.0 & 40.0 & 24.0 & 46.0 & 56.0 & \textbf{72.0} \\
\rowcolor[HTML]{FFFFFF}
\parbox[t]{0.35\linewidth}{put the fish on the plate\smallskip} & 0.0 & 40.0 & 14.0 & 48.0 & 52.0 & \textbf{78.0} \\
\rowcolor[HTML]{F2F2F2}
\parbox[t]{0.35\linewidth}{put the garlic in the middle drawer of the cabinet and close it\smallskip} & 0.0 & 56.0 & 54.0 & 52.0 & 50.0 & \textbf{70.0} \\
\rowcolor[HTML]{FFFFFF}
\parbox[t]{0.35\linewidth}{put the garlic in the pan\smallskip} & 2.0 & 68.0 & 32.0 & 76.0 & 58.0 & \textbf{78.0} \\
\rowcolor[HTML]{F2F2F2}
\parbox[t]{0.35\linewidth}{put the jam in the basket\smallskip} & 0.0 & 46.0 & 20.0 & 64.0 & 68.0 & \textbf{88.0} \\
\rowcolor[HTML]{FFFFFF}
\parbox[t]{0.35\linewidth}{put the jug in the basket\smallskip} & 0.0 & 10.0 & 20.0 & 10.0 & 32.0 & \textbf{64.0} \\
\rowcolor[HTML]{F2F2F2}
\parbox[t]{0.35\linewidth}{put the kiwi in the top drawer of the cabinet and close it\smallskip} & 0.0 & 28.0 & 28.0 & 42.0 & 34.0 & \textbf{60.0} \\
\rowcolor[HTML]{FFFFFF}
\parbox[t]{0.35\linewidth}{put the lemon on the plate\smallskip} & 0.0 & 36.0 & 32.0 & 48.0 & 48.0 & \textbf{74.0} \\
\rowcolor[HTML]{F2F2F2}
\parbox[t]{0.35\linewidth}{put the lime in the bowl\smallskip} & 0.0 & 70.0 & 60.0 & 80.0 & 84.0 & \textbf{96.0} \\
\rowcolor[HTML]{FFFFFF}
\parbox[t]{0.35\linewidth}{put the mango in the box and close it\smallskip} & 0.0 & 20.0 & 2.0 & 24.0 & 34.0 & \textbf{56.0} \\
\rowcolor[HTML]{F2F2F2}
\parbox[t]{0.35\linewidth}{put the mug in the left side of the dish drainer\smallskip} & 0.0 & 22.0 & 22.0 & 24.0 & 42.0 & \textbf{48.0} \\
\rowcolor[HTML]{FFFFFF}
\parbox[t]{0.35\linewidth}{put the mushroom in the bowl\smallskip} & 0.0 & 38.0 & 30.0 & 38.0 & 48.0 & \textbf{66.0} \\
\rowcolor[HTML]{F2F2F2}
\parbox[t]{0.35\linewidth}{put the onion on the tray\smallskip} & 0.0 & 56.0 & 52.0 & 64.0 & 66.0 & \textbf{70.0} \\
\rowcolor[HTML]{FFFFFF}
\parbox[t]{0.35\linewidth}{open the top drawer of the cabinet and put the avocado in it\smallskip} & 0.0 & 10.0 & 6.0 & 28.0 & 40.0 & \textbf{58.0} \\
\rowcolor[HTML]{F2F2F2}
\parbox[t]{0.35\linewidth}{open the slide cabinet and put the book in it\smallskip} & 6.0 & 58.0 & 32.0 & 68.0 & 64.0 & \textbf{70.0} \\
\rowcolor[HTML]{FFFFFF}
\parbox[t]{0.35\linewidth}{open the bottom drawer of the cabinet and put the broccoli in it\smallskip} & 0.0 & 0.0 & 4.0 & 16.0 & 0.0 & \textbf{22.0} \\
\rowcolor[HTML]{F2F2F2}
\parbox[t]{0.35\linewidth}{open the box and put the can in it\smallskip} & 2.0 & 22.0 & 20.0 & 32.0 & 42.0 & \textbf{62.0} \\
\rowcolor[HTML]{FFFFFF}
\parbox[t]{0.35\linewidth}{open the slide cabinet and put the candle in it\smallskip} & 2.0 & 24.0 & 14.0 & 12.0 & 20.0 & \textbf{32.0} \\
\rowcolor[HTML]{F2F2F2}
\parbox[t]{0.35\linewidth}{open the microwave and put the canned food in it\smallskip} & 0.0 & 24.0 & 20.0 & 18.0 & 24.0 & \textbf{52.0} \\
\rowcolor[HTML]{FFFFFF}
\parbox[t]{0.35\linewidth}{open the top drawer of the cabinet and put the carrot in it\smallskip} & 0.0 & 12.0 & 12.0 & 42.0 & 34.0 & \textbf{64.0} \\
\rowcolor[HTML]{F2F2F2}
\parbox[t]{0.35\linewidth}{open the box and put the cheese in it\smallskip} & 0.0 & 14.0 & 8.0 & 30.0 & 38.0 & \textbf{58.0} \\
\rowcolor[HTML]{FFFFFF}
\parbox[t]{0.35\linewidth}{open the microwave and put the eggplant in it\smallskip} & 0.0 & 14.0 & 4.0 & 28.0 & 18.0 & \textbf{50.0} \\
\rowcolor[HTML]{F2F2F2}
\parbox[t]{0.35\linewidth}{open the middle drawer of the cabinet and put the lemon in it\smallskip} & 0.0 & 12.0 & 6.0 & 40.0 & 46.0 & \textbf{62.0} \\
\rowcolor[HTML]{FFFFFF}
\parbox[t]{0.35\linewidth}{open the bottom drawer of the cabinet and put the mushroom in it\smallskip} & 0.0 & 2.0 & 2.0 & 10.0 & 2.0 & \textbf{20.0} \\
\rowcolor[HTML]{F2F2F2}
\parbox[t]{0.35\linewidth}{open the middle drawer of the cabinet and put the onion in it\smallskip} & 2.0 & 14.0 & 6.0 & 28.0 & 32.0 & \textbf{46.0} \\
\rowcolor[HTML]{FFFFFF}
\parbox[t]{0.35\linewidth}{open the top drawer of the cabinet and put the peach in it\smallskip} & 2.0 & 6.0 & 18.0 & 20.0 & 22.0 & \textbf{52.0} \\
\rowcolor[HTML]{F2F2F2}
\parbox[t]{0.35\linewidth}{open the microwave and put the squash in it\smallskip} & 0.0 & 6.0 & 4.0 & 10.0 & 28.0 & \textbf{64.0} \\
\rowcolor[HTML]{FFFFFF}
\parbox[t]{0.35\linewidth}{open the slide cabinet and put the tomato in it\smallskip} & 4.0 & 44.0 & 60.0 & 80.0 & 68.0 & \textbf{90.0} \\
\rowcolor[HTML]{F2F2F2}
\parbox[t]{0.35\linewidth}{open the slide cabinet and put the water bottle in it\smallskip} & 4.0 & 16.0 & 54.0 & 60.0 & 60.0 & \textbf{72.0} \\
\rowcolor[HTML]{FFFFFF}
\parbox[t]{0.35\linewidth}{open the slide cabinet and put the wine in it\smallskip} & 0.0 & 18.0 & 22.0 & 34.0 & 32.0 & \textbf{46.0} \\
\rowcolor[HTML]{F2F2F2}
\parbox[t]{0.35\linewidth}{put the orange on the cutting board\smallskip} & 2.0 & 22.0 & 14.0 & 54.0 & 48.0 & \textbf{56.0} \\
\rowcolor[HTML]{FFFFFF}
\parbox[t]{0.35\linewidth}{put the orange on the plate\smallskip} & 2.0 & 40.0 & 36.0 & 64.0 & 70.0 & \textbf{74.0} \\
\rowcolor[HTML]{F2F2F2}
\parbox[t]{0.35\linewidth}{put the peach in the basket\smallskip} & 0.0 & 52.0 & 24.0 & 66.0 & 64.0 & \textbf{88.0} \\
\rowcolor[HTML]{FFFFFF}
\parbox[t]{0.35\linewidth}{put the peach on the cutting board\smallskip} & 2.0 & 52.0 & 32.0 & 50.0 & 52.0 & \textbf{58.0} \\
\rowcolor[HTML]{F2F2F2}
\parbox[t]{0.35\linewidth}{put the peach on the tray\smallskip} & 0.0 & 50.0 & 46.0 & 60.0 & 56.0 & \textbf{74.0} \\
\rowcolor[HTML]{FFFFFF}
\parbox[t]{0.35\linewidth}{put the plate in the left side of the dish drainer\smallskip} & 0.0 & 2.0 & 2.0 & 8.0 & 14.0 & \textbf{16.0} \\
\rowcolor[HTML]{F2F2F2}
\parbox[t]{0.35\linewidth}{put the potato in the microwave and close it\smallskip} & 0.0 & 8.0 & 8.0 & 24.0 & 22.0 & \textbf{38.0} \\
\rowcolor[HTML]{FFFFFF}
\parbox[t]{0.35\linewidth}{put the rolling pin in the right side of the dish drainer\smallskip} & 0.0 & 10.0 & 8.0 & 20.0 & 22.0 & \textbf{36.0} \\
\rowcolor[HTML]{F2F2F2}
\parbox[t]{0.35\linewidth}{put the rolling pin in the pan\smallskip} & 2.0 & 34.0 & 30.0 & 58.0 & 56.0 & \textbf{60.0} \\
\rowcolor[HTML]{FFFFFF}
\parbox[t]{0.35\linewidth}{put the rolling pin on the tray\smallskip} & 2.0 & 30.0 & 26.0 & \textbf{62.0} & 50.0 & 50.0 \\
\rowcolor[HTML]{F2F2F2}
\parbox[t]{0.35\linewidth}{put the sponge in the middle drawer of the cabinet and close it\smallskip} & 0.0 & 12.0 & 0.0 & 10.0 & 12.0 & \textbf{24.0} \\
\rowcolor[HTML]{FFFFFF}
\parbox[t]{0.35\linewidth}{put the squash in the pan\smallskip} & 0.0 & 26.0 & 28.0 & 46.0 & \textbf{66.0} & 64.0 \\
\rowcolor[HTML]{F2F2F2}
\parbox[t]{0.35\linewidth}{put the squash on the tray\smallskip} & 2.0 & 24.0 & 26.0 & 56.0 & \textbf{68.0} & 66.0 \\
\rowcolor[HTML]{FFFFFF}
\parbox[t]{0.35\linewidth}{put the tomato on the cutting board\smallskip} & 8.0 & 72.0 & 58.0 & \textbf{88.0} & 76.0 & 86.0 \\
\rowcolor[HTML]{F2F2F2}
\parbox[t]{0.35\linewidth}{put the tomato in the pan\smallskip} & 8.0 & 64.0 & \textbf{92.0} & 86.0 & 82.0 & 88.0 \\
\rowcolor[HTML]{FFFFFF}
\parbox[t]{0.35\linewidth}{put the tomato on the tray\smallskip} & 8.0 & 72.0 & 74.0 & 90.0 & 70.0 & \textbf{96.0} \\
\rowcolor[HTML]{F2F2F2}
\parbox[t]{0.35\linewidth}{put the water bottle in the basket\smallskip} & 4.0 & 34.0 & 20.0 & 52.0 & 66.0 & \textbf{78.0} \\
\rowcolor[HTML]{FFFFFF}
\parbox[t]{0.35\linewidth}{put the water bottle on the tray\smallskip} & 10.0 & 30.0 & 38.0 & 50.0 & 50.0 & \textbf{74.0} \\
\rowcolor[HTML]{F2F2F2}
\parbox[t]{0.35\linewidth}{put the wine in the box\smallskip} & 0.0 & 34.0 & 18.0 & 58.0 & 70.0 & \textbf{76.0} \\
\rowcolor[HTML]{FFFFFF}
\parbox[t]{0.35\linewidth}{put the wine on the tray\smallskip} & 2.0 & 30.0 & 28.0 & 44.0 & 66.0 & \textbf{68.0} \\
\rowcolor[HTML]{F2F2F2}
\parbox[t]{0.35\linewidth}{put the yogurt in the basket\smallskip} & 2.0 & 26.0 & 10.0 & 44.0 & 48.0 & \textbf{56.0} \\
\bottomrule
\end{longtable}

%% file: tables/mesa-spatial-sr-table.tex
\begin{longtable}{lcccccc}
\caption{Success rates for all tasks in the MESA-Spatial task set.}\label{tab:mesa-spatial-per-task-sr} \\
\toprule
\rowcolor[HTML]{FFDEB4} \textbf{MESA-Spatial} & \textbf{DP} & \textbf{PG-FM} & \textbf{GR00T-N1.6} & $\boldsymbol{\pi}\mathbf{_0}$ & \textbf{FAST} & $\boldsymbol{\pi}\mathbf{_{0.5}}$ \\
\midrule
\endfirsthead
\toprule
\rowcolor[HTML]{FFDEB4} \textbf{MESA-Spatial} & \textbf{DP} & \textbf{PG-FM} & \textbf{GR00T-N1.6} & $\boldsymbol{\pi}\mathbf{_0}$ & \textbf{FAST} & $\boldsymbol{\pi}\mathbf{_{0.5}}$ \\
\midrule
\endhead
\rowcolor[HTML]{FFFFFF}
\parbox[t]{0.35\linewidth}{open the slide cabinet and put the tomato in it\smallskip} & 2.0 & 34.0 & 30.0 & 58.0 & 54.0 & \textbf{74.0} \\
\rowcolor[HTML]{F2F2F2}
\parbox[t]{0.35\linewidth}{open the slide cabinet and put the water bottle in it\smallskip} & 0.0 & 12.0 & 6.0 & 16.0 & 28.0 & \textbf{46.0} \\
\rowcolor[HTML]{FFFFFF}
\parbox[t]{0.35\linewidth}{put the rolling pin in the right side of the dish drainer\smallskip} & 0.0 & 4.0 & 6.0 & 8.0 & 32.0 & \textbf{38.0} \\
\rowcolor[HTML]{F2F2F2}
\parbox[t]{0.35\linewidth}{put the rolling pin in the pan\smallskip} & 4.0 & 12.0 & 36.0 & 26.0 & 46.0 & \textbf{62.0} \\
\rowcolor[HTML]{FFFFFF}
\parbox[t]{0.35\linewidth}{put the rolling pin on the tray\smallskip} & 4.0 & 28.0 & 22.0 & 28.0 & 54.0 & \textbf{72.0} \\
\rowcolor[HTML]{F2F2F2}
\parbox[t]{0.35\linewidth}{put the tomato on the cutting board\smallskip} & 0.0 & 54.0 & 28.0 & 58.0 & 72.0 & \textbf{78.0} \\
\rowcolor[HTML]{FFFFFF}
\parbox[t]{0.35\linewidth}{put the tomato in the pan\smallskip} & 0.0 & 64.0 & 42.0 & 76.0 & 74.0 & \textbf{90.0} \\
\rowcolor[HTML]{F2F2F2}
\parbox[t]{0.35\linewidth}{put the tomato on the tray\smallskip} & 0.0 & 54.0 & 46.0 & 68.0 & 76.0 & \textbf{86.0} \\
\rowcolor[HTML]{FFFFFF}
\parbox[t]{0.35\linewidth}{put the water bottle in the basket\smallskip} & 0.0 & 40.0 & 6.0 & 44.0 & 58.0 & \textbf{76.0} \\
\rowcolor[HTML]{F2F2F2}
\parbox[t]{0.35\linewidth}{put the water bottle on the tray\smallskip} & 0.0 & 48.0 & 10.0 & 40.0 & 42.0 & \textbf{64.0} \\
\bottomrule
\end{longtable}

%% file: tables/mesa-instance-sr-table.tex
\begin{longtable}{lcccccc}
\caption{Success rates for all tasks in the MESA-Instance task set.}\label{tab:mesa-instance-per-task-sr} \\
\toprule
\rowcolor[HTML]{FFDEB4} \textbf{MESA-Instance} & \textbf{DP} & \textbf{PG-FM} & \textbf{GR00T-N1.6} & $\boldsymbol{\pi}\mathbf{_0}$ & \textbf{FAST} & $\boldsymbol{\pi}\mathbf{_{0.5}}$ \\
\midrule
\endfirsthead
\toprule
\rowcolor[HTML]{FFDEB4} \textbf{MESA-Instance} & \textbf{DP} & \textbf{PG-FM} & \textbf{GR00T-N1.6} & $\boldsymbol{\pi}\mathbf{_0}$ & \textbf{FAST} & $\boldsymbol{\pi}\mathbf{_{0.5}}$ \\
\midrule
\endhead
\rowcolor[HTML]{FFFFFF}
\parbox[t]{0.35\linewidth}{put the apple on the tray\smallskip} & 10.0 & 38.0 & 46.0 & 48.0 & 56.0 & \textbf{64.0} \\
\rowcolor[HTML]{F2F2F2}
\parbox[t]{0.35\linewidth}{put the beer in the slide cabinet and close it\smallskip} & 0.0 & 6.0 & 4.0 & 10.0 & 6.0 & \textbf{14.0} \\
\rowcolor[HTML]{FFFFFF}
\parbox[t]{0.35\linewidth}{put the bell pepper in the box and close it\smallskip} & 0.0 & 10.0 & 4.0 & 10.0 & 14.0 & \textbf{30.0} \\
\rowcolor[HTML]{F2F2F2}
\parbox[t]{0.35\linewidth}{put the cheese on the plate\smallskip} & 0.0 & 4.0 & 2.0 & 2.0 & 12.0 & \textbf{28.0} \\
\rowcolor[HTML]{FFFFFF}
\parbox[t]{0.35\linewidth}{put the fish in the pan\smallskip} & 4.0 & 32.0 & 54.0 & 60.0 & 76.0 & \textbf{80.0} \\
\rowcolor[HTML]{F2F2F2}
\parbox[t]{0.35\linewidth}{put the jam in the basket\smallskip} & 0.0 & 50.0 & 24.0 & 52.0 & 44.0 & \textbf{78.0} \\
\rowcolor[HTML]{FFFFFF}
\parbox[t]{0.35\linewidth}{put the mushroom in the bowl\smallskip} & 0.0 & 28.0 & 24.0 & 46.0 & 46.0 & \textbf{72.0} \\
\rowcolor[HTML]{F2F2F2}
\parbox[t]{0.35\linewidth}{open the top drawer of the cabinet and put the carrot in it\smallskip} & 2.0 & 16.0 & 16.0 & 28.0 & 30.0 & \textbf{64.0} \\
\rowcolor[HTML]{FFFFFF}
\parbox[t]{0.35\linewidth}{open the top drawer of the cabinet and put the peach in it\smallskip} & 0.0 & 8.0 & 2.0 & 22.0 & 30.0 & \textbf{60.0} \\
\rowcolor[HTML]{F2F2F2}
\parbox[t]{0.35\linewidth}{open the microwave and put the squash in it\smallskip} & 0.0 & 2.0 & 2.0 & 8.0 & 26.0 & \textbf{30.0} \\
\rowcolor[HTML]{FFFFFF}
\parbox[t]{0.35\linewidth}{open the slide cabinet and put the wine in it\smallskip} & 2.0 & 2.0 & 8.0 & 10.0 & 12.0 & \textbf{40.0} \\
\rowcolor[HTML]{F2F2F2}
\parbox[t]{0.35\linewidth}{put the orange on the plate\smallskip} & 2.0 & 22.0 & 32.0 & 40.0 & 44.0 & \textbf{56.0} \\
\rowcolor[HTML]{FFFFFF}
\parbox[t]{0.35\linewidth}{put the peach in the basket\smallskip} & 0.0 & 38.0 & 36.0 & 74.0 & 58.0 & \textbf{82.0} \\
\rowcolor[HTML]{F2F2F2}
\parbox[t]{0.35\linewidth}{put the peach on the cutting board\smallskip} & 0.0 & 40.0 & 48.0 & 40.0 & 50.0 & \textbf{64.0} \\
\rowcolor[HTML]{FFFFFF}
\parbox[t]{0.35\linewidth}{put the peach on the tray\smallskip} & 0.0 & 54.0 & 50.0 & 74.0 & \textbf{78.0} & \textbf{78.0} \\
\rowcolor[HTML]{F2F2F2}
\parbox[t]{0.35\linewidth}{put the potato in the microwave and close it\smallskip} & 0.0 & 2.0 & 2.0 & 18.0 & 12.0 & \textbf{30.0} \\
\rowcolor[HTML]{FFFFFF}
\parbox[t]{0.35\linewidth}{put the squash in the pan\smallskip} & 2.0 & 12.0 & 30.0 & 20.0 & 44.0 & \textbf{56.0} \\
\rowcolor[HTML]{F2F2F2}
\parbox[t]{0.35\linewidth}{put the squash on the tray\smallskip} & 4.0 & 24.0 & 26.0 & 36.0 & 36.0 & \textbf{52.0} \\
\rowcolor[HTML]{FFFFFF}
\parbox[t]{0.35\linewidth}{put the wine in the box\smallskip} & 2.0 & 22.0 & 18.0 & 34.0 & 40.0 & \textbf{50.0} \\
\rowcolor[HTML]{F2F2F2}
\parbox[t]{0.35\linewidth}{put the wine on the tray\smallskip} & 10.0 & 44.0 & 26.0 & 48.0 & 68.0 & \textbf{80.0} \\
\bottomrule
\end{longtable}

%% file: tables/mesa-composite-sr-table.tex
\begin{longtable}{lcccccc}
\caption{Success rates for all tasks in the MESA-Composite task set.}\label{tab:mesa-composite-per-task-sr} \\
\toprule
\rowcolor[HTML]{FFDEB4} \textbf{MESA-Composite} & \textbf{DP} & \textbf{PG-FM} & \textbf{GR00T-N1.6} & $\boldsymbol{\pi}\mathbf{_0}$ & \textbf{FAST} & $\boldsymbol{\pi}\mathbf{_{0.5}}$ \\
\midrule
\endfirsthead
\toprule
\rowcolor[HTML]{FFDEB4} \textbf{MESA-Composite} & \textbf{DP} & \textbf{PG-FM} & \textbf{GR00T-N1.6} & $\boldsymbol{\pi}\mathbf{_0}$ & \textbf{FAST} & $\boldsymbol{\pi}\mathbf{_{0.5}}$ \\
\midrule
\endhead
\rowcolor[HTML]{FFFFFF}
\parbox[t]{0.35\linewidth}{put the apple on the plate\smallskip} & 0.0 & 66.0 & 54.0 & 70.0 & 64.0 & \textbf{84.0} \\
\rowcolor[HTML]{F2F2F2}
\parbox[t]{0.35\linewidth}{put the broccoli and mushroom on the cutting board\smallskip} & 0.0 & 0.0 & 0.0 & 0.0 & 2.0 & \textbf{4.0} \\
\rowcolor[HTML]{FFFFFF}
\parbox[t]{0.35\linewidth}{put the carrot in the pan\smallskip} & 2.0 & 30.0 & 42.0 & 58.0 & 60.0 & \textbf{74.0} \\
\rowcolor[HTML]{F2F2F2}
\parbox[t]{0.35\linewidth}{put the cheese in the basket\smallskip} & 0.0 & 28.0 & 6.0 & 36.0 & \textbf{62.0} & \textbf{62.0} \\
\rowcolor[HTML]{FFFFFF}
\parbox[t]{0.35\linewidth}{put the cheese and mug on the tray\smallskip} & 0.0 & 0.0 & 0.0 & \textbf{2.0} & 0.0 & \textbf{2.0} \\
\rowcolor[HTML]{F2F2F2}
\parbox[t]{0.35\linewidth}{put the cup in the left side of the dish drainer\smallskip} & 4.0 & 22.0 & \textbf{34.0} & 24.0 & 28.0 & \textbf{34.0} \\
\rowcolor[HTML]{FFFFFF}
\parbox[t]{0.35\linewidth}{put the fish and garlic in the pan\smallskip} & \textbf{0.0} & \textbf{0.0} & \textbf{0.0} & \textbf{0.0} & \textbf{0.0} & \textbf{0.0} \\
\rowcolor[HTML]{F2F2F2}
\parbox[t]{0.35\linewidth}{put the lemon on the cutting board\smallskip} & 0.0 & 34.0 & 40.0 & 54.0 & 44.0 & \textbf{76.0} \\
\rowcolor[HTML]{FFFFFF}
\parbox[t]{0.35\linewidth}{put the lemon and lime in the basket\smallskip} & 0.0 & 0.0 & 0.0 & 2.0 & 6.0 & \textbf{12.0} \\
\rowcolor[HTML]{F2F2F2}
\parbox[t]{0.35\linewidth}{put the mushroom on the tray\smallskip} & 0.0 & 52.0 & 48.0 & 52.0 & 56.0 & \textbf{74.0} \\
\rowcolor[HTML]{FFFFFF}
\parbox[t]{0.35\linewidth}{put the onion in the bowl\smallskip} & 0.0 & 40.0 & 20.0 & 70.0 & 62.0 & \textbf{80.0} \\
\rowcolor[HTML]{F2F2F2}
\parbox[t]{0.35\linewidth}{open the middle drawer of the cabinet and put the apple in it\smallskip} & 0.0 & 14.0 & 12.0 & 38.0 & 24.0 & \textbf{50.0} \\
\rowcolor[HTML]{FFFFFF}
\parbox[t]{0.35\linewidth}{open the box and put the apple in it and close it\smallskip} & 0.0 & 0.0 & 0.0 & 4.0 & 2.0 & \textbf{44.0} \\
\rowcolor[HTML]{F2F2F2}
\parbox[t]{0.35\linewidth}{open the box and put the beer in it\smallskip} & 0.0 & 6.0 & 10.0 & 10.0 & 14.0 & \textbf{38.0} \\
\rowcolor[HTML]{FFFFFF}
\parbox[t]{0.35\linewidth}{open the middle drawer of the cabinet and put the broccoli in it\smallskip} & 0.0 & 14.0 & 6.0 & 30.0 & 34.0 & \textbf{66.0} \\
\rowcolor[HTML]{F2F2F2}
\parbox[t]{0.35\linewidth}{open the top drawer of the cabinet and put the carrot in it and close it\smallskip} & 0.0 & 0.0 & 4.0 & 10.0 & 14.0 & \textbf{46.0} \\
\rowcolor[HTML]{FFFFFF}
\parbox[t]{0.35\linewidth}{open the slide cabinet and put the jam in it\smallskip} & 0.0 & 20.0 & 16.0 & 52.0 & 48.0 & \textbf{74.0} \\
\rowcolor[HTML]{F2F2F2}
\parbox[t]{0.35\linewidth}{open the middle drawer of the cabinet and put the mushroom in it and close it\smallskip} & 0.0 & 2.0 & 0.0 & 6.0 & 10.0 & \textbf{40.0} \\
\rowcolor[HTML]{FFFFFF}
\parbox[t]{0.35\linewidth}{open the microwave and put the potato in it and close it\smallskip} & 0.0 & 0.0 & 0.0 & 0.0 & 0.0 & \textbf{2.0} \\
\rowcolor[HTML]{F2F2F2}
\parbox[t]{0.35\linewidth}{put the sponge in the right side of the dish drainer\smallskip} & 0.0 & 26.0 & 8.0 & 32.0 & 48.0 & \textbf{72.0} \\
\bottomrule
\end{longtable}

%% file: tables/mesa-object-sr-table.tex
\begin{longtable}{lcccccc}
\caption{Success rates for all tasks in the MESA-Category task set.}\label{tab:mesa-category-per-task-sr} \\
\toprule
\rowcolor[HTML]{FFDEB4} \textbf{MESA-Category} & \textbf{DP} & \textbf{PG-FM} & \textbf{GR00T-N1.6} & $\boldsymbol{\pi}\mathbf{_0}$ & \textbf{FAST} & $\boldsymbol{\pi}\mathbf{_{0.5}}$ \\
\midrule
\endfirsthead
\toprule
\rowcolor[HTML]{FFDEB4} \textbf{MESA-Category} & \textbf{DP} & \textbf{PG-FM} & \textbf{GR00T-N1.6} & $\boldsymbol{\pi}\mathbf{_0}$ & \textbf{FAST} & $\boldsymbol{\pi}\mathbf{_{0.5}}$ \\
\midrule
\endhead
\rowcolor[HTML]{FFFFFF}
\parbox[t]{0.35\linewidth}{put the beet in the pot\smallskip} & 0.0 & 0.0 & 12.0 & 8.0 & 10.0 & \textbf{38.0} \\
\rowcolor[HTML]{F2F2F2}
\parbox[t]{0.35\linewidth}{put the brussel sprout in the pan\smallskip} & 0.0 & 2.0 & 6.0 & \textbf{8.0} & 4.0 & \textbf{8.0} \\
\rowcolor[HTML]{FFFFFF}
\parbox[t]{0.35\linewidth}{put the chili pepper in the bowl\smallskip} & 0.0 & 20.0 & \textbf{38.0} & 22.0 & 4.0 & 26.0 \\
\rowcolor[HTML]{F2F2F2}
\parbox[t]{0.35\linewidth}{put the cupcake in the basket\smallskip} & 2.0 & 14.0 & \textbf{28.0} & 14.0 & 8.0 & 26.0 \\
\rowcolor[HTML]{FFFFFF}
\parbox[t]{0.35\linewidth}{put the ginger on the cutting board\smallskip} & 0.0 & 10.0 & 12.0 & 0.0 & \textbf{26.0} & 20.0 \\
\rowcolor[HTML]{F2F2F2}
\parbox[t]{0.35\linewidth}{put the grapes in the basket\smallskip} & 2.0 & 12.0 & 16.0 & 24.0 & 8.0 & \textbf{28.0} \\
\rowcolor[HTML]{FFFFFF}
\parbox[t]{0.35\linewidth}{put the ice cream in the bowl\smallskip} & 0.0 & 4.0 & 2.0 & \textbf{34.0} & 4.0 & 32.0 \\
\rowcolor[HTML]{F2F2F2}
\parbox[t]{0.35\linewidth}{put the kebabs in the pan\smallskip} & 0.0 & 20.0 & 18.0 & 12.0 & 16.0 & \textbf{44.0} \\
\rowcolor[HTML]{FFFFFF}
\parbox[t]{0.35\linewidth}{put the olive oil in the basket\smallskip} & 0.0 & 4.0 & 2.0 & 6.0 & 6.0 & \textbf{46.0} \\
\rowcolor[HTML]{F2F2F2}
\parbox[t]{0.35\linewidth}{put the pear in the pot\smallskip} & 4.0 & 0.0 & \textbf{20.0} & 12.0 & 2.0 & 16.0 \\
\rowcolor[HTML]{FFFFFF}
\parbox[t]{0.35\linewidth}{put the pomegranate in the top drawer of the cabinet\smallskip} & 0.0 & \textbf{8.0} & 6.0 & 2.0 & 2.0 & 4.0 \\
\rowcolor[HTML]{F2F2F2}
\parbox[t]{0.35\linewidth}{put the potato on the baking sheet\smallskip} & 2.0 & 4.0 & 16.0 & 34.0 & 8.0 & \textbf{42.0} \\
\rowcolor[HTML]{FFFFFF}
\parbox[t]{0.35\linewidth}{put the radish on the cutting board\smallskip} & 2.0 & 4.0 & 8.0 & 8.0 & 8.0 & \textbf{32.0} \\
\rowcolor[HTML]{F2F2F2}
\parbox[t]{0.35\linewidth}{put the raspberry in the basket\smallskip} & 0.0 & 14.0 & 10.0 & 8.0 & 6.0 & \textbf{18.0} \\
\rowcolor[HTML]{FFFFFF}
\parbox[t]{0.35\linewidth}{put the salt shaker in the box and close it\smallskip} & 0.0 & 2.0 & 2.0 & 4.0 & 4.0 & \textbf{34.0} \\
\rowcolor[HTML]{F2F2F2}
\parbox[t]{0.35\linewidth}{put the sausage on the baking sheet\smallskip} & 2.0 & 8.0 & \textbf{20.0} & 6.0 & 0.0 & 14.0 \\
\rowcolor[HTML]{FFFFFF}
\parbox[t]{0.35\linewidth}{put the scone on the plate\smallskip} & 0.0 & 0.0 & 12.0 & 2.0 & \textbf{14.0} & 12.0 \\
\rowcolor[HTML]{F2F2F2}
\parbox[t]{0.35\linewidth}{put the strawberry in the bowl\smallskip} & 0.0 & 28.0 & 48.0 & 20.0 & 34.0 & \textbf{74.0} \\
\rowcolor[HTML]{FFFFFF}
\parbox[t]{0.35\linewidth}{put the sushi on the plate\smallskip} & 0.0 & 0.0 & 2.0 & 8.0 & 2.0 & \textbf{10.0} \\
\rowcolor[HTML]{F2F2F2}
\parbox[t]{0.35\linewidth}{put the thermos in the slide cabinet\smallskip} & 0.0 & 0.0 & \textbf{4.0} & 0.0 & 2.0 & \textbf{4.0} \\
\bottomrule
\end{longtable}

%% file: tables/bimesa_id_per_task_success.tex
{\small\setlength{\tabcolsep}{3pt}
\begin{longtable}{lcccccccc}
\caption{Success rates for all tasks in the BiMESA-ID task set.} \\
\toprule
\rowcolor[HTML]{FFDEB4} \textbf{BiMESA-ID} & \textbf{DP} & \textbf{PG-Bin} & \textbf{PG-FM} & \textbf{GR00T-N1.6} & $\boldsymbol{\pi}\mathbf{_0}$ & \textbf{FAST} & $\boldsymbol{\pi}\mathbf{_{0.5}}$ & \makecell{\textbf{GPT-6.1-}\\\textbf{Sol}} \\
\midrule
\endfirsthead
\toprule
\rowcolor[HTML]{FFDEB4} \textbf{BiMESA-ID} & \textbf{DP} & \textbf{PG-Bin} & \textbf{PG-FM} & \textbf{GR00T-N1.6} & $\boldsymbol{\pi}\mathbf{_0}$ & \textbf{FAST} & $\boldsymbol{\pi}\mathbf{_{0.5}}$ & \makecell{\textbf{GPT-6.1-}\\\textbf{Sol}} \\
\midrule
\endhead
\rowcolor[HTML]{FFFFFF}
\parbox[t]{0.31\linewidth}{put the apple in the bowl {\scriptsize\mbox{(left arm)}}\smallskip} & 16.0 & 76.0 & 76.0 & 86.0 & 78.0 & 94.0 & \textbf{98.0} & 80.0 \\
\rowcolor[HTML]{F2F2F2}
\parbox[t]{0.31\linewidth}{put the apple in the box and close it\smallskip} & 0.0 & 0.0 & 8.0 & 16.0 & 16.0 & \textbf{50.0} & 46.0 & 10.0 \\
\rowcolor[HTML]{FFFFFF}
\parbox[t]{0.31\linewidth}{put the apple on the tray {\scriptsize\mbox{(right arm)}}\smallskip} & 28.0 & 48.0 & 74.0 & 78.0 & 66.0 & 88.0 & \textbf{92.0} & 70.0 \\
\rowcolor[HTML]{F2F2F2}
\parbox[t]{0.31\linewidth}{put the beer on the plate {\scriptsize\mbox{(left arm)}}\smallskip} & 26.0 & 70.0 & 70.0 & 76.0 & 56.0 & 88.0 & \textbf{98.0} & 60.0 \\
\rowcolor[HTML]{FFFFFF}
\parbox[t]{0.31\linewidth}{put the beer in the slide cabinet and close it\smallskip} & 0.0 & 0.0 & 0.0 & 6.0 & 10.0 & \textbf{22.0} & 20.0 & 0.0 \\
\rowcolor[HTML]{F2F2F2}
\parbox[t]{0.31\linewidth}{put the bell pepper in the basket {\scriptsize\mbox{(right arm)}}\smallskip} & 4.0 & 16.0 & 62.0 & 46.0 & 62.0 & 76.0 & \textbf{80.0} & 40.0 \\
\rowcolor[HTML]{FFFFFF}
\parbox[t]{0.31\linewidth}{put the bell pepper on the tray {\scriptsize\mbox{(right arm)}}\smallskip} & 22.0 & 24.0 & 74.0 & 66.0 & 80.0 & 88.0 & \textbf{90.0} & 70.0 \\
\rowcolor[HTML]{F2F2F2}
\parbox[t]{0.31\linewidth}{put the bottled drink in the basket {\scriptsize\mbox{(left arm)}}\smallskip} & 12.0 & 52.0 & 36.0 & 54.0 & 54.0 & \textbf{64.0} & \textbf{64.0} & 40.0 \\
\rowcolor[HTML]{FFFFFF}
\parbox[t]{0.31\linewidth}{put the can in the pan {\scriptsize\mbox{(left arm)}}\smallskip} & 10.0 & 78.0 & 58.0 & 64.0 & 62.0 & 84.0 & \textbf{92.0} & 80.0 \\
\rowcolor[HTML]{F2F2F2}
\parbox[t]{0.31\linewidth}{put the candle on the tray {\scriptsize\mbox{(right arm)}}\smallskip} & 20.0 & 28.0 & 58.0 & 74.0 & 60.0 & 86.0 & \textbf{88.0} & 30.0 \\
\rowcolor[HTML]{FFFFFF}
\parbox[t]{0.31\linewidth}{put the coffee cup on the tray {\scriptsize\mbox{(right arm)}}\smallskip} & 10.0 & 22.0 & 52.0 & 78.0 & 56.0 & \textbf{80.0} & \textbf{80.0} & 50.0 \\
\rowcolor[HTML]{F2F2F2}
\parbox[t]{0.31\linewidth}{put the garlic in the pan {\scriptsize\mbox{(right arm)}}\smallskip} & 12.0 & 22.0 & 64.0 & 62.0 & 44.0 & 72.0 & \textbf{80.0} & 60.0 \\
\rowcolor[HTML]{FFFFFF}
\parbox[t]{0.31\linewidth}{put the alcohol in the basket {\scriptsize\mbox{(handoff)}}\smallskip} & 10.0 & 14.0 & 26.0 & 34.0 & \textbf{36.0} & 6.0 & \textbf{36.0} & 0.0 \\
\rowcolor[HTML]{F2F2F2}
\parbox[t]{0.31\linewidth}{put the beer on the plate {\scriptsize\mbox{(handoff)}}\smallskip} & 20.0 & 8.0 & 46.0 & 52.0 & 64.0 & 30.0 & \textbf{68.0} & 40.0 \\
\rowcolor[HTML]{FFFFFF}
\parbox[t]{0.31\linewidth}{put the bottled drink in the basket {\scriptsize\mbox{(handoff)}}\smallskip} & 18.0 & 28.0 & 42.0 & 52.0 & 50.0 & 28.0 & \textbf{62.0} & 0.0 \\
\rowcolor[HTML]{F2F2F2}
\parbox[t]{0.31\linewidth}{put the bottled water in the basket {\scriptsize\mbox{(handoff)}}\smallskip} & 6.0 & 16.0 & \textbf{32.0} & 18.0 & 12.0 & 10.0 & \textbf{32.0} & 10.0 \\
\rowcolor[HTML]{FFFFFF}
\parbox[t]{0.31\linewidth}{put the can in the pan {\scriptsize\mbox{(handoff)}}\smallskip} & 10.0 & 26.0 & 50.0 & 38.0 & 48.0 & 26.0 & \textbf{70.0} & 60.0 \\
\rowcolor[HTML]{F2F2F2}
\parbox[t]{0.31\linewidth}{put the candle on the tray {\scriptsize\mbox{(handoff)}}\smallskip} & 18.0 & 22.0 & 12.0 & 40.0 & 22.0 & \textbf{42.0} & 40.0 & 10.0 \\
\rowcolor[HTML]{FFFFFF}
\parbox[t]{0.31\linewidth}{put the coffee cup on the tray {\scriptsize\mbox{(handoff)}}\smallskip} & 14.0 & 12.0 & 22.0 & 16.0 & 34.0 & 38.0 & \textbf{68.0} & 50.0 \\
\rowcolor[HTML]{F2F2F2}
\parbox[t]{0.31\linewidth}{put the condiment on the plate {\scriptsize\mbox{(handoff)}}\smallskip} & 16.0 & 20.0 & 46.0 & 60.0 & 68.0 & 36.0 & \textbf{74.0} & 40.0 \\
\rowcolor[HTML]{FFFFFF}
\parbox[t]{0.31\linewidth}{put the jam in the right side of the dish drainer {\scriptsize\mbox{(handoff)}}\smallskip} & 4.0 & 2.0 & 22.0 & 6.0 & 30.0 & 14.0 & \textbf{60.0} & 10.0 \\
\rowcolor[HTML]{F2F2F2}
\parbox[t]{0.31\linewidth}{put the jug in the basket {\scriptsize\mbox{(handoff)}}\smallskip} & 14.0 & 10.0 & 18.0 & 30.0 & 10.0 & 12.0 & \textbf{32.0} & 20.0 \\
\rowcolor[HTML]{FFFFFF}
\parbox[t]{0.31\linewidth}{put the ketchup on the plate {\scriptsize\mbox{(handoff)}}\smallskip} & 28.0 & 16.0 & 46.0 & 64.0 & 62.0 & 30.0 & \textbf{86.0} & 20.0 \\
\rowcolor[HTML]{F2F2F2}
\parbox[t]{0.31\linewidth}{put the milk on the tray {\scriptsize\mbox{(handoff)}}\smallskip} & 6.0 & 20.0 & 22.0 & 24.0 & 26.0 & 10.0 & \textbf{48.0} & 40.0 \\
\rowcolor[HTML]{FFFFFF}
\parbox[t]{0.31\linewidth}{put the soap dispenser on the tray {\scriptsize\mbox{(handoff)}}\smallskip} & 16.0 & 24.0 & 36.0 & 38.0 & 30.0 & 20.0 & \textbf{56.0} & 30.0 \\
\rowcolor[HTML]{F2F2F2}
\parbox[t]{0.31\linewidth}{put the spray in the basket {\scriptsize\mbox{(handoff)}}\smallskip} & 4.0 & 16.0 & 18.0 & 22.0 & 34.0 & 12.0 & \textbf{52.0} & 20.0 \\
\rowcolor[HTML]{FFFFFF}
\parbox[t]{0.31\linewidth}{put the water bottle on the tray {\scriptsize\mbox{(handoff)}}\smallskip} & 8.0 & 18.0 & 60.0 & 26.0 & 60.0 & 20.0 & \textbf{74.0} & 40.0 \\
\rowcolor[HTML]{F2F2F2}
\parbox[t]{0.31\linewidth}{put the wine on the cutting board {\scriptsize\mbox{(handoff)}}\smallskip} & 4.0 & 16.0 & 64.0 & 44.0 & 56.0 & 20.0 & \textbf{68.0} & 30.0 \\
\rowcolor[HTML]{FFFFFF}
\parbox[t]{0.31\linewidth}{put the yogurt in the bowl {\scriptsize\mbox{(handoff)}}\smallskip} & 8.0 & 10.0 & 32.0 & 24.0 & 34.0 & 22.0 & \textbf{56.0} & 30.0 \\
\rowcolor[HTML]{F2F2F2}
\parbox[t]{0.31\linewidth}{put the jam in the basket {\scriptsize\mbox{(right arm)}}\smallskip} & 10.0 & 14.0 & 62.0 & 60.0 & 46.0 & 70.0 & \textbf{76.0} & 40.0 \\
\rowcolor[HTML]{FFFFFF}
\parbox[t]{0.31\linewidth}{put the kiwi on the cutting board {\scriptsize\mbox{(right arm)}}\smallskip} & 28.0 & 54.0 & 86.0 & 84.0 & 70.0 & 88.0 & \textbf{94.0} & 60.0 \\
\rowcolor[HTML]{F2F2F2}
\parbox[t]{0.31\linewidth}{put the kiwi on the plate {\scriptsize\mbox{(right arm)}}\smallskip} & 34.0 & 66.0 & 88.0 & 86.0 & 76.0 & 92.0 & \textbf{96.0} & 80.0 \\
\rowcolor[HTML]{FFFFFF}
\parbox[t]{0.31\linewidth}{put the lemon on the plate {\scriptsize\mbox{(right arm)}}\smallskip} & 16.0 & 30.0 & 88.0 & 84.0 & 70.0 & 92.0 & \textbf{96.0} & 90.0 \\
\rowcolor[HTML]{F2F2F2}
\parbox[t]{0.31\linewidth}{put the lemon in the box and close it\smallskip} & 0.0 & 6.0 & 6.0 & 8.0 & 20.0 & 66.0 & \textbf{70.0} & 0.0 \\
\rowcolor[HTML]{FFFFFF}
\parbox[t]{0.31\linewidth}{put the lime in the bowl {\scriptsize\mbox{(right arm)}}\smallskip} & 22.0 & 52.0 & 64.0 & 84.0 & 70.0 & 88.0 & \textbf{98.0} & 70.0 \\
\rowcolor[HTML]{F2F2F2}
\parbox[t]{0.31\linewidth}{put the lime on the tray {\scriptsize\mbox{(left arm)}}\smallskip} & 26.0 & 82.0 & 44.0 & 80.0 & 76.0 & \textbf{92.0} & 90.0 & 70.0 \\
\rowcolor[HTML]{FFFFFF}
\parbox[t]{0.31\linewidth}{put the mango in the basket {\scriptsize\mbox{(right arm)}}\smallskip} & 14.0 & 14.0 & 50.0 & 48.0 & 52.0 & 74.0 & \textbf{90.0} & 40.0 \\
\rowcolor[HTML]{F2F2F2}
\parbox[t]{0.31\linewidth}{put the mango on the plate {\scriptsize\mbox{(left arm)}}\smallskip} & 26.0 & 62.0 & 68.0 & 52.0 & 58.0 & 86.0 & \textbf{90.0} & 70.0 \\
\rowcolor[HTML]{FFFFFF}
\parbox[t]{0.31\linewidth}{put the mango in the slide cabinet and close it\smallskip} & 0.0 & 8.0 & 16.0 & 24.0 & 48.0 & 34.0 & \textbf{52.0} & 0.0 \\
\rowcolor[HTML]{F2F2F2}
\parbox[t]{0.31\linewidth}{put the mushroom in the bowl {\scriptsize\mbox{(right arm)}}\smallskip} & 2.0 & 22.0 & 42.0 & 46.0 & 52.0 & 62.0 & \textbf{68.0} & 30.0 \\
\rowcolor[HTML]{FFFFFF}
\parbox[t]{0.31\linewidth}{put the onion on the cutting board {\scriptsize\mbox{(right arm)}}\smallskip} & 28.0 & 38.0 & \textbf{92.0} & 70.0 & 58.0 & 86.0 & \textbf{92.0} & 60.0 \\
\rowcolor[HTML]{F2F2F2}
\parbox[t]{0.31\linewidth}{put the onion in the pan {\scriptsize\mbox{(left arm)}}\smallskip} & 26.0 & 66.0 & 82.0 & 66.0 & 56.0 & 86.0 & \textbf{94.0} & 60.0 \\
\rowcolor[HTML]{FFFFFF}
\parbox[t]{0.31\linewidth}{put the onion on the tray {\scriptsize\mbox{(right arm)}}\smallskip} & 26.0 & 32.0 & 74.0 & 76.0 & 58.0 & 76.0 & \textbf{84.0} & 70.0 \\
\rowcolor[HTML]{F2F2F2}
\parbox[t]{0.31\linewidth}{open the box and put the apple in it\smallskip} & 0.0 & 2.0 & 4.0 & 4.0 & 16.0 & 32.0 & \textbf{54.0} & 10.0 \\
\rowcolor[HTML]{FFFFFF}
\parbox[t]{0.31\linewidth}{open the slide cabinet and put the beer in it\smallskip} & 0.0 & 2.0 & 10.0 & 20.0 & 26.0 & 38.0 & 38.0 & \textbf{50.0} \\
\rowcolor[HTML]{F2F2F2}
\parbox[t]{0.31\linewidth}{open the slide cabinet and put the book in it\smallskip} & 0.0 & 0.0 & 0.0 & 28.0 & 24.0 & 20.0 & \textbf{36.0} & 10.0 \\
\rowcolor[HTML]{FFFFFF}
\parbox[t]{0.31\linewidth}{open the slide cabinet and put the can in it\smallskip} & 0.0 & 10.0 & 16.0 & 34.0 & 32.0 & 54.0 & \textbf{64.0} & 50.0 \\
\rowcolor[HTML]{F2F2F2}
\parbox[t]{0.31\linewidth}{open the box and put the can in it\smallskip} & 0.0 & 4.0 & 8.0 & 18.0 & 20.0 & 34.0 & \textbf{48.0} & 10.0 \\
\rowcolor[HTML]{FFFFFF}
\parbox[t]{0.31\linewidth}{open the box and put the can in it and close it\smallskip} & 0.0 & 0.0 & 0.0 & 0.0 & 2.0 & 6.0 & \textbf{10.0} & 0.0 \\
\rowcolor[HTML]{F2F2F2}
\parbox[t]{0.31\linewidth}{open the slide cabinet and put the candle in it\smallskip} & 0.0 & 0.0 & 0.0 & 22.0 & 26.0 & 54.0 & \textbf{58.0} & 10.0 \\
\rowcolor[HTML]{FFFFFF}
\parbox[t]{0.31\linewidth}{open the box and put the mango in it and close it\smallskip} & 0.0 & 0.0 & 2.0 & 0.0 & 6.0 & 16.0 & \textbf{26.0} & 0.0 \\
\rowcolor[HTML]{F2F2F2}
\parbox[t]{0.31\linewidth}{open the slide cabinet and put the orange in it and close it\smallskip} & 0.0 & 0.0 & 0.0 & 18.0 & 16.0 & 30.0 & \textbf{42.0} & 20.0 \\
\rowcolor[HTML]{FFFFFF}
\parbox[t]{0.31\linewidth}{put the orange on the cutting board {\scriptsize\mbox{(right arm)}}\smallskip} & 34.0 & 64.0 & 82.0 & 76.0 & 68.0 & 88.0 & \textbf{100.0} & 80.0 \\
\rowcolor[HTML]{F2F2F2}
\parbox[t]{0.31\linewidth}{put the orange on the plate {\scriptsize\mbox{(right arm)}}\smallskip} & 32.0 & 52.0 & 82.0 & 80.0 & 72.0 & 86.0 & \textbf{94.0} & 90.0 \\
\rowcolor[HTML]{FFFFFF}
\parbox[t]{0.31\linewidth}{put the tangerine on the cutting board {\scriptsize\mbox{(left arm)}}\smallskip} & 18.0 & 72.0 & 72.0 & 90.0 & 78.0 & 90.0 & 90.0 & \textbf{100.0} \\
\rowcolor[HTML]{F2F2F2}
\parbox[t]{0.31\linewidth}{put the tangerine on the tray {\scriptsize\mbox{(right arm)}}\smallskip} & 28.0 & 52.0 & 78.0 & 76.0 & 72.0 & 86.0 & 88.0 & \textbf{90.0} \\
\rowcolor[HTML]{FFFFFF}
\parbox[t]{0.31\linewidth}{put the yogurt in the basket {\scriptsize\mbox{(right arm)}}\smallskip} & 8.0 & 6.0 & 56.0 & 52.0 & 28.0 & \textbf{66.0} & \textbf{66.0} & 50.0 \\
\bottomrule
\end{longtable}
}

%% file: tables/bimesa_spatial_per_task_success.tex
{\small\setlength{\tabcolsep}{3pt}
\begin{longtable}{lcccccccc}
\caption{Success rates for all tasks in the BiMESA-Spatial task set.} \\
\toprule
\rowcolor[HTML]{FFDEB4} \textbf{BiMESA-Spatial} & \textbf{DP} & \textbf{PG-Bin} & \textbf{PG-FM} & \textbf{GR00T-N1.6} & $\boldsymbol{\pi}\mathbf{_0}$ & \textbf{FAST} & $\boldsymbol{\pi}\mathbf{_{0.5}}$ & \makecell{\textbf{GPT-6.1-}\\\textbf{Sol}} \\
\midrule
\endfirsthead
\toprule
\rowcolor[HTML]{FFDEB4} \textbf{BiMESA-Spatial} & \textbf{DP} & \textbf{PG-Bin} & \textbf{PG-FM} & \textbf{GR00T-N1.6} & $\boldsymbol{\pi}\mathbf{_0}$ & \textbf{FAST} & $\boldsymbol{\pi}\mathbf{_{0.5}}$ & \makecell{\textbf{GPT-6.1-}\\\textbf{Sol}} \\
\midrule
\endhead
\rowcolor[HTML]{FFFFFF}
\parbox[t]{0.31\linewidth}{put the apple in the bowl {\scriptsize\mbox{(right arm)}}\smallskip} & 24.0 & 56.0 & 76.0 & 84.0 & 76.0 & 88.0 & \textbf{92.0} & 50.0 \\
\rowcolor[HTML]{F2F2F2}
\parbox[t]{0.31\linewidth}{put the apple on the tray {\scriptsize\mbox{(left arm)}}\smallskip} & 24.0 & 78.0 & 72.0 & 82.0 & 58.0 & \textbf{86.0} & 80.0 & 70.0 \\
\rowcolor[HTML]{FFFFFF}
\parbox[t]{0.31\linewidth}{put the beer on the plate {\scriptsize\mbox{(right arm)}}\smallskip} & 22.0 & 36.0 & 64.0 & 72.0 & 42.0 & 80.0 & \textbf{88.0} & 50.0 \\
\rowcolor[HTML]{F2F2F2}
\parbox[t]{0.31\linewidth}{put the beer in the slide cabinet and close it {\scriptsize\mbox{(left arm, right fixture)}}\smallskip} & 0.0 & 0.0 & 0.0 & 0.0 & 0.0 & 4.0 & \textbf{6.0} & 0.0 \\
\rowcolor[HTML]{FFFFFF}
\parbox[t]{0.31\linewidth}{put the bell pepper in the basket {\scriptsize\mbox{(pick L, place R)}}\smallskip} & 20.0 & 32.0 & 52.0 & 44.0 & 56.0 & 60.0 & \textbf{80.0} & 10.0 \\
\rowcolor[HTML]{F2F2F2}
\parbox[t]{0.31\linewidth}{put the bell pepper on the tray {\scriptsize\mbox{(left arm)}}\smallskip} & 36.0 & 58.0 & 76.0 & 54.0 & 60.0 & \textbf{80.0} & 70.0 & 70.0 \\
\rowcolor[HTML]{FFFFFF}
\parbox[t]{0.31\linewidth}{put the bottled drink in the basket {\scriptsize\mbox{(right arm)}}\smallskip} & 2.0 & 12.0 & 42.0 & 38.0 & 32.0 & \textbf{60.0} & 58.0 & 0.0 \\
\rowcolor[HTML]{F2F2F2}
\parbox[t]{0.31\linewidth}{put the can in the pan {\scriptsize\mbox{(right arm)}}\smallskip} & 14.0 & 46.0 & 68.0 & 68.0 & 60.0 & \textbf{86.0} & \textbf{86.0} & 70.0 \\
\rowcolor[HTML]{FFFFFF}
\parbox[t]{0.31\linewidth}{put the candle on the tray {\scriptsize\mbox{(pick R, place L)}}\smallskip} & 14.0 & 22.0 & 20.0 & \textbf{38.0} & 28.0 & 32.0 & 36.0 & 30.0 \\
\rowcolor[HTML]{F2F2F2}
\parbox[t]{0.31\linewidth}{put the coffee cup on the tray {\scriptsize\mbox{(pick R, place L)}}\smallskip} & 10.0 & 16.0 & 30.0 & 22.0 & 26.0 & 28.0 & \textbf{48.0} & 20.0 \\
\rowcolor[HTML]{FFFFFF}
\parbox[t]{0.31\linewidth}{put the garlic in the pan {\scriptsize\mbox{(pick L, place R)}}\smallskip} & 10.0 & 24.0 & 48.0 & 34.0 & 48.0 & 58.0 & \textbf{90.0} & 70.0 \\
\rowcolor[HTML]{F2F2F2}
\parbox[t]{0.31\linewidth}{put the beer on the plate {\scriptsize\mbox{(handoff R$\to$L)}}\smallskip} & 50.0 & 26.0 & 48.0 & 56.0 & 56.0 & 62.0 & \textbf{90.0} & 30.0 \\
\rowcolor[HTML]{FFFFFF}
\parbox[t]{0.31\linewidth}{put the can in the pan {\scriptsize\mbox{(handoff R$\to$L)}}\smallskip} & 22.0 & 22.0 & 62.0 & 44.0 & 64.0 & 62.0 & \textbf{80.0} & 30.0 \\
\rowcolor[HTML]{F2F2F2}
\parbox[t]{0.31\linewidth}{put the condiment on the plate {\scriptsize\mbox{(handoff L$\to$R)}}\smallskip} & 16.0 & 32.0 & 26.0 & 42.0 & 38.0 & 12.0 & \textbf{56.0} & 30.0 \\
\rowcolor[HTML]{FFFFFF}
\parbox[t]{0.31\linewidth}{put the jug in the basket {\scriptsize\mbox{(handoff L$\to$R)}}\smallskip} & 2.0 & 2.0 & 12.0 & 22.0 & 24.0 & 14.0 & \textbf{30.0} & 10.0 \\
\rowcolor[HTML]{F2F2F2}
\parbox[t]{0.31\linewidth}{put the milk on the tray {\scriptsize\mbox{(handoff L$\to$R)}}\smallskip} & 4.0 & 12.0 & 36.0 & 22.0 & 14.0 & 30.0 & \textbf{52.0} & 30.0 \\
\rowcolor[HTML]{FFFFFF}
\parbox[t]{0.31\linewidth}{put the water bottle on the tray {\scriptsize\mbox{(handoff R$\to$L)}}\smallskip} & 8.0 & 16.0 & 24.0 & 26.0 & 22.0 & 24.0 & 26.0 & \textbf{50.0} \\
\rowcolor[HTML]{F2F2F2}
\parbox[t]{0.31\linewidth}{put the wine on the cutting board {\scriptsize\mbox{(handoff R$\to$L)}}\smallskip} & 14.0 & 18.0 & 20.0 & 46.0 & 46.0 & 44.0 & \textbf{52.0} & 40.0 \\
\rowcolor[HTML]{FFFFFF}
\parbox[t]{0.31\linewidth}{put the yogurt in the bowl {\scriptsize\mbox{(handoff L$\to$R)}}\smallskip} & 8.0 & 30.0 & 30.0 & 18.0 & 34.0 & 42.0 & \textbf{60.0} & 10.0 \\
\rowcolor[HTML]{F2F2F2}
\parbox[t]{0.31\linewidth}{put the jam in the basket {\scriptsize\mbox{(pick L, place R)}}\smallskip} & 14.0 & 34.0 & 30.0 & 36.0 & 54.0 & 44.0 & \textbf{72.0} & 40.0 \\
\rowcolor[HTML]{FFFFFF}
\parbox[t]{0.31\linewidth}{put the kiwi on the cutting board {\scriptsize\mbox{(pick L, place R)}}\smallskip} & 26.0 & 24.0 & 66.0 & 46.0 & 66.0 & 74.0 & \textbf{84.0} & 30.0 \\
\rowcolor[HTML]{F2F2F2}
\parbox[t]{0.31\linewidth}{put the kiwi on the plate {\scriptsize\mbox{(pick R, place L)}}\smallskip} & 26.0 & 52.0 & 70.0 & 44.0 & 68.0 & 68.0 & \textbf{88.0} & 40.0 \\
\rowcolor[HTML]{FFFFFF}
\parbox[t]{0.31\linewidth}{put the lemon on the plate {\scriptsize\mbox{(pick R, place L)}}\smallskip} & 14.0 & 28.0 & 74.0 & 38.0 & 72.0 & 80.0 & \textbf{94.0} & 90.0 \\
\rowcolor[HTML]{F2F2F2}
\parbox[t]{0.31\linewidth}{put the lime in the bowl {\scriptsize\mbox{(left arm)}}\smallskip} & 20.0 & 78.0 & 74.0 & 82.0 & 70.0 & \textbf{94.0} & 92.0 & 20.0 \\
\rowcolor[HTML]{FFFFFF}
\parbox[t]{0.31\linewidth}{put the lime on the tray {\scriptsize\mbox{(right arm)}}\smallskip} & 28.0 & 34.0 & 88.0 & 86.0 & 76.0 & 82.0 & \textbf{96.0} & 70.0 \\
\rowcolor[HTML]{F2F2F2}
\parbox[t]{0.31\linewidth}{put the mango in the basket {\scriptsize\mbox{(left arm)}}\smallskip} & 8.0 & 44.0 & 62.0 & 64.0 & 62.0 & 72.0 & \textbf{86.0} & 40.0 \\
\rowcolor[HTML]{FFFFFF}
\parbox[t]{0.31\linewidth}{put the mango on the plate {\scriptsize\mbox{(right arm)}}\smallskip} & 42.0 & 38.0 & 88.0 & 64.0 & 80.0 & 94.0 & \textbf{98.0} & 80.0 \\
\rowcolor[HTML]{F2F2F2}
\parbox[t]{0.31\linewidth}{put the mushroom in the bowl {\scriptsize\mbox{(pick L, place R)}}\smallskip} & 6.0 & 14.0 & 38.0 & 22.0 & 40.0 & 56.0 & \textbf{66.0} & 10.0 \\
\rowcolor[HTML]{FFFFFF}
\parbox[t]{0.31\linewidth}{put the onion on the cutting board {\scriptsize\mbox{(pick R, place L)}}\smallskip} & 32.0 & 32.0 & 82.0 & 36.0 & 58.0 & 78.0 & \textbf{94.0} & 40.0 \\
\rowcolor[HTML]{F2F2F2}
\parbox[t]{0.31\linewidth}{put the onion in the pan {\scriptsize\mbox{(right arm)}}\smallskip} & 22.0 & 34.0 & 80.0 & 58.0 & 62.0 & 78.0 & \textbf{96.0} & 80.0 \\
\rowcolor[HTML]{FFFFFF}
\parbox[t]{0.31\linewidth}{put the onion on the tray {\scriptsize\mbox{(pick L, place R)}}\smallskip} & 34.0 & 30.0 & 44.0 & 50.0 & 52.0 & 60.0 & 78.0 & \textbf{80.0} \\
\rowcolor[HTML]{F2F2F2}
\parbox[t]{0.31\linewidth}{open the slide cabinet and put the candle in it {\scriptsize\mbox{(right arm, left fixture)}}\smallskip} & 0.0 & 2.0 & 0.0 & 4.0 & 8.0 & 18.0 & \textbf{24.0} & 10.0 \\
\rowcolor[HTML]{FFFFFF}
\parbox[t]{0.31\linewidth}{open the slide cabinet and put the orange in it and close it {\scriptsize\mbox{(right arm, left fixture)}}\smallskip} & 0.0 & 0.0 & 0.0 & 2.0 & 2.0 & 6.0 & 0.0 & \textbf{20.0} \\
\rowcolor[HTML]{F2F2F2}
\parbox[t]{0.31\linewidth}{put the orange on the cutting board {\scriptsize\mbox{(left arm)}}\smallskip} & 30.0 & 60.0 & 76.0 & 76.0 & 78.0 & 88.0 & 84.0 & \textbf{90.0} \\
\rowcolor[HTML]{FFFFFF}
\parbox[t]{0.31\linewidth}{put the orange on the plate {\scriptsize\mbox{(pick L, place R)}}\smallskip} & 46.0 & 18.0 & 58.0 & 60.0 & 64.0 & 62.0 & \textbf{94.0} & 50.0 \\
\rowcolor[HTML]{F2F2F2}
\parbox[t]{0.31\linewidth}{put the tangerine on the cutting board {\scriptsize\mbox{(right arm)}}\smallskip} & 22.0 & 44.0 & 86.0 & 78.0 & 62.0 & 84.0 & 84.0 & \textbf{90.0} \\
\rowcolor[HTML]{FFFFFF}
\parbox[t]{0.31\linewidth}{put the tangerine on the tray {\scriptsize\mbox{(pick R, place L)}}\smallskip} & 38.0 & 26.0 & 66.0 & 44.0 & 64.0 & 76.0 & \textbf{88.0} & 30.0 \\
\rowcolor[HTML]{F2F2F2}
\parbox[t]{0.31\linewidth}{put the yogurt in the basket {\scriptsize\mbox{(pick R, place L)}}\smallskip} & 8.0 & 20.0 & 42.0 & 52.0 & 48.0 & 42.0 & \textbf{80.0} & 10.0 \\
\bottomrule
\end{longtable}
}

%% file: tables/bimesa_instance_per_task_success.tex
{\small\setlength{\tabcolsep}{3pt}
\begin{longtable}{lcccccccc}
\caption{Success rates for all tasks in the BiMESA-Instance task set.} \\
\toprule
\rowcolor[HTML]{FFDEB4} \textbf{BiMESA-Instance} & \textbf{DP} & \textbf{PG-Bin} & \textbf{PG-FM} & \textbf{GR00T-N1.6} & $\boldsymbol{\pi}\mathbf{_0}$ & \textbf{FAST} & $\boldsymbol{\pi}\mathbf{_{0.5}}$ & \makecell{\textbf{GPT-6.1-}\\\textbf{Sol}} \\
\midrule
\endfirsthead
\toprule
\rowcolor[HTML]{FFDEB4} \textbf{BiMESA-Instance} & \textbf{DP} & \textbf{PG-Bin} & \textbf{PG-FM} & \textbf{GR00T-N1.6} & $\boldsymbol{\pi}\mathbf{_0}$ & \textbf{FAST} & $\boldsymbol{\pi}\mathbf{_{0.5}}$ & \makecell{\textbf{GPT-6.1-}\\\textbf{Sol}} \\
\midrule
\endhead
\rowcolor[HTML]{FFFFFF}
\parbox[t]{0.31\linewidth}{put the apple in the bowl\smallskip} & 12.0 & 56.0 & 54.0 & 62.0 & 42.0 & 64.0 & \textbf{72.0} & 40.0 \\
\rowcolor[HTML]{F2F2F2}
\parbox[t]{0.31\linewidth}{put the apple in the box and close it\smallskip} & 0.0 & 0.0 & 4.0 & 14.0 & 6.0 & 28.0 & \textbf{52.0} & 0.0 \\
\rowcolor[HTML]{FFFFFF}
\parbox[t]{0.31\linewidth}{put the apple on the tray\smallskip} & 18.0 & 30.0 & 20.0 & 62.0 & 22.0 & 54.0 & 42.0 & \textbf{80.0} \\
\rowcolor[HTML]{F2F2F2}
\parbox[t]{0.31\linewidth}{put the beer on the plate\smallskip} & 12.0 & 2.0 & 20.0 & 52.0 & 4.0 & 40.0 & 54.0 & \textbf{60.0} \\
\rowcolor[HTML]{FFFFFF}
\parbox[t]{0.31\linewidth}{put the bell pepper in the basket\smallskip} & 10.0 & 6.0 & 28.0 & \textbf{36.0} & 14.0 & 32.0 & 34.0 & 20.0 \\
\rowcolor[HTML]{F2F2F2}
\parbox[t]{0.31\linewidth}{put the bell pepper on the tray\smallskip} & 6.0 & 16.0 & 38.0 & 38.0 & 26.0 & 34.0 & 48.0 & \textbf{80.0} \\
\rowcolor[HTML]{FFFFFF}
\parbox[t]{0.31\linewidth}{put the beer on the plate {\scriptsize\mbox{(handoff)}}\smallskip} & 4.0 & 4.0 & 18.0 & 26.0 & 20.0 & 10.0 & 38.0 & \textbf{80.0} \\
\rowcolor[HTML]{F2F2F2}
\parbox[t]{0.31\linewidth}{put the can in the pan {\scriptsize\mbox{(handoff)}}\smallskip} & 18.0 & 8.0 & 26.0 & 24.0 & 38.0 & 20.0 & \textbf{60.0} & 40.0 \\
\rowcolor[HTML]{FFFFFF}
\parbox[t]{0.31\linewidth}{put the candle on the tray {\scriptsize\mbox{(handoff)}}\smallskip} & 10.0 & 10.0 & 14.0 & 14.0 & 6.0 & 24.0 & 20.0 & \textbf{30.0} \\
\rowcolor[HTML]{F2F2F2}
\parbox[t]{0.31\linewidth}{put the jam in the right side of the dish drainer {\scriptsize\mbox{(handoff)}}\smallskip} & 0.0 & 4.0 & \textbf{24.0} & 4.0 & 8.0 & 6.0 & 10.0 & 0.0 \\
\rowcolor[HTML]{FFFFFF}
\parbox[t]{0.31\linewidth}{put the water bottle on the tray {\scriptsize\mbox{(handoff)}}\smallskip} & 10.0 & 4.0 & 26.0 & 30.0 & 36.0 & 20.0 & \textbf{62.0} & 40.0 \\
\rowcolor[HTML]{F2F2F2}
\parbox[t]{0.31\linewidth}{put the jam in the basket\smallskip} & 4.0 & 18.0 & 32.0 & \textbf{62.0} & 18.0 & 46.0 & \textbf{62.0} & 60.0 \\
\rowcolor[HTML]{FFFFFF}
\parbox[t]{0.31\linewidth}{put the mushroom in the bowl\smallskip} & 6.0 & 16.0 & 58.0 & 62.0 & 40.0 & 52.0 & \textbf{78.0} & 60.0 \\
\rowcolor[HTML]{F2F2F2}
\parbox[t]{0.31\linewidth}{open the box and put the apple in it\smallskip} & 2.0 & 6.0 & 6.0 & 8.0 & 12.0 & 30.0 & \textbf{32.0} & 0.0 \\
\rowcolor[HTML]{FFFFFF}
\parbox[t]{0.31\linewidth}{open the slide cabinet and put the beer in it\smallskip} & 0.0 & 0.0 & 0.0 & 14.0 & 18.0 & \textbf{34.0} & 26.0 & 30.0 \\
\rowcolor[HTML]{F2F2F2}
\parbox[t]{0.31\linewidth}{open the slide cabinet and put the orange in it and close it\smallskip} & 0.0 & 0.0 & 4.0 & 10.0 & 10.0 & \textbf{20.0} & 18.0 & \textbf{20.0} \\
\rowcolor[HTML]{FFFFFF}
\parbox[t]{0.31\linewidth}{put the orange on the cutting board\smallskip} & 26.0 & 34.0 & 80.0 & 70.0 & 60.0 & 76.0 & \textbf{84.0} & 60.0 \\
\rowcolor[HTML]{F2F2F2}
\parbox[t]{0.31\linewidth}{put the orange on the plate\smallskip} & 32.0 & 36.0 & 62.0 & 72.0 & 50.0 & 70.0 & 74.0 & \textbf{90.0} \\
\bottomrule
\end{longtable}
}

%% file: tables/bimesa_composite_per_task_success.tex
{\small\setlength{\tabcolsep}{3pt}
\begin{longtable}{lcccccccc}
\caption{Success rates for all tasks in the BiMESA-Composite task set.} \\
\toprule
\rowcolor[HTML]{FFDEB4} \textbf{BiMESA-Composite} & \textbf{DP} & \textbf{PG-Bin} & \textbf{PG-FM} & \textbf{GR00T-N1.6} & $\boldsymbol{\pi}\mathbf{_0}$ & \textbf{FAST} & $\boldsymbol{\pi}\mathbf{_{0.5}}$ & \makecell{\textbf{GPT-6.1-}\\\textbf{Sol}} \\
\midrule
\endfirsthead
\toprule
\rowcolor[HTML]{FFDEB4} \textbf{BiMESA-Composite} & \textbf{DP} & \textbf{PG-Bin} & \textbf{PG-FM} & \textbf{GR00T-N1.6} & $\boldsymbol{\pi}\mathbf{_0}$ & \textbf{FAST} & $\boldsymbol{\pi}\mathbf{_{0.5}}$ & \makecell{\textbf{GPT-6.1-}\\\textbf{Sol}} \\
\midrule
\endhead
\rowcolor[HTML]{FFFFFF}
\parbox[t]{0.31\linewidth}{put the apple on the plate\smallskip} & 32.0 & 62.0 & 76.0 & 78.0 & 74.0 & 76.0 & \textbf{90.0} & \textbf{90.0} \\
\rowcolor[HTML]{F2F2F2}
\parbox[t]{0.31\linewidth}{put the broccoli and the mushroom on the cutting board\smallskip} & 0.0 & 0.0 & 0.0 & 0.0 & 0.0 & 2.0 & 0.0 & \textbf{30.0} \\
\rowcolor[HTML]{FFFFFF}
\parbox[t]{0.31\linewidth}{put the carrot in the pan\smallskip} & 2.0 & 4.0 & 4.0 & 14.0 & 2.0 & 2.0 & 32.0 & \textbf{100.0} \\
\rowcolor[HTML]{F2F2F2}
\parbox[t]{0.31\linewidth}{put the fish and the garlic in the pan\smallskip} & 0.0 & 0.0 & 0.0 & 4.0 & 0.0 & 2.0 & 2.0 & \textbf{20.0} \\
\rowcolor[HTML]{FFFFFF}
\parbox[t]{0.31\linewidth}{put the lemon on the cutting board\smallskip} & 28.0 & 54.0 & 84.0 & 70.0 & 72.0 & 90.0 & 94.0 & \textbf{100.0} \\
\rowcolor[HTML]{F2F2F2}
\parbox[t]{0.31\linewidth}{put the lemon and the lime in the basket\smallskip} & 0.0 & 0.0 & 0.0 & 2.0 & 0.0 & 2.0 & 2.0 & \textbf{10.0} \\
\rowcolor[HTML]{FFFFFF}
\parbox[t]{0.31\linewidth}{put the mushroom on the tray\smallskip} & 6.0 & 38.0 & 40.0 & 64.0 & 32.0 & 72.0 & \textbf{82.0} & 40.0 \\
\rowcolor[HTML]{F2F2F2}
\parbox[t]{0.31\linewidth}{put the onion in the bowl\smallskip} & 18.0 & 58.0 & 60.0 & 74.0 & 52.0 & 74.0 & \textbf{92.0} & 40.0 \\
\rowcolor[HTML]{FFFFFF}
\parbox[t]{0.31\linewidth}{open the box and put the apple in it and close it\smallskip} & 0.0 & 0.0 & 0.0 & 4.0 & 0.0 & 16.0 & \textbf{28.0} & 10.0 \\
\rowcolor[HTML]{F2F2F2}
\parbox[t]{0.31\linewidth}{open the slide cabinet and put the beer in it and close it\smallskip} & 0.0 & 0.0 & 0.0 & 0.0 & 0.0 & 2.0 & 8.0 & \textbf{20.0} \\
\rowcolor[HTML]{FFFFFF}
\parbox[t]{0.31\linewidth}{open the box and put the beer in it\smallskip} & 0.0 & 0.0 & 4.0 & 0.0 & 6.0 & 14.0 & \textbf{24.0} & 10.0 \\
\rowcolor[HTML]{F2F2F2}
\parbox[t]{0.31\linewidth}{open the slide cabinet and put the book in it and close it\smallskip} & 0.0 & 0.0 & 0.0 & 0.0 & 0.0 & 0.0 & 0.0 & \textbf{10.0} \\
\rowcolor[HTML]{FFFFFF}
\parbox[t]{0.31\linewidth}{open the slide cabinet and put the jam in it\smallskip} & 0.0 & 0.0 & 0.0 & 16.0 & 32.0 & 48.0 & \textbf{70.0} & 20.0 \\
\rowcolor[HTML]{F2F2F2}
\parbox[t]{0.31\linewidth}{open the box and put the lemon in it and close it\smallskip} & 0.0 & 0.0 & 0.0 & 0.0 & 0.0 & \textbf{24.0} & 14.0 & 10.0 \\
\rowcolor[HTML]{FFFFFF}
\parbox[t]{0.31\linewidth}{open the slide cabinet and put the mushroom in it and close it\smallskip} & 0.0 & 0.0 & 0.0 & 6.0 & 6.0 & 8.0 & \textbf{24.0} & 10.0 \\
\bottomrule
\end{longtable}
}

%% file: tables/bimesa_category_per_task_success.tex
{\small\setlength{\tabcolsep}{3pt}
\begin{longtable}{lcccccccc}
\caption{Success rates for all tasks in the BiMESA-Category task set.} \\
\toprule
\rowcolor[HTML]{FFDEB4} \textbf{BiMESA-Category} & \textbf{DP} & \textbf{PG-Bin} & \textbf{PG-FM} & \textbf{GR00T-N1.6} & $\boldsymbol{\pi}\mathbf{_0}$ & \textbf{FAST} & $\boldsymbol{\pi}\mathbf{_{0.5}}$ & \makecell{\textbf{GPT-6.1-}\\\textbf{Sol}} \\
\midrule
\endfirsthead
\toprule
\rowcolor[HTML]{FFDEB4} \textbf{BiMESA-Category} & \textbf{DP} & \textbf{PG-Bin} & \textbf{PG-FM} & \textbf{GR00T-N1.6} & $\boldsymbol{\pi}\mathbf{_0}$ & \textbf{FAST} & $\boldsymbol{\pi}\mathbf{_{0.5}}$ & \makecell{\textbf{GPT-6.1-}\\\textbf{Sol}} \\
\midrule
\endhead
\rowcolor[HTML]{FFFFFF}
\parbox[t]{0.31\linewidth}{put the beet in the pot\smallskip} & 2.0 & 0.0 & 4.0 & 16.0 & 2.0 & 2.0 & 8.0 & \textbf{30.0} \\
\rowcolor[HTML]{F2F2F2}
\parbox[t]{0.31\linewidth}{put the brussel sprout in the pan\smallskip} & 20.0 & 14.0 & 18.0 & 42.0 & 6.0 & 24.0 & 14.0 & \textbf{100.0} \\
\rowcolor[HTML]{FFFFFF}
\parbox[t]{0.31\linewidth}{put the chili pepper in the bowl\smallskip} & 0.0 & 0.0 & 0.0 & 8.0 & 0.0 & 2.0 & 2.0 & \textbf{40.0} \\
\rowcolor[HTML]{F2F2F2}
\parbox[t]{0.31\linewidth}{put the cupcake in the basket\smallskip} & 2.0 & 4.0 & 26.0 & \textbf{32.0} & 10.0 & 20.0 & 26.0 & 10.0 \\
\rowcolor[HTML]{FFFFFF}
\parbox[t]{0.31\linewidth}{put the ginger on the cutting board\smallskip} & 6.0 & 20.0 & 14.0 & 24.0 & 8.0 & 28.0 & 10.0 & \textbf{90.0} \\
\rowcolor[HTML]{F2F2F2}
\parbox[t]{0.31\linewidth}{put the grapes in the basket\smallskip} & 2.0 & 0.0 & 6.0 & \textbf{24.0} & 0.0 & 16.0 & 12.0 & 20.0 \\
\rowcolor[HTML]{FFFFFF}
\parbox[t]{0.31\linewidth}{put the olive oil bottle on the tray {\scriptsize\mbox{(handoff)}}\smallskip} & 18.0 & 4.0 & 34.0 & 12.0 & 30.0 & 6.0 & 34.0 & \textbf{90.0} \\
\rowcolor[HTML]{F2F2F2}
\parbox[t]{0.31\linewidth}{put the pomegranate in the bowl {\scriptsize\mbox{(handoff)}}\smallskip} & 20.0 & 20.0 & 30.0 & 40.0 & 52.0 & 34.0 & 26.0 & \textbf{60.0} \\
\rowcolor[HTML]{FFFFFF}
\parbox[t]{0.31\linewidth}{put the thermos in the basket {\scriptsize\mbox{(handoff)}}\smallskip} & 4.0 & 14.0 & 18.0 & 30.0 & 20.0 & 6.0 & 12.0 & \textbf{70.0} \\
\rowcolor[HTML]{F2F2F2}
\parbox[t]{0.31\linewidth}{open the box and put the cupcake in it\smallskip} & 0.0 & 0.0 & 0.0 & 0.0 & 6.0 & 8.0 & \textbf{26.0} & 0.0 \\
\rowcolor[HTML]{FFFFFF}
\parbox[t]{0.31\linewidth}{open the box and put the grapes in it and close it\smallskip} & 0.0 & 0.0 & 0.0 & 0.0 & 0.0 & \textbf{4.0} & 0.0 & 0.0 \\
\rowcolor[HTML]{F2F2F2}
\parbox[t]{0.31\linewidth}{open the slide cabinet and put the pear in it\smallskip} & 0.0 & 4.0 & 6.0 & 38.0 & 20.0 & 44.0 & \textbf{52.0} & 40.0 \\
\rowcolor[HTML]{FFFFFF}
\parbox[t]{0.31\linewidth}{open the slide cabinet and put the potato in it\smallskip} & 0.0 & 0.0 & 0.0 & 40.0 & 4.0 & 38.0 & 28.0 & \textbf{60.0} \\
\rowcolor[HTML]{F2F2F2}
\parbox[t]{0.31\linewidth}{put the pear in the pot\smallskip} & 12.0 & 14.0 & 18.0 & 40.0 & 6.0 & 38.0 & 50.0 & \textbf{90.0} \\
\rowcolor[HTML]{FFFFFF}
\parbox[t]{0.31\linewidth}{put the radish on the cutting board\smallskip} & 0.0 & 0.0 & 4.0 & 34.0 & 0.0 & 22.0 & 6.0 & \textbf{70.0} \\
\rowcolor[HTML]{F2F2F2}
\parbox[t]{0.31\linewidth}{put the sausage on the baking sheet\smallskip} & 2.0 & 0.0 & 0.0 & 4.0 & 0.0 & 0.0 & 0.0 & \textbf{50.0} \\
\rowcolor[HTML]{FFFFFF}
\parbox[t]{0.31\linewidth}{put the scone on the plate\smallskip} & 10.0 & 8.0 & 8.0 & 60.0 & 16.0 & 44.0 & 10.0 & \textbf{70.0} \\
\bottomrule
\end{longtable}
}

%% file: 7.3-mesa-details.tex
\section{\abbr{} Details}
\label{sec:supp_task_skeletons}

\subsection{Background: BDDL}
\label{sec:supp_bddl_background}

The \textit{BEHAVIOR Domain Definition Language (BDDL)} \citep{McDermott1998PDDLthePD, srivastava2021behaviorbenchmarkeverydayhousehold} allows users to specify task initial and goal states using logical predicates, and has been used for several high-profile benchmarks \cite{liu2024libero, li2024behavior1khumancenteredembodiedai}.
More recently, \citet{saxena2025mimiclabs} adapts it to MimicGen, specifying the object-centric subtasks using demonstration predicates and enabling MimicGen-ready environment generation purely in the BDDL language.

\subsection{Implemented Task Skeletons}
We implement task skeletons as simple Python classes specifying task-relevant objects, initialization predicates, demonstration predicates, and goal predicates.
Our pipeline currently includes the \texttt{PickAndPlace} skeleton, which encompasses all pick-and-place tasks; \texttt{MultiStepArticulatedManipulation}, which encompasses tasks involving articulated objects, such as opening or closing them and placing objects in them; and, for the bimanual setting, \texttt{BimanualHandoff}.

\subsection{Simulation Platform}
We build on MimicLabs \cite{saxena2025mimiclabs}, which integrates infrastructure from robosuite \cite{robosuite2020}, MimicGen \cite{mandlekar2023mimicgen}, LIBERO \cite{liu2024libero} and RoboCasa \cite{nasiriany2024robocasa}, and uses the MuJoCo simulator \cite{todorov2012mujoco}.
For our single-arm Franka setup, we use the Franka Panda embodiment \cite{franka_panda} asset from robosuite with a Robotiq 2F-85 gripper whose asset was sourced from MuJoCo Menagerie \cite{menagerie2022github}. The dataset is designed to be a digital cousin of the DROID \cite{khazatsky2024droid} dataset.
Our environment includes left and right shoulder cameras and a wrist camera, but we only use the left shoulder and wrist perspectives for our experiments.
For our bimanual YAM setup, we use the YAM asset from MuJoCo Menagerie \cite{menagerie2022github} and include two wrist cameras and an egocentric camera. 

\begin{algorithm}[h!]
\caption{Distractor sampling with visual confusion guarantees.}
\label{alg:distractor_sampling}
\begin{algorithmic}[1]
\Require target distractor object counts $n_O,n_D$, task-relevant objects $o$, $d$, maps between objects and destinations $\mathcal{O}_d$, $\mathcal{D}_o$ \Ensure list of distractor object keys \vspace{2pt}

\State $\mathcal{O}_{\mathrm{placed}} \leftarrow \{o\}$, $\mathcal{D}_{\mathrm{placed}} \leftarrow \{d\}$

\While{$|\mathcal{O}_{\mathrm{placed}}|<n_O$ or $|\mathcal{D}_{\mathrm{placed}}|<n_D$} \If{$|\mathcal{O}_{\mathrm{placed}}|<n_O$} \State Sample object $o \in \bigcup_{d \in \mathcal{D}_{\mathrm{placed}}} \mathcal{O}_d \setminus \mathcal{O}_{\mathrm{placed}}$ \State $\mathcal{O}_{\mathrm{placed}} \leftarrow \mathcal{O}_{\mathrm{placed}} \cup \{o\}$ \EndIf \If{$|\mathcal{D}_{\mathrm{placed}}| <n_D$} \State Sample destination $d \in \bigcup_{o \in \mathcal{O}_{\mathrm{placed}}} \mathcal{D}_o \setminus \mathcal{D}_{\mathrm{placed}}$ \State $\mathcal{D}_{\mathrm{placed}} \leftarrow \mathcal{D}_{\mathrm{placed}} \cup \{d\}$ \EndIf
    
\EndWhile

\noindent \Return $\mathcal{O}_{\mathrm{placed}}, \mathcal{D}_{\mathrm{placed}}$

\end{algorithmic}
\end{algorithm}

\subsection{Ensuring Visual Confusion via Distractor Sampling}
\label{sec:supp_distractors}
This section expands on the distractor sampling procedure referenced in the main paper, which aims to maintain high observation-conditioned task entropy $H(\mathcal{T}|\mathcal{I})$ and reduce causal confusion.

Each scene contains a mixture of task-relevant objects and task-irrelevant distractor objects.
For example, for the task ``\texttt{put the apple in the basket}'', the task-relevant objects would be the apple and the basket, and the task-irrelevant objects might include objects such as a plate.
The goal is that for each initial state $i \in \mathcal{I}$ corresponding to a task $t \in \mathcal{T}$, there exists some other task $t' \in \mathcal{T}$, $t' \neq t$, such that the task-relevant objects for $t'$ appear in $i$.
To achieve this, we define $\mathcal{O}$ as the set of graspable objects (e.g., apples) and $\mathcal{D}$ to be the set of destination objects (e.g., baskets).
For each object $o \in \mathcal{O}$, we define a set $\mathcal{D}_o \subset \mathcal{D}$ as the set of destination objects in which $o$ is placed during some task in MESA-70. Similarly, for each destination $d \in \mathcal{D}$, we define $\mathcal{O}_d \subset \mathcal{O}$ to be the set of objects placed in $d$.
From here, we start by placing the task-relevant objects and then iteratively add more objects, sampling from the combined sets according to the placed items, until we have the desired number of distractor objects.
This process is summarized for pick and place tasks in \autoref{alg:distractor_sampling}, and we apply a similar algorithm to choose valid articulated objects.

\subsection{Example: \texttt{PickAndPlace} Skeleton}
\label{sec:supp_pick_and_place_skeleton}
This section provides a sample \texttt{PickAndPlace} skeleton.

\input{figures/skeleton_sample}

\subsection{Example: BDDL File}
\label{sec:example-bddl-file}

Here we provide a sample BDDL file for the task \texttt{``open the box and put the apple in it and close it''}.
Note that whereas prior work saves these in the \texttt{.bddl} file format, we found this to add unnecessary implementation complexity since they are ultimately parsed into Python dictionaries.
To avoid this, we simply save them as JSON files as we see below.
This BDDL file shows how we instantiate several distractor objects, including a bell pepper, tray, bottled water, pan, tomato and squash.
In \texttt{distractor_goal_specs} we see how the possible distractor tasks are enumerated.
In \texttt{demonstration_states} we see the predicates that are used when decomposing the task for MimicGen.

Notice in \texttt{regions} that the initialization regions all cover the same area of the table.
Unlike prior works such as LIBERO \cite{liu2024libero}, which constrains each object to a small region without inter-object overlap, we allow objects to spawn across the full table and filter out overlapping initial configurations.
To do this, we start by labeling each object with a 2D bounding box of its projection onto the table, which is already done for RoboCasa \cite{nasiriany2024robocasa} objects and we do manually for other objects.
Next, we sort objects by the sizes of their bounding boxes and iterate through them, randomly placing an object until it doesn't intersect with any already placed objects.

\input{figures/bddl_sample}

\subsection{Example: Initial State Distribution}

In \autoref{fig:init-states} we visualize 20 initialization states for the \texttt{``open the box and put the apple in it and close it''} task.
The top left image corresponds to the BDDL file shown in \autoref{sec:example-bddl-file}.
Notice that each one has a unique set of distractor objects, posing substantial vision-language grounding challenges.

\input{figures/init_states/init_states}

%% file: figures/skeleton_sample.tex
\begin{Verbatim}[commandchars=\\\{\},breaklines,fontsize=\small,numbers=left]

\PYG{k+kn}{from}\PYG{+w}{ }\PYG{n+nn}{dataclasses}\PYG{+w}{ }\PYG{k+kn}{import} \PYG{n}{dataclass}\PYG{p}{,} \PYG{n}{field}
\PYG{k+kn}{from}\PYG{+w}{ }\PYG{n+nn}{typing}\PYG{+w}{ }\PYG{k+kn}{import} \PYG{n}{List}\PYG{p}{,} \PYG{n}{Optional}\PYG{p}{,} \PYG{n}{Tuple}

\PYG{k+kn}{from}\PYG{+w}{ }\PYG{n+nn}{.objects}\PYG{+w}{ }\PYG{k+kn}{import} \PYG{p}{(}
    \PYG{n}{get\PYGZus{}grasp\PYGZus{}object\PYGZus{}spec}\PYG{p}{,}
    \PYG{n}{get\PYGZus{}dest\PYGZus{}object\PYGZus{}spec}\PYG{p}{,}
\PYG{p}{)}

\PYG{k+kn}{from}\PYG{+w}{ }\PYG{n+nn}{.task}\PYG{+w}{ }\PYG{k+kn}{import} \PYG{n}{BaseTask}\PYG{p}{,} \PYG{n}{BaseTaskConfig}\PYG{p}{,} \PYG{n}{register\PYGZus{}task}
\PYG{k+kn}{from}\PYG{+w}{ }\PYG{n+nn}{.utils}\PYG{+w}{ }\PYG{k+kn}{import} \PYG{p}{(}
    \PYG{n}{make\PYGZus{}named\PYGZus{}grasp\PYGZus{}object\PYGZus{}spec}\PYG{p}{,}
    \PYG{n}{make\PYGZus{}named\PYGZus{}dest\PYGZus{}object\PYGZus{}spec}\PYG{p}{,}
    \PYG{n}{NamedGraspObjectSpec}\PYG{p}{,}
    \PYG{n}{NamedDestObjectSpec}\PYG{p}{,}
    \PYG{n}{NamedArticulatedObjectSpec}\PYG{p}{,}
\PYG{p}{)}

\PYG{n+nd}{@dataclass}\PYG{p}{(}\PYG{n}{frozen}\PYG{o}{=}\PYG{k+kc}{True}\PYG{p}{,} \PYG{n}{kw\PYGZus{}only}\PYG{o}{=}\PYG{k+kc}{True}\PYG{p}{)}
\PYG{k}{class}\PYG{+w}{ }\PYG{n+nc}{PickAndPlaceTaskConfig}\PYG{p}{(}\PYG{n}{BaseTaskConfig}\PYG{p}{):}
\PYG{+w}{    }\PYG{l+s+sd}{\PYGZdq{}\PYGZdq{}\PYGZdq{}Configuration for generating pick\PYGZhy{}and\PYGZhy{}place task suites.\PYGZdq{}\PYGZdq{}\PYGZdq{}}

    \PYG{n}{dest\PYGZus{}object}\PYG{p}{:} \PYG{n+nb}{str}
    \PYG{n}{pick\PYGZus{}objects}\PYG{p}{:} \PYG{n+nb}{str} \PYG{o}{|} \PYG{n}{List}\PYG{p}{[}\PYG{n+nb}{str}\PYG{p}{]}
    \PYG{n}{dest\PYGZus{}region}\PYG{p}{:} \PYG{n+nb}{str}
    \PYG{n}{goal\PYGZus{}predicate}\PYG{p}{:} \PYG{n+nb}{str}
    \PYG{n}{class\PYGZus{}name}\PYG{p}{:} \PYG{n+nb}{str} \PYG{o}{=} \PYG{l+s+s2}{\PYGZdq{}PickAndPlaceTask\PYGZdq{}}

\PYG{n+nd}{@register\PYGZus{}task}\PYG{p}{(}\PYG{l+s+s2}{\PYGZdq{}pick\PYGZus{}and\PYGZus{}place\PYGZdq{}}\PYG{p}{,} \PYG{n}{config\PYGZus{}cls}\PYG{o}{=}\PYG{n}{PickAndPlaceTaskConfig}\PYG{p}{)}
\PYG{k}{class}\PYG{+w}{ }\PYG{n+nc}{PickAndPlaceTask}\PYG{p}{(}\PYG{n}{BaseTask}\PYG{p}{):}
    \PYG{k}{def}\PYG{+w}{ }\PYG{n+nf+fm}{\PYGZus{}\PYGZus{}init\PYGZus{}\PYGZus{}}\PYG{p}{(}\PYG{n+nb+bp}{self}\PYG{p}{,} \PYG{n}{config}\PYG{p}{:} \PYG{n}{PickAndPlaceTaskConfig}\PYG{p}{)} \PYG{o}{\PYGZhy{}\PYGZgt{}} \PYG{k+kc}{None}\PYG{p}{:}
        \PYG{n+nb}{super}\PYG{p}{()}\PYG{o}{.}\PYG{n+nf+fm}{\PYGZus{}\PYGZus{}init\PYGZus{}\PYGZus{}}\PYG{p}{(}\PYG{n}{config}\PYG{o}{=}\PYG{n}{config}\PYG{p}{)}

        \PYG{k}{if} \PYG{n+nb}{isinstance}\PYG{p}{(}\PYG{n}{config}\PYG{o}{.}\PYG{n}{pick\PYGZus{}objects}\PYG{p}{,} \PYG{n+nb}{str}\PYG{p}{):}
            \PYG{n}{pick\PYGZus{}objects} \PYG{o}{=} \PYG{p}{[}\PYG{n}{config}\PYG{o}{.}\PYG{n}{pick\PYGZus{}objects}\PYG{p}{]}
        \PYG{k}{else}\PYG{p}{:}
            \PYG{n}{pick\PYGZus{}objects} \PYG{o}{=} \PYG{n}{config}\PYG{o}{.}\PYG{n}{pick\PYGZus{}objects}
        \PYG{n}{pick\PYGZus{}objects} \PYG{o}{=} \PYG{p}{[}\PYG{n}{get\PYGZus{}grasp\PYGZus{}object\PYGZus{}spec}\PYG{p}{(}\PYG{n}{object\PYGZus{}name}\PYG{p}{)} \PYG{k}{for} \PYG{n}{object\PYGZus{}name} \PYG{o+ow}{in} \PYG{n}{pick\PYGZus{}objects}\PYG{p}{]}
        \PYG{n+nb+bp}{self}\PYG{o}{.}\PYG{n}{\PYGZus{}pick\PYGZus{}objects}\PYG{p}{:} \PYG{n}{List}\PYG{p}{[}\PYG{n}{NamedGraspObjectSpec}\PYG{p}{]} \PYG{o}{=} \PYG{p}{[]}
        \PYG{k}{for} \PYG{n}{i}\PYG{p}{,} \PYG{n}{spec} \PYG{o+ow}{in} \PYG{n+nb}{enumerate}\PYG{p}{(}\PYG{n}{pick\PYGZus{}objects}\PYG{p}{):}
            \PYG{n}{object\PYGZus{}name} \PYG{o}{=} \PYG{l+s+sa}{f}\PYG{l+s+s2}{\PYGZdq{}object\PYGZus{}}\PYG{l+s+si}{\PYGZob{}}\PYG{n}{i}\PYG{l+s+si}{\PYGZcb{}}\PYG{l+s+s2}{\PYGZdq{}}
            \PYG{n+nb+bp}{self}\PYG{o}{.}\PYG{n}{\PYGZus{}pick\PYGZus{}objects}\PYG{o}{.}\PYG{n}{append}\PYG{p}{(}\PYG{n}{make\PYGZus{}named\PYGZus{}grasp\PYGZus{}object\PYGZus{}spec}\PYG{p}{(}\PYG{n}{object\PYGZus{}name}\PYG{p}{,} \PYG{n}{spec}\PYG{p}{))}

        \PYG{n}{dest\PYGZus{}object} \PYG{o}{=} \PYG{n}{get\PYGZus{}dest\PYGZus{}object\PYGZus{}spec}\PYG{p}{(}\PYG{n}{config}\PYG{o}{.}\PYG{n}{dest\PYGZus{}object}\PYG{p}{)}
        \PYG{n+nb+bp}{self}\PYG{o}{.}\PYG{n}{\PYGZus{}dest\PYGZus{}object}\PYG{p}{:} \PYG{n}{NamedDestObjectSpec} \PYG{o}{=} \PYG{n}{make\PYGZus{}named\PYGZus{}dest\PYGZus{}object\PYGZus{}spec}\PYG{p}{(}\PYG{l+s+s2}{\PYGZdq{}dest\PYGZus{}object\PYGZdq{}}\PYG{p}{,} \PYG{n}{dest\PYGZus{}object}\PYG{p}{)}
        \PYG{n+nb+bp}{self}\PYG{o}{.}\PYG{n}{\PYGZus{}dest\PYGZus{}region} \PYG{o}{=} \PYG{n}{config}\PYG{o}{.}\PYG{n}{dest\PYGZus{}region}
        \PYG{n+nb+bp}{self}\PYG{o}{.}\PYG{n}{\PYGZus{}goal\PYGZus{}predicate} \PYG{o}{=} \PYG{n}{config}\PYG{o}{.}\PYG{n}{goal\PYGZus{}predicate}

    \PYG{k}{def}\PYG{+w}{ }\PYG{n+nf}{build\PYGZus{}task\PYGZus{}objects}\PYG{p}{(}
        \PYG{n+nb+bp}{self}
    \PYG{p}{)} \PYG{o}{\PYGZhy{}\PYGZgt{}} \PYG{n}{Tuple}\PYG{p}{[}\PYG{n}{List}\PYG{p}{[}\PYG{n}{NamedGraspObjectSpec}\PYG{p}{],} \PYG{n}{List}\PYG{p}{[}\PYG{n}{NamedDestObjectSpec}\PYG{p}{],} \PYG{n}{List}\PYG{p}{[}\PYG{n}{NamedArticulatedObjectSpec}\PYG{p}{]]:}
        \PYG{k}{return} \PYG{n+nb+bp}{self}\PYG{o}{.}\PYG{n}{\PYGZus{}pick\PYGZus{}objects}\PYG{p}{,} \PYG{p}{[}\PYG{n+nb+bp}{self}\PYG{o}{.}\PYG{n}{\PYGZus{}dest\PYGZus{}object}\PYG{p}{],} \PYG{p}{[]}

    \PYG{k}{def}\PYG{+w}{ }\PYG{n+nf}{build\PYGZus{}goal\PYGZus{}state}\PYG{p}{(}
        \PYG{n+nb+bp}{self}
    \PYG{p}{)} \PYG{o}{\PYGZhy{}\PYGZgt{}} \PYG{n}{List}\PYG{p}{[}\PYG{n+nb}{object}\PYG{p}{]:}
        \PYG{n}{goal\PYGZus{}state}\PYG{p}{:} \PYG{n}{List}\PYG{p}{[}\PYG{n+nb}{object}\PYG{p}{]} \PYG{o}{=} \PYG{p}{[}\PYG{l+s+s2}{\PYGZdq{}and\PYGZdq{}}\PYG{p}{]}
        \PYG{k}{for} \PYG{n}{pick\PYGZus{}object} \PYG{o+ow}{in} \PYG{n+nb+bp}{self}\PYG{o}{.}\PYG{n}{\PYGZus{}pick\PYGZus{}objects}\PYG{p}{:}
            \PYG{k}{if} \PYG{n+nb+bp}{self}\PYG{o}{.}\PYG{n}{\PYGZus{}goal\PYGZus{}predicate} \PYG{o}{==} \PYG{l+s+s2}{\PYGZdq{}on\PYGZdq{}}\PYG{p}{:}
                \PYG{n}{goal\PYGZus{}state}\PYG{o}{.}\PYG{n}{append}\PYG{p}{([}\PYG{l+s+s2}{\PYGZdq{}on\PYGZdq{}}\PYG{p}{,} \PYG{n}{pick\PYGZus{}object}\PYG{o}{.}\PYG{n}{id}\PYG{p}{,} \PYG{n+nb+bp}{self}\PYG{o}{.}\PYG{n}{\PYGZus{}dest\PYGZus{}object}\PYG{o}{.}\PYG{n}{id}\PYG{p}{])}
            \PYG{k}{else}\PYG{p}{:}
                \PYG{k}{assert} \PYG{n+nb+bp}{self}\PYG{o}{.}\PYG{n}{\PYGZus{}dest\PYGZus{}region} \PYG{o+ow}{is} \PYG{o+ow}{not} \PYG{k+kc}{None}
                \PYG{n}{contain\PYGZus{}region\PYGZus{}name} \PYG{o}{=} \PYG{l+s+sa}{f}\PYG{l+s+s2}{\PYGZdq{}}\PYG{l+s+si}{\PYGZob{}}\PYG{n+nb+bp}{self}\PYG{o}{.}\PYG{n}{\PYGZus{}dest\PYGZus{}object}\PYG{o}{.}\PYG{n}{id}\PYG{l+s+si}{\PYGZcb{}}\PYG{l+s+s2}{\PYGZus{}}\PYG{l+s+si}{\PYGZob{}}\PYG{n+nb+bp}{self}\PYG{o}{.}\PYG{n}{\PYGZus{}dest\PYGZus{}region}\PYG{l+s+si}{\PYGZcb{}}\PYG{l+s+s2}{\PYGZdq{}}
                \PYG{n}{goal\PYGZus{}state}\PYG{o}{.}\PYG{n}{append}\PYG{p}{([}\PYG{l+s+s2}{\PYGZdq{}in\PYGZdq{}}\PYG{p}{,} \PYG{n}{pick\PYGZus{}object}\PYG{o}{.}\PYG{n}{id}\PYG{p}{,} \PYG{n}{contain\PYGZus{}region\PYGZus{}name}\PYG{p}{])}
        \PYG{k}{return} \PYG{n}{goal\PYGZus{}state}

    \PYG{k}{def}\PYG{+w}{ }\PYG{n+nf}{build\PYGZus{}demonstration\PYGZus{}states}\PYG{p}{(}
        \PYG{n+nb+bp}{self}\PYG{p}{,}
    \PYG{p}{)} \PYG{o}{\PYGZhy{}\PYGZgt{}} \PYG{n}{List}\PYG{p}{[}\PYG{n+nb}{list}\PYG{p}{]:}
        \PYG{n}{steps}\PYG{p}{:} \PYG{n}{List}\PYG{p}{[}\PYG{n+nb}{list}\PYG{p}{]} \PYG{o}{=} \PYG{p}{[]}
        \PYG{k}{for} \PYG{n}{pick\PYGZus{}object} \PYG{o+ow}{in} \PYG{n+nb+bp}{self}\PYG{o}{.}\PYG{n}{\PYGZus{}pick\PYGZus{}objects}\PYG{p}{:}
            \PYG{n}{steps}\PYG{o}{.}\PYG{n}{append}\PYG{p}{([}\PYG{l+s+s2}{\PYGZdq{}grasp\PYGZdq{}}\PYG{p}{,} \PYG{n}{pick\PYGZus{}object}\PYG{o}{.}\PYG{n}{id}\PYG{p}{])}
            \PYG{k}{if} \PYG{n+nb+bp}{self}\PYG{o}{.}\PYG{n}{\PYGZus{}goal\PYGZus{}predicate} \PYG{o}{==} \PYG{l+s+s2}{\PYGZdq{}on\PYGZdq{}}\PYG{p}{:}
                \PYG{n}{steps}\PYG{o}{.}\PYG{n}{append}\PYG{p}{([}\PYG{l+s+s2}{\PYGZdq{}on\PYGZdq{}}\PYG{p}{,} \PYG{n}{pick\PYGZus{}object}\PYG{o}{.}\PYG{n}{id}\PYG{p}{,} \PYG{n+nb+bp}{self}\PYG{o}{.}\PYG{n}{\PYGZus{}dest\PYGZus{}object}\PYG{o}{.}\PYG{n}{id}\PYG{p}{])}
            \PYG{k}{else}\PYG{p}{:}
                \PYG{k}{assert} \PYG{n+nb+bp}{self}\PYG{o}{.}\PYG{n}{\PYGZus{}dest\PYGZus{}region} \PYG{o+ow}{is} \PYG{o+ow}{not} \PYG{k+kc}{None}
                \PYG{n}{contain\PYGZus{}region\PYGZus{}name} \PYG{o}{=} \PYG{l+s+sa}{f}\PYG{l+s+s2}{\PYGZdq{}}\PYG{l+s+si}{\PYGZob{}}\PYG{n+nb+bp}{self}\PYG{o}{.}\PYG{n}{\PYGZus{}dest\PYGZus{}object}\PYG{o}{.}\PYG{n}{id}\PYG{l+s+si}{\PYGZcb{}}\PYG{l+s+s2}{\PYGZus{}}\PYG{l+s+si}{\PYGZob{}}\PYG{n+nb+bp}{self}\PYG{o}{.}\PYG{n}{\PYGZus{}dest\PYGZus{}region}\PYG{l+s+si}{\PYGZcb{}}\PYG{l+s+s2}{\PYGZdq{}}
                \PYG{n}{steps}\PYG{o}{.}\PYG{n}{append}\PYG{p}{([}\PYG{l+s+s2}{\PYGZdq{}in\PYGZdq{}}\PYG{p}{,} \PYG{n}{pick\PYGZus{}object}\PYG{o}{.}\PYG{n}{id}\PYG{p}{,} \PYG{n}{contain\PYGZus{}region\PYGZus{}name}\PYG{p}{])}
        \PYG{k}{return} \PYG{n}{steps}

\end{Verbatim}

%% file: figures/bddl_sample.tex

%% file: figures/init_states/init_states.tex
\begin{figure}[t]
    \centering
    
    \includegraphics[width=0.23\linewidth]{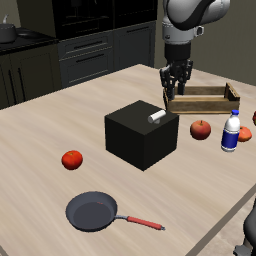}
    \includegraphics[width=0.23\linewidth]{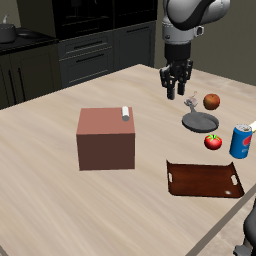}
    \includegraphics[width=0.23\linewidth]{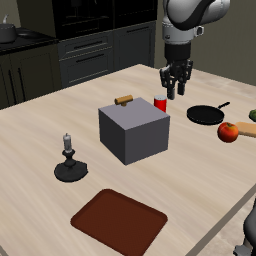}
    \includegraphics[width=0.23\linewidth]{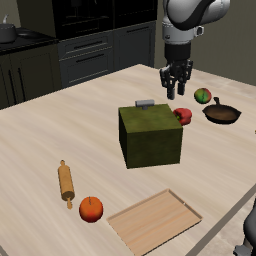}
    
    \medskip
    
    \includegraphics[width=0.23\linewidth]{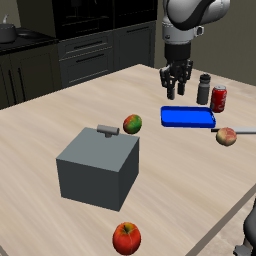}
    \includegraphics[width=0.23\linewidth]{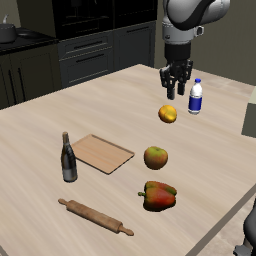}
    \includegraphics[width=0.23\linewidth]{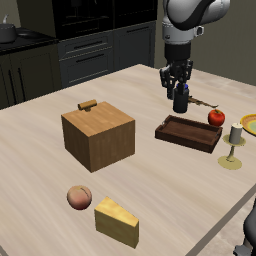}
    \includegraphics[width=0.23\linewidth]{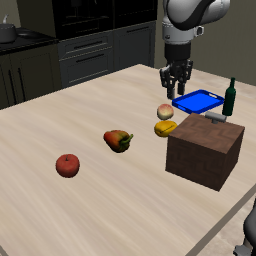}
    
    \medskip
    
    \includegraphics[width=0.23\linewidth]{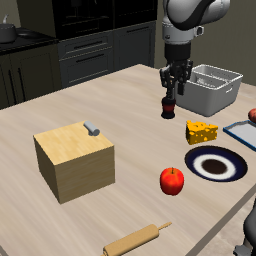}
    \includegraphics[width=0.23\linewidth]{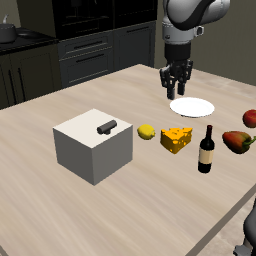}
    \includegraphics[width=0.23\linewidth]{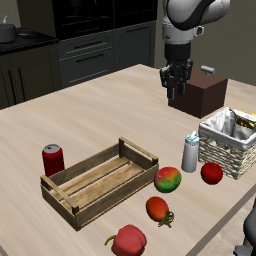}
    \includegraphics[width=0.23\linewidth]{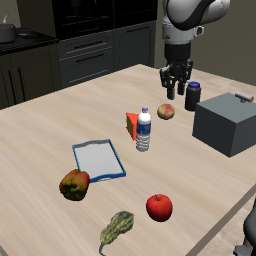}
    
    \medskip
    
    \includegraphics[width=0.23\linewidth]{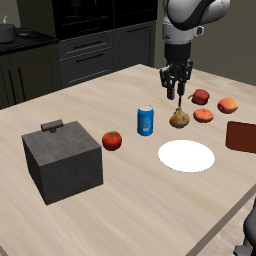}
    \includegraphics[width=0.23\linewidth]{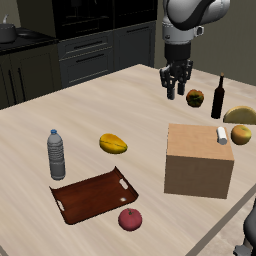}
    \includegraphics[width=0.23\linewidth]{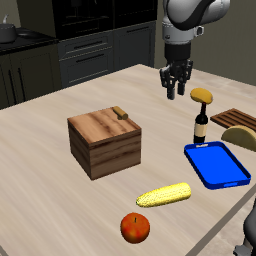}
    \includegraphics[width=0.23\linewidth]{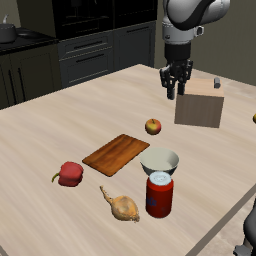}
    
    \medskip
    
    \includegraphics[width=0.23\linewidth]{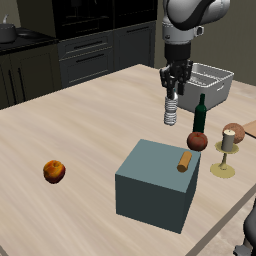}
    \includegraphics[width=0.23\linewidth]{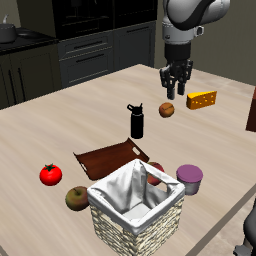}
    \includegraphics[width=0.23\linewidth]{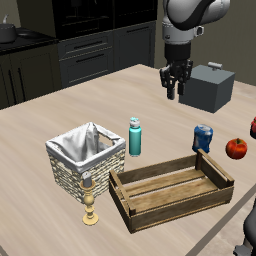}
    \includegraphics[width=0.23\linewidth]{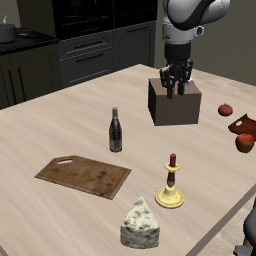}
    
    \caption{We visualize 20 initialization states for the \texttt{``open the box and put the apple in it and close it''} task.}
    \label{fig:init-states}
\end{figure}

%% file: 7.4-data-gen-details.tex
\section{Data Generation Details}
\label{sec:supp_mesa_gen}
\label{sec:mimicgen}

In this section, we describe \abbr{}-Gen's pipeline for scalable data generation with substantial variety in tasks, objects and initial configurations.
We first provide background on MimicGen in \autoref{sec:background}, then describe our protocol for collecting human source demonstrations in \autoref{sec:supp_source_demos}, then describe our method for enabling data generation with unseen tasks via subtask stitching in \autoref{sec:supp_subtask_stitching}, and finally describe our trajectory optimizations for handling large source-target object pose differences in \autoref{sec:supp_traj_opt}.

\subsection{Background: MimicGen}
\label{sec:background}

\textit{MimicGen} \citep{mandlekar2023mimicgen} is a framework for generating large robot demonstration datasets from a small set of human-collected seed trajectories.
The core idea is to decompose a demonstration trajectory $\tau = \{(T_o(t),T_e(t))\}_{t=1}^N$ into object-centric segments where $T_o(t)\in SE(3)$ and $T_e(t)\in SE(3)$ denote the object and end-effector poses, respectively.
For each segment, MimicGen computes the relative transform $T_{\mathrm{rel}}(t) = T_o(t)^{-1}\,T_e(t),$ which decouples motion from the original global frame.
Given a new target object pose $\tilde{T}_o(t)$, the corresponding adapted end-effector pose is reconstructed as $\tilde{T}_e(t)=\tilde{T}_o(t)\,T_{\mathrm{rel}}(t).$ By stitching these transformed segments together in their original temporal order, MimicGen generates new demonstrations that are physically consistent in novel contexts.

\subsection{Source Demonstration Collection Protocol}
\label{sec:supp_source_demos}
We collect human demonstrations for source tasks to seed MimicGen and subtask stitching.
For the single-arm setting, we collect 1125 demonstrations across 80 tasks using a Meta Quest VR setup \cite{OrbikEbert2021OculusReader}.
During data collection, control commands are executed at 20 Hz.
For the Quest controller, translational and rotational inputs are normalized to $\pm 1.0$ along axes and passed directly to the controller without action clipping, using a gain of 1.5. Source tasks are selected to broadly cover the objects and predicates supported in the benchmark.
We filter tasks by a data generation rate cutoff of $10\%$ after collecting 5 demonstrations per task.
The 2,152 teleoperated BiMESA demonstrations (cross-arm pick-and-place and handoffs; \appref{sec:supp_bench_construction}) are collected with the same Meta Quest hardware.

\subsection{Subtask Stitching for Unseen Tasks}
\label{sec:supp_subtask_stitching}
\label{sec:piecewise}
While MimicGen is built to enable scalable data generation, it still requires human data collection for each task.
This requires linear human effort in the number of created tasks, which is cumbersome for researchers with limited resources.
To alleviate this, we use subtask stitching which, instead of using only demonstrations from the target task, uses \abbr{}'s predicate structure to enable data generation for unseen tasks by composing object-centric subtasks sourced from other tasks with compatible predicates.
For example, for a target task ``\texttt{pick up the apple and put it on the plate}'', \abbr{}-Gen might combine the \texttt{Grasp(apple)} subtask from ``\texttt{put the apple in the drawer}'' with the \texttt{On(banana, plate)} subtask of ``\texttt{put the banana on the plate}''.

\medskip \noindent \textbf{Demonstration Lookup System:} We construct a hierarchical index offline by parsing all demonstration HDF5 files and extracting subtask-level metadata.
This index is organized primarily by predicate type (e.g., \textit{grasp, in, on, open, close, stack, turnon, turnoff}), and further refined by using object-centric semantic attributes such as shape category, object category, insertion type, destination region, and object identity.
The exact hierarchy depth and key structure depend on the predicate semantics; for example, \textit{grasp} predicates are indexed by the grasped object's shape and category, whereas \textit{in} predicates additionally encode both the source object and destination container properties.

Each leaf node of the index contains a list of concrete subtask segments extracted from demonstrations, including the path to the source HDF5 file, the demonstration identifier, object references involved in the interaction, the subtask's index within the original trajectory, and the termination signal defining the subtask boundary.
This is sufficient to recover the full state-action segment corresponding to the subtask and to re-contextualize it for a new task.

At generation time, we query this index using a predicate-specific set of search keys from the current subtask specification.
These keys are computed using object metadata functions that map object instances to semantic categories like shape class or insertion type.
Retrieval is performed by a recursive tree walk that follows the exact key if present.
If a key is not present, we take the union across all children at that depth, which broadens the search to semantically related subtasks.

From the retrieved candidate list, we select using either \textit{RandomStrategy} or one of several nearest-neighbor strategies: \textit{NearestNeighborObjectStrategy} (weighted object pose distance), \textit{NearestNeighborRobotDistanceStrategy} (robot travel distance to the first transformed pose), or \textit{NearestNeighborEefObjectTranslationStrategy} (relative end-effector-object translation).
Nearest-neighbor selection computes distances for all candidates and samples uniformly among the top-$k$ closest (default $k=3$).

\medskip \noindent \textbf{Stitching Procedure:} Each subtask is first trimmed from its source demonstration using start and end indices derived from predicate-specific termination signals stored in the HDF5 logs.
For non-final subtasks, boundaries are detected via the first valid rising edge of the corresponding terminal signal following the previous subtask; the final subtask extends to the end of the demonstration.
To increase diversity, subtask end boundaries may be randomly offset within a task-specified range, except for the final subtask, which is fixed to prevent truncation.

When concatenating subtasks, the system does not resample trajectories globally but instead inserts short transition segments to ensure temporal continuity.
Transitions optionally insert i) a linear interpolation segment that smoothly moves the end-effector from the final pose of the previous subtask to the initial pose of the next, and ii) a fix-pose hold segment that maintains the first pose of the next subtask for a fixed number of steps.
To avoid duplicate commands, the first waypoint of each appended segment is removed prior to concatenation.
After all subtasks are merged, their actions are concatenated into a single action sequence.

The environment is reset once at the beginning of the stitched trajectory, and all subsequent subtasks execute sequentially without intermediate resets.
We transform subtask trajectories to the current scene by preserving the relative pose between the end-effector and the reference object.
When objects differ in canonical rotation axes, an additional rotation correction is applied to align coordinate conventions.

Demonstrations with missing or invalid subtask termination signals are skipped during dataset parsing.
We enforce non-empty subtasks, strictly increasing subtask indices, and valid ordering of subtasks.
On generation, if no compatible subtasks are found even after using the broader semantic fallbacks mentioned previously, generation terminates.

\subsection{Trajectory Optimizations for Large Pose Differences}
\label{sec:supp_traj_opt}
\label{sec:start-opt}

As described in \autoref{sec:background}, MimicGen works by applying a rigid transformation based on the target object pose in the source and current task.
This is effective when source and target object poses are relatively close to one another and the target object is quasistatically stable, but can lead to unpredictable behaviors when either of these assumptions is broken.
In this subsection, we outline changes to the MimicGen pipeline leading to behaviors that quantitatively lead to higher data generation rates and qualitatively lead to more sensible behaviors.

\begin{figure}
    \centering
    \includegraphics[width=\linewidth]{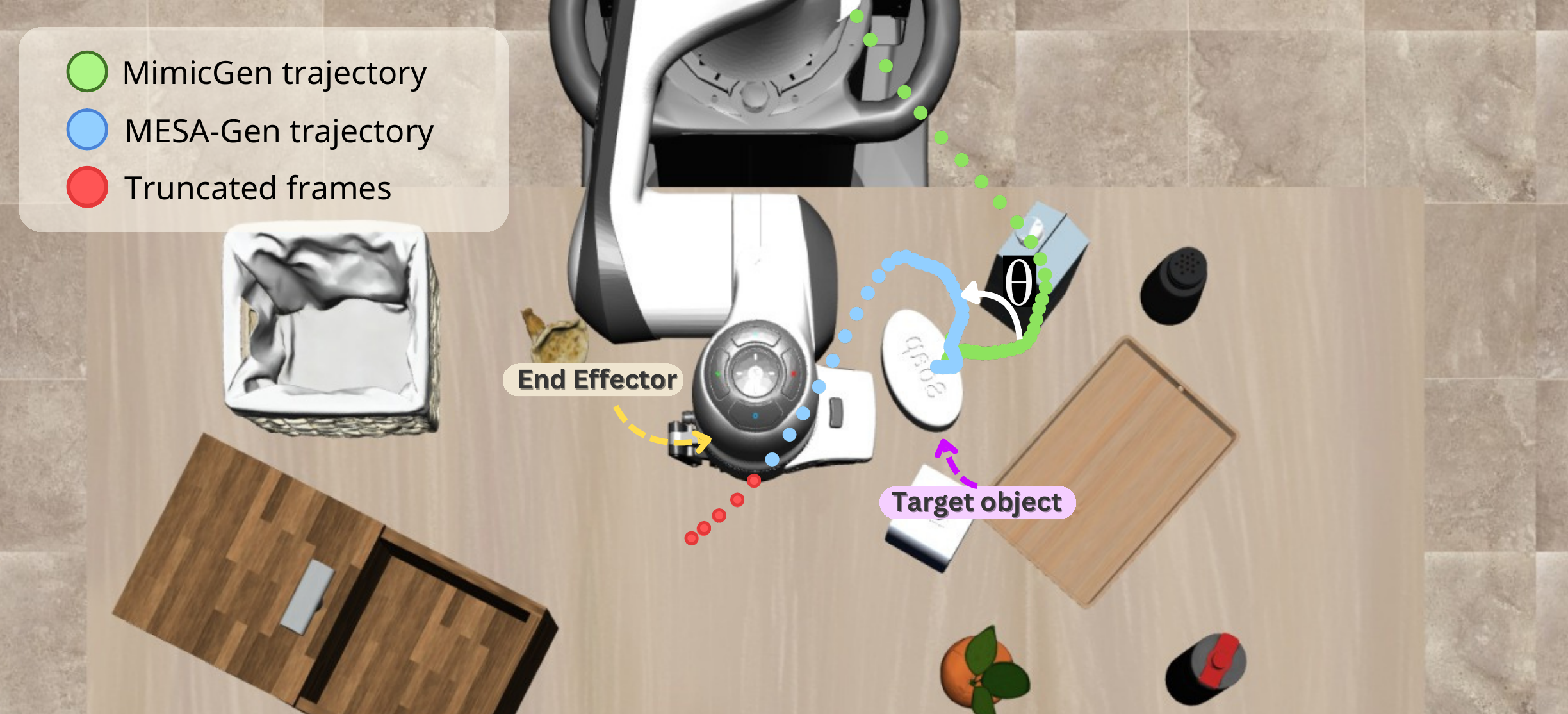}
    \caption{Due to the large change in object pose, the original MimicGen trajectory ({\color{LimeGreen} $\CIRCLE$}) starts from a position near the robot base, leading to a large interpolation and self-collision. We optimize the trajectory ({\color{SkyBlue} $\CIRCLE$}) by rotating it about the vertical axis through the object position and truncating unnecessary frames ({\color{Red}$\CIRCLE$}) to minimize the Euclidean distance from the end-effector to a point along the trajectory.}
    \label{fig:start-opt}
\end{figure}

\medskip \noindent \textbf{Removing Roll and Pitch in MimicGen Transforms:} MimicGen can produce unstable trajectories for objects without stable resting poses.
We observe that most tabletop objects either have effectively fixed roll and pitch under our initialization distribution or admit roll- and pitch-invariant grasps, so we ignore roll and pitch when computing MimicGen transformations.
Concretely, given an input rotation $R \in SO(3)$, we extract static XYZ Euler angles $(\phi, \theta, \psi)$ (roll, pitch, yaw) via \texttt{mat2euler(R, axes="sxyz")}, and reconstruct a pure-yaw rotation by setting roll and pitch to zero.
We enable this when grasping or placing objects and ignore it for other predicates (for example opening or closing articulated objects).
Additionally, we ignore yaw for \textit{on} predicates, or when interacting with objects rotationally symmetric about the vertical axis.
These include, for example, bottles, cans and apples.

\medskip \noindent \textbf{Trajectory Start Optimization:}
\label{sec:supp_start_opt_math}
Another issue with MimicGen arises when there is a large change between the source and target object poses, sometimes leading to unnatural behaviors.
We show an example of this in \autoref{fig:start-opt}, where the source trajectory comes from a configuration where the soap was farther away from the robot, leading to a trajectory beginning with a large unnecessary interpolation and self-collision.
In the best case this is unnatural and suboptimal; in the worst case it leads to dangerous behaviors and kinematic singularities.
To address this, we optimize the trajectory by rotating it about the vertical axis going through the target object to minimize the $L_2$ distance between the current end-effector pose and the start of the trajectory, and then truncate any additional frames so that the starting frame is the closest to the end-effector.

Let $\{\tau_i\}_0^N$ be a sequence of end-effector poses output by MimicGen, and let $T_o(0)$ and $T_e(0)$ be the poses of the target object and end-effector at the beginning of the rollout.
For each pose $\tau_i$, we solve for an angle $\theta_i \in (-\pi,\pi]$ defining a rotation about the vertical axis through $T_o(0)$ that best aligns the rotated pose with the current end-effector position:
\begin{equation}
\tilde{\tau}_i(\theta) \;=\; T_o(0)\,R_z(\theta)\,T_o(0)^{-1}\,\tau_i,
\end{equation}

\begin{equation}
\theta_i \;=\; \arg\min_{\theta \in (-\pi,\pi]} 
\left\| p\!\left(\tilde{\tau}_i(\theta)\right) - p\!\left(T_e(0)\right) \right\|_2,
\end{equation}
where $R_z(\theta)$ denotes a rotation about the world vertical axis and $p(\cdot)$ extracts the translational component of an $SE(3)$ transform.
We then select the starting index as the waypoint from the first half of the trajectory that can be made closest to the current end-effector position under such a rotation:
\begin{equation}
i^\star \;=\; \arg\min_{i \in \{0,\dots,\frac{N}{2}\}} \min_{\theta_i \in (-\pi,\pi]} 
\left\| p\!\left(\tilde{\tau}_i(\theta_i)\right) - p\!\left(T_e(0)\right) \right\|_2,
\end{equation}
and execute the rotated, truncated trajectory $\{\tilde{\tau}_i(\theta_{i^\star})\}_{i=i^\star}^{N}$.
This ensures execution begins from the point along the optimally rotated trajectory that is closest to the current end-effector pose, avoiding large initial deviations.

While this operation leads to more natural trajectory starting points, it may also lead to incorrect grasps for objects where end-effector yaw is important, while producing natural motions without modification for objects where it is not.
To alleviate this, we use an LLM to label objects based on whether they are geometrically symmetric under yaw rotation, and slerp end-effector rotation back to the original MimicGen output when they are not.

The next issue is that the current end-effector rotation is sometimes far from the rotation at the beginning of the MimicGen trajectory, even after optimizing the starting pose, causing unnecessary rapid rotations at the beginning of the subtask.
To alleviate this, we compute an offset rotation $R_{\mathrm{offset}}$ aligning the beginning of the trajectory with the current end-effector rotation.
We then compute a sequence of offsets slerping from $R_{\mathrm{offset}}$ to the identity and left multiply these by the end-effector orientations in our trajectory.

%% file: 7.5-bench-details.tex
\section{MESA-Bench Details}
\label{sec:supp_bench_lists}

All \abbr{} tasks are listed in \autoref{sec:per-task-sr}.

\subsection{Benchmark Construction}
\label{sec:supp_bench_construction}

We use the \abbr{}-Gen pipeline (\autoref{sec:mesa}) to generate hundreds of tasks for \abbr{}-Bench.
We define a handful of task skeletons such as \texttt{PickAndPlace}, \texttt{MultiStepArticulatedManipulation}, and (for the bimanual setting) \texttt{BimanualHandoff}, which requires picking with one arm and placing with the other.
We use GPT-5.1 to label each object with container compatibility (e.g., it is reasonable to place an apple but not a bottle of wine in a bowl) and generate BDDL files for all compatible task instantiations.

We then collect human source demonstrations covering a wide range of subtasks, run \abbr{}-Gen on the full task set, and iterate on subtasks where data generation succeeds at a low rate.
After iterating, we collect 1125 human demonstrations across 80 single-arm tasks.
We then keep all tasks with a data generation rate above $10\%$, yielding 100 demonstrations each for 806 single-arm tasks.
A potential concern is that generated data is biased toward regions with high data generation success rates; we study this in \appref{sec:supp_bias_study}.

The bimanual dataset, BiMESA-57, consists of a mixture of MimicGen and manually
collected data. \abbr{}-Gen can automatically synthesize spatial variants of a
source demonstration whenever the manipulation is carried out by a \emph{single}
arm, but cross-arm motions---a pick with one arm placed by the other, or a
hand-to-hand handoff---require coordinated bimanual control that the single-arm
trajectory transform cannot reproduce. We therefore generate single-arm task
variants with MimicGen and teleoperate the bimanual ones, then merge both sources
into one dataset of 6{,}833 demonstrations across 57 tasks.

We organize the tasks into three groups. \textbf{Spatial pick-and-place} (27 tasks,
4{,}690 demos) instantiates each \texttt{PickAndPlace} task under four side
configurations defined by which arm picks and on which side the destination lies:
the two \emph{aligned} configurations (one arm both picks and places) are
self-contained single-arm motions and are generated with MimicGen, while the two
\emph{cross} configurations require coordinated two-arm motion and are
teleoperated. For each task we hold out one configuration for evaluation and train
on the remaining three. \textbf{Handoff} (17 tasks, 843 demos) covers
\texttt{BimanualHandoff} tasks, which are inherently cross-arm and thus collected
entirely by teleoperation (48--50 demos per task); we train a single handoff direction
and reserve the reverse direction for evaluation. \textbf{Single-arm articulated
manipulation} (13 tasks, 1{,}300 demos) covers
\texttt{MultiStepArticulatedManipulation} tasks---opening, placing into, and
closing sliding cabinets and boxes---which a single arm performs end to end and are
generated with MimicGen (100 demos per task, balanced 50/50 across left- and
right-arm execution). In total, 4{,}681 demonstrations are MimicGen-generated and
2{,}152 are teleoperated for the BiMESA set.

\subsection{Single-Arm Suite Construction}

Here we list details about all of our single-arm task suites.

\medskip \noindent \textbf{\abbr{}-70}

This evaluation set is composed of the 70 tasks from the \abbr{}-70 training dataset.
Notably, it includes in-distribution variants of all the tasks from \abbr{}-Spatial and \abbr{}-Instance.

\medskip \noindent \textbf{\abbr{}-Spatial}

By default, $x$ coordinates are sampled from $\mathrm{Uniform}(0.05, 0.4)$ and $y$ from $\mathrm{Uniform}(-0.4, 0.4)$, which we empirically found to be the subset of the table where the robot could reliably interact with objects.
We partition the table into two regions, $\mathcal{R}_{\mathrm{ID}}$ for the training set and \abbr{}-70 evaluation set, and $\mathcal{R}_{\mathrm{OOD}}$ for the \abbr{}-Spatial out-of-distribution evaluation set.
For $\mathcal{R}_{\mathrm{ID}}$ we instead sample $y$ coordinates from $\mathrm{Uniform}(-0.4, 0.0)$ and for $\mathcal{R}_{\mathrm{OOD}}$ we sample them from $\mathrm{Uniform}(0.0, 0.4)$.

\medskip \noindent \textbf{\abbr{}-Instance}

For each task, we partition the set of object assets into disjoint in-distribution and out-of-distribution subsets, training exclusively on the in-distribution assets and evaluating on held-out instances at test time.
As with MESA-Spatial, the in-distribution asset variants are included in MESA-70, enabling direct comparison under changes in object appearance and geometry.

For three objects, namely the peach, squash and wine bottle, we hold out some assets from the RoboCasa-objaverse object set.
For the rest, we replace the original object instance with RoboCasa AI-generated assets, which are never seen during training.

\medskip \noindent \textbf{\abbr{}-Composite}

These are unseen composite tasks composed of subtasks seen in the \abbr{}-70 dataset.

We include 12 tasks with familiar structure.
Some are easy, such as \texttt{``put the apple on the plate''}, which composes skills from seen tasks \texttt{``put the apple on the tray''} and \texttt{``put the cheese on the plate''}.
Others have familiar structure but are designed adversarially.
For example, \texttt{``put the cup in the left side of the dish drainer''} is similar to the seen task \texttt{``put the cup in the right side of the dish drainer''}, but to complete it correctly the policy must notice this small change in task description and adjust the placement accordingly.

We include 8 tasks with unfamiliar structure.
Four of these are two-step pick and place tasks, such as \texttt{``put the fish and garlic in the pan''}, which are unfamiliar since the training data only includes one-step pick and place tasks such as \texttt{``put the fish in the pan''}.
The other four are articulated manipulation tasks involving opening objects, placing things in them, and then closing them.
These are unfamiliar since the training tasks only opened or closed objects, but never both.

\medskip \noindent \textbf{\abbr{}-Category}

These tasks all involve unseen grasp objects sourced from the RoboCasa AI-generated assets.
In addition, some include the \texttt{pot} and \texttt{baking sheet} placement locations, which are also unseen.
For example, for the task \texttt{``put the beet in the pot''}, the policy has never seen a beet or pot during finetuning.

\subsection{Bimanual Suite Construction}
\label{sec:supp_bimanual_suites}

The bimanual setting uses a YAM-style setup. Tasks fall into three families: bimanual pick-and-place (objects spawn on either side of the table; the policy chooses which arm picks and which arm places), single-arm articulated manipulation (one arm opens/closes a cabinet or sliding-top box and the same arm picks and places), and bimanual handoffs (one arm picks the object and hands it to the other arm, which places it).

\medskip \noindent \textbf{BiMESA-57}

This evaluation set is composed of the 57 tasks from the BiMESA-57 training dataset, evaluated on held-out initial-state seeds drawn from each task's evaluation split.
It includes the in-distribution variants of all tasks in BiMESA-Spatial.

\medskip \noindent \textbf{BiMESA-Spatial}

All 38 BiMESA-Spatial tasks require the policy to use a different arm than was used during training.
Because two arms are available, relocating the scene changes \emph{which arm must act and how the arms coordinate}, rather than simply where a single end-effector must reach---a form of spatial generalization with no single-arm analogue.
\begin{itemize}
    \item \textit{Pick-and-place held-out side combination} (27 tasks): each task is trained with three of four combinations of $\{\text{pick-left}, \text{pick-right}\} \times \{\text{dest-left}, \text{dest-right}\}$ and evaluated on the held-out fourth combination.
    \item \textit{Handoff direction reversal} (8 tasks): the policy is trained on handoffs in one direction (e.g., pick-left then place-right) and evaluated in the opposite direction. Four tasks are evaluated on unseen left-to-right handoffs and four on unseen right-to-left handoffs.
    \item \textit{Articulated fixture-side misalignment} (3 tasks): training pairs the pick-side and fixture-side (e.g., pick with the left arm while operating a left-side cabinet); evaluation places the fixture on a different side, requiring cross-body coordination.
\end{itemize}
\medskip \noindent \textbf{BiMESA-Instance}

For 18 tasks (9 pick-and-place, 5 handoff, 4 articulated), the pick object or destination is replaced with an AI-generated variant (suffixed \texttt{\_ai}) at evaluation time.
The in-distribution variants of these objects appear in BiMESA-57, enabling direct comparison.

\medskip \noindent \textbf{BiMESA-Composite}

These 15 unseen tasks compose subtasks seen during training:
\begin{itemize}
    \item 5 pick-and-place pairs where the individual pick and destination objects were seen, but never together.
    \item 3 two-step pick-and-place tasks that place two seen objects into a shared destination (e.g., \texttt{``put the lemon and lime in the basket''}).
    \item 4 unseen open-and-place pairs and 3 longer-horizon open-place-close sequences with unseen target objects, drawing on seen open/close skills.
\end{itemize}

\medskip \noindent \textbf{BiMESA-Category}

These 17 tasks involve pick objects (and sometimes destinations) drawn from categories never seen during training, sourced primarily from RoboCasa AI-generated assets.
They span 10 pick-and-place, 3 handoff, and 4 articulated manipulation tasks.

\subsection{Training and Auxiliary Datasets}
\label{sec:app-dataset-detail}

Our main \abbr{}-70 training set includes 7,000 demonstrations across the 70 tasks in \abbr{}-70, which is about 1M frames.
We also include an auxiliary dataset of 100 demonstrations per task for 736 tasks, and train on a subset of this with 10 demonstrations per task, which also has about 1M frames.
The bimanual training set, BiMESA-57, contains 6,833 demonstrations across the 57 in-distribution bimanual tasks (roughly 120 per task on average).

\subsection{Evaluation Details}

To minimize bias caused by unlucky starting configurations, we generate a fixed set of starting configurations.
We use these configurations for all of our experiments and include them in our open-source code release.
Within each embodiment and task, rollout $i$ restores the same object set and starting configuration for each method, ensuring comparisons use identical starting conditions.
When generating these starting configurations, we ensure that the target object is not occluded or out of frame and that the target task has not already been completed.

We evaluate each policy for 50 rollouts for each task, with each rollout consisting of a different set of distractor objects. 
Note that since the distractor objects and initial positions are sampled completely independently of the training configurations, \textit{every single initial observation in the benchmark is unique}, and solving this benchmark through memorization alone is infeasible.

\subsection{Statistical Uncertainty}
\label{sec:uncertainty}

With 50 rollouts per task, each finetuned method receives 3,500 / 500 / 1,000 / 1,000 / 1,000 rollouts on \abbr{}-ID / Spatial / Instance / Composite / Category and 2,850 / 1,900 / 900 / 750 / 850 on the corresponding BiMESA suites; GPT-6.1-Sol receives 570 / 380 / 180 / 150 / 170 (10 per task).
We report 95\% Wilson intervals on the pooled success counts of each suite (\autoref{tab:main-results-ci} and \autoref{tab:main-results-bimesa-ci}) and on the successful and distractor-completing rollouts for the language following rate (\autoref{tab:language-following-ci} and \autoref{tab:language-following-bimesa-ci}).
Suite averages use a normal interval from per-task binomial variances.

Since rollout $i$ uses the same initial state for every method, we compare finetuned methods with an exact McNemar test on paired rollouts, uncorrected for multiple comparisons, and call a difference significant at $p<0.05$.
On \abbr{}-ID, all five adjacent pairs in the ranking differ significantly, and $\pi_{0.5}$ is significantly ahead of every other method on all five single-arm suites.
On BiMESA-ID, 4 of 6 adjacent pairs differ significantly, and $\pi_{0.5}$ is significantly ahead on every BiMESA suite except Category, where GR00T-N1.6 is.
OOD rankings are less certain: 5 of 20 single-arm and 6 of 24 bimanual adjacent pairs cannot be separated.
Most of these are near-ties within 1.6 points; the other two are limited by suite size, as the \abbr{}-Composite and Category intervals are about twice as wide as those of \abbr{}-ID.

These intervals and tests capture only rollout variation for a single trained checkpoint per method, not variation across training seeds or task selection.

\input{tables/uncertainty}

%% file: tables/uncertainty.tex
\providecommand{\ci}[2]{\,{\tiny[#1,\,#2]}}

\begin{table}[h]
\centering
\scriptsize
\begin{adjustbox}{max width=\linewidth}
\begin{tabular}{lcccccc}
\toprule
\rowcolor[HTML]{FFDEB4}
Setting & DP & PG-FM & GR00T-N1.6 & $\pi_0$ & FAST & $\pi_{0.5}$ \\
\midrule
MESA-ID & 1.6\ci{1.2}{2.1} & 30.7\ci{29.2}{32.2} & 25.0\ci{23.6}{26.5} & 43.1\ci{41.5}{44.7} & 46.5\ci{44.9}{48.2} & 62.8\ci{61.2}{64.4} \\
MESA-Spatial & 1.0\ci{0.4}{2.3} & 35.0\ci{30.9}{39.3} & 23.2\ci{19.7}{27.1} & 42.2\ci{37.9}{46.6} & 53.6\ci{49.2}{57.9} & 68.6\ci{64.4}{72.5} \\
MESA-Instance & 1.9\ci{1.2}{2.9} & 22.7\ci{20.2}{25.4} & 22.7\ci{20.2}{25.4} & 34.0\ci{31.1}{37.0} & 39.1\ci{36.1}{42.2} & 55.4\ci{52.3}{58.5} \\
MESA-Composite & 0.3\ci{0.1}{0.9} & 17.7\ci{15.5}{20.2} & 15.0\ci{12.9}{17.3} & 27.5\ci{24.8}{30.3} & 28.9\ci{26.2}{31.8} & 46.7\ci{43.6}{49.8} \\
MESA-Category & 0.7\ci{0.3}{1.4} & 7.7\ci{6.2}{9.5} & 14.1\ci{12.1}{16.4} & 11.6\ci{9.8}{13.7} & 8.4\ci{6.8}{10.3} & 26.4\ci{23.8}{29.2} \\
\midrule
Average & 1.1\ci{0.8}{1.4} & 22.8\ci{21.7}{23.8} & 20.0\ci{19.0}{21.0} & 31.7\ci{30.5}{32.8} & 35.3\ci{34.1}{36.5} & 52.0\ci{50.7}{53.2} \\
\bottomrule
\end{tabular}
\end{adjustbox}
\caption{\textbf{Single-arm success rates (\%)} with 95\% Wilson intervals for \autoref{tab:main-results}.}
\label{tab:main-results-ci}
\end{table}

\begin{table}[h]
\centering
\scriptsize
\begin{adjustbox}{max width=\linewidth}
\begin{tabular}{lcccccccc}
\toprule
\rowcolor[HTML]{FFDEB4}
Setting & DP & PG-Bin & PG-FM & GR00T-N1.6 & $\pi_0$ & FAST & $\pi_{0.5}$ & GPT-6.1-Sol \\
\midrule
BiMESA-ID & 13.0\ci{11.8}{14.3} & 26.7\ci{25.1}{28.4} & 43.8\ci{42.0}{45.7} & 47.0\ci{45.2}{48.9} & 45.9\ci{44.1}{47.7} & 53.7\ci{51.8}{55.5} & 68.7\ci{66.9}{70.3} & 41.2\ci{37.3}{45.3} \\
BiMESA-Spatial & 18.8\ci{17.1}{20.7} & 30.3\ci{28.2}{32.4} & 50.8\ci{48.5}{53.0} & 46.2\ci{43.9}{48.4} & 49.2\ci{47.0}{51.5} & 58.1\ci{55.9}{60.3} & 70.5\ci{68.4}{72.5} & 41.8\ci{37.0}{46.9} \\
BiMESA-Instance & 9.4\ci{7.7}{11.5} & 13.9\ci{11.8}{16.3} & 28.6\ci{25.7}{31.6} & 36.7\ci{33.6}{39.9} & 23.9\ci{21.2}{26.8} & 36.7\ci{33.6}{39.9} & 48.1\ci{44.9}{51.4} & 43.9\ci{36.8}{51.2} \\
BiMESA-Composite & 5.7\ci{4.3}{7.6} & 14.4\ci{12.1}{17.1} & 17.9\ci{15.3}{20.8} & 22.1\ci{19.3}{25.2} & 18.4\ci{15.8}{21.3} & 28.8\ci{25.7}{32.1} & 37.5\ci{34.1}{41.0} & 34.7\ci{27.5}{42.6} \\
BiMESA-Category & 5.8\ci{4.4}{7.5} & 6.0\ci{4.6}{7.8} & 10.9\ci{9.0}{13.2} & 26.1\ci{23.3}{29.2} & 10.6\ci{8.7}{12.8} & 19.8\ci{17.2}{22.6} & 18.6\ci{16.1}{21.3} & 52.4\ci{44.9}{59.7} \\
\midrule
Average & 10.6\ci{9.9}{11.2} & 18.3\ci{17.5}{19.0} & 30.4\ci{29.5}{31.3} & 35.6\ci{34.6}{36.6} & 29.6\ci{28.7}{30.5} & 39.4\ci{38.4}{40.4} & 48.7\ci{47.7}{49.7} & 42.8\ci{40.5}{45.1} \\
\bottomrule
\end{tabular}
\end{adjustbox}
\caption{\textbf{Bimanual success rates (\%)} with 95\% Wilson intervals for \autoref{tab:main-results}.}
\label{tab:main-results-bimesa-ci}
\end{table}

\begin{table}[h]
\centering
\scriptsize
\begin{adjustbox}{max width=\linewidth}
\begin{tabular}{lcccccc}
\toprule
\rowcolor[HTML]{FFDEB4}
Setting & DP & PG-FM & GR00T-N1.6 & $\pi_0$ & FAST & $\pi_{0.5}$ \\
\midrule
MESA-ID & 18.9\ci{14.8}{23.7} & 80.0\ci{77.8}{82.1} & 71.4\ci{68.8}{73.8} & 79.7\ci{77.8}{81.4} & 83.7\ci{82.0}{85.3} & 87.6\ci{86.3}{88.8} \\
MESA-Spatial & 11.4\ci{5.0}{24.0} & 77.4\ci{71.6}{82.4} & 53.7\ci{47.0}{60.2} & 73.0\ci{67.6}{77.8} & 78.6\ci{73.9}{82.6} & 84.5\ci{80.6}{87.7} \\
MESA-Instance & 23.5\ci{15.6}{33.8} & 66.6\ci{61.4}{71.4} & 60.5\ci{55.5}{65.3} & 66.5\ci{62.3}{70.5} & 69.0\ci{65.0}{72.6} & 76.9\ci{73.7}{79.9} \\
MESA-Composite & 6.4\ci{2.2}{17.2} & 65.1\ci{59.2}{70.5} & 56.2\ci{50.2}{62.0} & 65.5\ci{60.8}{69.9} & 66.3\ci{61.7}{70.6} & 81.5\ci{78.1}{84.5} \\
MESA-Category & 12.1\ci{6.0}{22.9} & 32.6\ci{27.0}{38.8} & 42.5\ci{37.3}{47.8} & 33.1\ci{28.4}{38.2} & 28.0\ci{23.2}{33.3} & 44.1\ci{40.1}{48.1} \\
\midrule
Average & 14.4\ci{10.9}{18.0} & 64.3\ci{62.1}{66.6} & 56.9\ci{54.5}{59.2} & 63.6\ci{61.7}{65.5} & 65.1\ci{63.3}{66.9} & 74.9\ci{73.5}{76.3} \\
\bottomrule
\end{tabular}
\end{adjustbox}
\caption{\textbf{Single-arm language following rates (\%)} with 95\% Wilson intervals for \autoref{tab:language-following}.}
\label{tab:language-following-ci}
\end{table}

\begin{table}[h]
\centering
\scriptsize
\begin{adjustbox}{max width=\linewidth}
\begin{tabular}{lcccccccc}
\toprule
\rowcolor[HTML]{FFDEB4}
Setting & DP & PG-Bin & PG-FM & GR00T-N1.6 & $\pi_0$ & FAST & $\pi_{0.5}$ & GPT-6.1-Sol \\
\midrule
BiMESA-ID & 34.9\ci{32.1}{37.9} & 78.7\ci{76.0}{81.2} & 84.6\ci{82.7}{86.4} & 87.7\ci{86.0}{89.2} & 73.4\ci{71.3}{75.4} & 86.9\ci{85.3}{88.4} & 92.9\ci{91.7}{93.9} & 73.2\ci{68.1}{77.8} \\
BiMESA-Spatial & 45.3\ci{41.8}{48.7} & 75.7\ci{72.5}{78.6} & 86.2\ci{84.0}{88.1} & 77.5\ci{74.9}{79.8} & 72.6\ci{70.1}{75.0} & 83.0\ci{80.9}{84.9} & 89.1\ci{87.5}{90.6} & 69.7\ci{63.5}{75.3} \\
BiMESA-Instance & 25.5\ci{21.1}{30.5} & 53.0\ci{46.6}{59.2} & 64.4\ci{59.6}{69.0} & 73.3\ci{69.1}{77.2} & 48.0\ci{43.4}{52.6} & 64.8\ci{60.6}{68.9} & 74.0\ci{70.3}{77.4} & 65.8\ci{57.0}{73.7} \\
BiMESA-Composite & 14.5\ci{11.0}{19.0} & 65.1\ci{57.5}{71.9} & 66.7\ci{59.9}{72.8} & 71.2\ci{65.1}{76.7} & 42.3\ci{37.1}{47.8} & 66.7\ci{61.4}{71.6} & 80.1\ci{75.6}{83.9} & 74.3\ci{63.0}{83.1} \\
BiMESA-Category & 15.7\ci{12.1}{20.2} & 27.4\ci{21.5}{34.2} & 30.1\ci{25.3}{35.4} & 55.6\ci{50.7}{60.4} & 21.2\ci{17.6}{25.3} & 35.1\ci{30.9}{39.4} & 28.7\ci{25.1}{32.6} & 79.5\ci{71.1}{85.9} \\
\midrule
Average & 27.2\ci{25.5}{28.9} & 60.0\ci{57.5}{62.4} & 66.4\ci{64.4}{68.4} & 73.1\ci{71.2}{74.9} & 51.5\ci{49.8}{53.2} & 67.3\ci{65.6}{69.0} & 73.0\ci{71.6}{74.3} & 72.5\ci{69.1}{75.9} \\
\bottomrule
\end{tabular}
\end{adjustbox}
\caption{\textbf{Bimanual language following rates (\%)} with 95\% Wilson intervals for \autoref{tab:language-following}.}
\label{tab:language-following-bimesa-ci}
\end{table}

\begin{table}[h]
\centering
\scriptsize
\begin{adjustbox}{max width=\linewidth}
\begin{tabular}{lccccc}
\toprule
\rowcolor[HTML]{FFDEB4}
 & ID & Spatial & Instance & Composite & Category \\
\midrule
DP & 1.6\ci{1.2}{2.1} & 1.0\ci{0.4}{2.3} & 1.9\ci{1.2}{2.9} & 0.3\ci{0.1}{0.9} & 0.7\ci{0.3}{1.4} \\
DP+Aux & 2.3\ci{1.8}{2.8} & 2.4\ci{1.4}{4.1} & 2.2\ci{1.5}{3.3} & 0.6\ci{0.3}{1.3} & 0.3\ci{0.1}{0.9} \\
\quad {\tiny McNemar $p$} & {\tiny 0.03} & {\tiny 0.07} & {\tiny 0.76} & {\tiny 0.51} & {\tiny 0.34} \\
\midrule
$\pi_0$ & 43.1\ci{41.5}{44.7} & 42.2\ci{37.9}{46.6} & 34.0\ci{31.1}{37.0} & 27.5\ci{24.8}{30.3} & 11.6\ci{9.8}{13.7} \\
$\pi_0$+Aux & 43.4\ci{41.7}{45.0} & 54.4\ci{50.0}{58.7} & 33.2\ci{30.4}{36.2} & 28.8\ci{26.1}{31.7} & 8.7\ci{7.1}{10.6} \\
\quad {\tiny McNemar $p$} & {\tiny 0.79} & {\tiny $<$0.001} & {\tiny 0.68} & {\tiny 0.44} & {\tiny 0.02} \\
\midrule
$\pi_{0.5}$ & 62.8\ci{61.2}{64.4} & 68.6\ci{64.4}{72.5} & 55.4\ci{52.3}{58.5} & 46.7\ci{43.6}{49.8} & 26.4\ci{23.8}{29.2} \\
$\pi_{0.5}$+Aux & 59.0\ci{57.4}{60.6} & 69.2\ci{65.0}{73.1} & 53.6\ci{50.5}{56.7} & 45.6\ci{42.5}{48.7} & 23.0\ci{20.5}{25.7} \\
\quad {\tiny McNemar $p$} & {\tiny $<$0.001} & {\tiny 0.86} & {\tiny 0.34} & {\tiny 0.55} & {\tiny 0.04} \\
\bottomrule
\end{tabular}
\end{adjustbox}
\caption{\textbf{Auxiliary co-training success rates (\%)} with 95\% Wilson intervals for \autoref{tab:diversity}. McNemar $p$ compares each model with its +Aux counterpart.}
\label{tab:diversity-ci}
\end{table}

%% file: 7.6-training-details.tex
\section{Training Details for Baselines}
\label{sec:supp_training_details}
Here we share additional details about our baselines.

\subsection{Shared Details}

Unless noted otherwise, all VLA / VLM-based policies except for GR00T are trained for 50k gradient steps with a batch size of 128. For GR00T, we use a batch size of 1024 as recommended by the authors on single-arm MESA and a batch size of 128 on BiMESA. The PaliGemma baselines and $\pi$ models are trained using the JAX version of the official \texttt{openpi} repository, and GR00T is trained using the official \texttt{Isaac-GR00T} repository.
We use the default optimization parameters from these repositories for all experiments.
In the single-arm setting, all models are trained with a left shoulder camera image and a wrist camera image. The bimanual setup provides two wrist cameras and an egocentric camera.
All models are evaluated using the final model at the end of training.
All models use relative joint angle action space, which is a common choice in the pretraining data for our main VLA baselines.
Unless otherwise noted, all models are trained with a chunk size matching the control frequency of 20 Hz and we replan after executing the first 5 actions.
In general, we didn't tune hyperparameters unless performance was surprisingly poor.

\paragraph{Policy inputs and privileged information.}
During training and evaluation, policies are conditioned only on camera images, proprioception (joint positions and gripper widths), and the language instruction; demonstration actions provide the training targets.
No baseline receives ground-truth object poses, object-identity metadata, task predicates, or symmetry-derived grasp yaw as policy inputs.
Privileged simulator information is used by the demonstration-generation pipeline and by the evaluator to score task completion, not supplied to the policies.

\subsection{Baseline-Specific Details}
\begin{itemize}
    \item \textbf{Multitask Diffusion Policy (DP)}~\cite{diffusionpolicy}: We use the DiT Block Policy from \citet{dasari2024ditpi} with 6 blocks, an embed dimension of 512 and feedforward dim of 3200, yielding a total of 80M parameters. This architecture uses a ResNet-18 \cite{resnet} backbone with FiLM \cite{film} to extract $7 \times 7$ feature vectors from each image. These image features are concatenated with the projected state and language embedding before being fed to a transformer encoder. The outputs of the encoder are pooled and used together with the diffusion timestep as AdaLN \cite{peebles2023scalablediffusionmodelstransformers} conditioning for the diffusion transformer decoder.
    We use the AdamW \cite{loshchilov2019decoupledweightdecayregularization} optimizer with a learning rate of $1 \times 10^{-4}$, weight decay of $1 \times 10^{-6}$ and cosine annealing learning rate decay. We train for 320,000 gradient steps.
    
    \item \textbf{PaliGemma-Binning (PG-Bin)}: This baseline finetunes a PaliGemma-3B \cite{beyer2024paligemmaversatile3bvlm} backbone to predict actions discretized using RT-2-style \cite{rt-2} binning. We use the implementation in the \texttt{openpi} repository, originally intended for use as a baseline in \citet{atreya2025roboarena}. The baseline discretizes each action dimension into 256 bins. Since this baseline is extremely token-inefficient, we only predict chunks of 5 actions to save compute. The model autoregressively predicts these binned actions.

    \paragraph{PG-Bin single-arm omission.}
    We omit PG-Bin results on the single-arm MESA suites due to a bug in our PG-Bin inference implementation that affected only the single-arm setting. The issue did not affect PG-Bin training, the BiMESA PG-Bin evaluation, or any other baseline. We therefore report PG-Bin only on BiMESA and exclude it from single-arm averages and comparisons.
    
    \item \textbf{PaliGemma-Flow Matching (PG-FM)}: This baseline finetunes a PaliGemma-3B backbone with a flow matching head; i.e., $\pi_0$ \cite{black2024pi0} without robot pretraining. It is implemented in the \texttt{openpi} repository and is identical to the $\pi_0$ baseline, but using PaliGemma pretrained weights for the backbone and initializing the action head from scratch.
    \item \textbf{GR00T-N1.6}~\cite{nvidia2025gr00tn1openfoundation}: This baseline attaches a flow matching action head to a Cosmos-2B VLM \cite{nvidia2025cosmosreason1physicalcommonsense}. It is implemented in the official \texttt{Isaac-GR00T} repository and is trained using the default configurations. Notably, this includes leaving the backbone frozen. On single-arm MESA, we originally trained with a batch size of 128 and retrained with a batch size of 1024, which we found to lead to a modest improvement. BiMESA results use a batch size of 128.
    \item $\boldsymbol{\pi}\mathbf{_0}$~\cite{black2024pi0}, $\boldsymbol{\pi}\mathbf{_0}$-\textbf{FAST}~\cite{pertsch2025fastefficientactiontokenization}, $\boldsymbol{\pi}\mathbf{_{0.5}}$~\cite{intelligence2025pi05}: $\pi_0$ attaches a flow matching head to a PaliGemma-3B backbone and pretrains on a large cross-embodiment robot dataset. $\pi_0$-FAST replaces the flow matching head with discretized action chunk prediction. $\pi_{0.5}$ trains a $\pi_0$-like architecture with knowledge insulation \cite{driess2025knowledgeinsulatingvisionlanguageactionmodels}. These were all trained using the default configuration from the \texttt{openpi} repository. For $\pi_{0.5}$ we finetuned with and without knowledge insulation, found this not to make much difference, and report the version without.
\end{itemize}

\subsection{Training Compute Usage}

An important goal for our framework is to allow academic researchers to quickly iterate with minimal compute requirements.
We find that \abbr{}'s dataset was small enough to achieve saturation after a relatively quick training cycle.
For the \texttt{openpi} repository models, the full 50k training steps took 48 hours on four L40S GPUs or 21 hours on two H100s.
We also saw that $\pi_{0.5}$ achieved strong performance even after 5k gradient steps, and that it achieved strong performance with only 10\% of our data, meaning it may be possible to shrink the budget even further.
For GR00T, the version with a batch size 128 took about 12 hours on one H100 and the version with batch size 1024 took about 30 hours with four H100s.